\documentclass[11pt,letterpaper]{article}

\usepackage[letterpaper,top=1in,bottom=1in,left=0.7in,right=0.7in]{geometry}
\usepackage{graphicx}

\let\jsacOrigScalebox\scalebox
\renewcommand{\scalebox}[2]{\ifdim#1pt<1pt\relax\resizebox{\linewidth}{!}{#2}\else\jsacOrigScalebox{#1}{#2}\fi}
\usepackage{amsmath,amsfonts,amssymb}
\usepackage[ruled,vlined,linesnumbered]{algorithm2e}
\usepackage{array}
\usepackage{textcomp}
\usepackage{stfloats}
\usepackage{url}
\usepackage{verbatim}
\usepackage{cite}
\usepackage[hidelinks]{hyperref}
\usepackage{booktabs}
\usepackage{nicefrac}
\usepackage{microtype}
\usepackage{xcolor}
\usepackage{footnote}
\usepackage{enumitem}
\usepackage{multirow}
\usepackage{makecell}
\usepackage{mathtools}
\usepackage{amsthm}
\usepackage{newtxtext}
\usepackage{caption}
\usepackage{subcaption}
\usepackage[capitalize,noabbrev]{cleveref}

\theoremstyle{plain}
\newtheorem{theorem}{Theorem}

\newtheorem{lemma}{Lemma}

\theoremstyle{definition}

\newtheorem{remark}{Remark}

\newcommand{\algoname}{{DD-SGD-GC}}
\newcommand{\sysname}{{\textsc{DeCo-SGD}}}
\newcommand{\decoTimelineFigWidth}{0.86\linewidth}

\newcommand{\sr}{}
\newcommand{\er}{}

\AtBeginDocument{\colorlet{red}{black}\colorlet{blue}{black}}

\providecommand{\IEEEPARstart}[2]{\textbf{#1}#2}
\newenvironment{IEEEkeywords}{\par\medskip\noindent\textbf{Index Terms---}\,}{\par\medskip}

\title{Taming Latency and Bandwidth: A Theoretical Framework and Adaptive Algorithm for Communication-Constrained Training\\[0.4em]\large(Extended Version)}

\providecommand{\blfootnote}[1]{\begingroup\renewcommand{\thefootnote}{}\footnotetext{#1}\endgroup}

\author{%
  Rongwei Lu$^{1,*}$,\quad Chunyang Li$^{1,*}$,\quad Jingyan Jiang$^{1,*}$,\quad Xingguang Wei$^{2}$,\quad Zhi Wang$^{1,\dagger}$ \\[0.7em]
  {\small $^{1}$Shenzhen International Graduate School, Tsinghua University} \\
  {\small $^{2}$University of Science and Technology of China} \\[0.7em]
  {\small \texttt{lurw24@mails.tsinghua.edu.cn} (R. Lu)\quad \texttt{wangzhi@sz.tsinghua.edu.cn} (Z. Wang$^{\dagger}$)}%
}

\date{}

\begin{document}

\maketitle
\blfootnote{$^{*}$The first three authors contributed equally to this work. $^{\dagger}$Corresponding author: Zhi Wang.
This is an extended version of the article accepted for publication in the \emph{IEEE Journal on Selected Areas in Communications} (DOI: 10.1109/JSAC.2026.3735988); it includes the complete appendix.
\textcopyright\ 2026 IEEE. Personal use is permitted, but republication/redistribution requires IEEE permission. See \url{https://www.ieee.org/publications/rights/index.html} for more information.}

\begin{abstract}
Regional energy caps constrain the scalability of individual data centers for large-scale model training. This single-center training paradigm works when model size remains manageable, but exponential growth in model size and computational demand challenges it. A natural alternative is to distribute training across multiple data centers over wide-area networks. This pools distributed resources, but suffers from high latency and low, time-varying bandwidth, sharply reducing throughput.
Employing gradient compression and delayed aggregation jointly can alleviate communication problems but introduces a complex three-way trade-off among compression ratio, staleness (delayed synchronization steps), and convergence rate. Existing work lacks theoretical guidance and thus proposes only fixed strategies, agnostic to computation and communication conditions. We address this with a new theoretical framework, decomposing the joint optimization problem into a traditional process plus multiple analyzable noise terms. Our analysis derives the first convergence rate for this setting and shows that increasing staleness exponentially amplifies the detrimental effect of compression. Leveraging these insights, we propose \sysname, which dynamically adjusts the compression ratio and staleness based on the real-time communication and computation conditions. \sysname\ achieves up to \sr$10.82\times$\er\ and \sr$1.36\times$\er\ speed-ups\ over distributed SGD and the state-of-the-art strategy in high-latency and low, varying bandwidth networks, respectively.
\end{abstract}


\section{Introduction}
\label{sec:intro}

\IEEEPARstart{T}{he} rapid progress of large language models (LLMs) (\textit{e.g.}, GPT-4 \cite{gpt-4}, DeepSeek-v3 \cite{deepseek}) has been accompanied by an explosive growth in model parameters. This expansion drives a steep rise in the computational resources required for training \cite{ScalaLaws1}. Training in a single AI computing center, where thousands of GPUs are tightly coupled via high-speed interconnects (\textit{e.g.}, PCIe and NVLink \cite{nvidia}), is increasingly constrained by regional energy capacity. This bottleneck is already evident in major hubs; for instance, data centers in Virginia are projected to consume up to $51\%$ of the state’s electricity by 2030 \cite{GEI-Datacenter-25}, while utilities in Ohio have imposed moratoriums on new connections due to transmission limits \cite{Williams2024}. As model sizes continue to grow exponentially, the paradigm of training in a hyperscale compute center is approaching its scalability limits. This observation motivates a new phase of large-scale training that extends beyond a single compute center or energy domain. Distributed training over wide-area networks (WANs) offers a practical and sustainable way to overcome regional power limits and coordinate multiple compute centers globally.

\begin{figure}[t]
    \centering
    \includegraphics[width=0.45\textwidth]{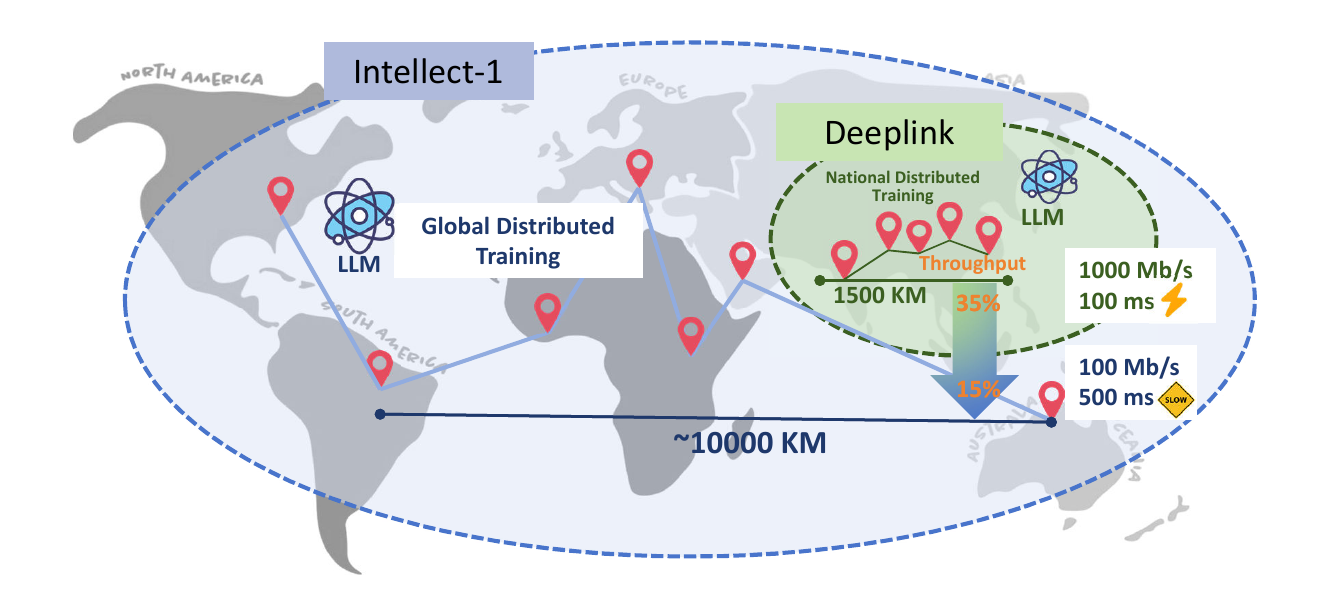}
    \caption{Pioneering LLM training over WANs. DeepLink connects compute clusters across two cities over $1{,}500$ km and Intellect-1 trains across continents. Training over WANs faces high-latency and low-bandwidth network conditions, leading to severe degradation in system throughput.}
    \label{Traing in WAN}
\end{figure}

Recently, some pioneering projects have already demonstrated the feasibility of training over WANs as shown in Fig.~\ref{Traing in WAN}.
DeepLink connected datacenters across two cities in China, more than $1{,}500$ km apart, to train a model exceeding $100$B parameters \cite{deeplink}. Similarly,
Intellect-1 collaboratively trained a $10$B-parameter model on up to $112$ H$100$s distributed across five countries and three continents \cite{jaghouar2024opendiloco}. These highlight the potential of multi-region training over WANs to mitigate regional power constraints and begin to harness geographically distributed compute at scale.
Distributed training over WANs is fundamentally communication-constrained, facing high latency and limited bandwidth. In this setting, the standard data-parallel paradigm requires workers to perform local computation and then exchange full gradients with the server. The server then aggregates the gradients, updates the model, and broadcasts the new parameters to workers. 
Among algorithms employing this paradigm, Distributed Stochastic Gradient Descent
(D-SGD) and its variants are the most widely adopted, thanks to their clean
iterative formulation and practical scalability \cite{SIGCOMM24StellaTrain}.
However, under WAN conditions, the need to transmit full gradients leads to severe degradation in end-to-end training throughput (\textit{i.e.}, the number of samples processed per second) for these methods \cite{dong25emnlpsurvey}.

\begin{figure}[t]
    \centering
    \includegraphics[width=0.45\textwidth]{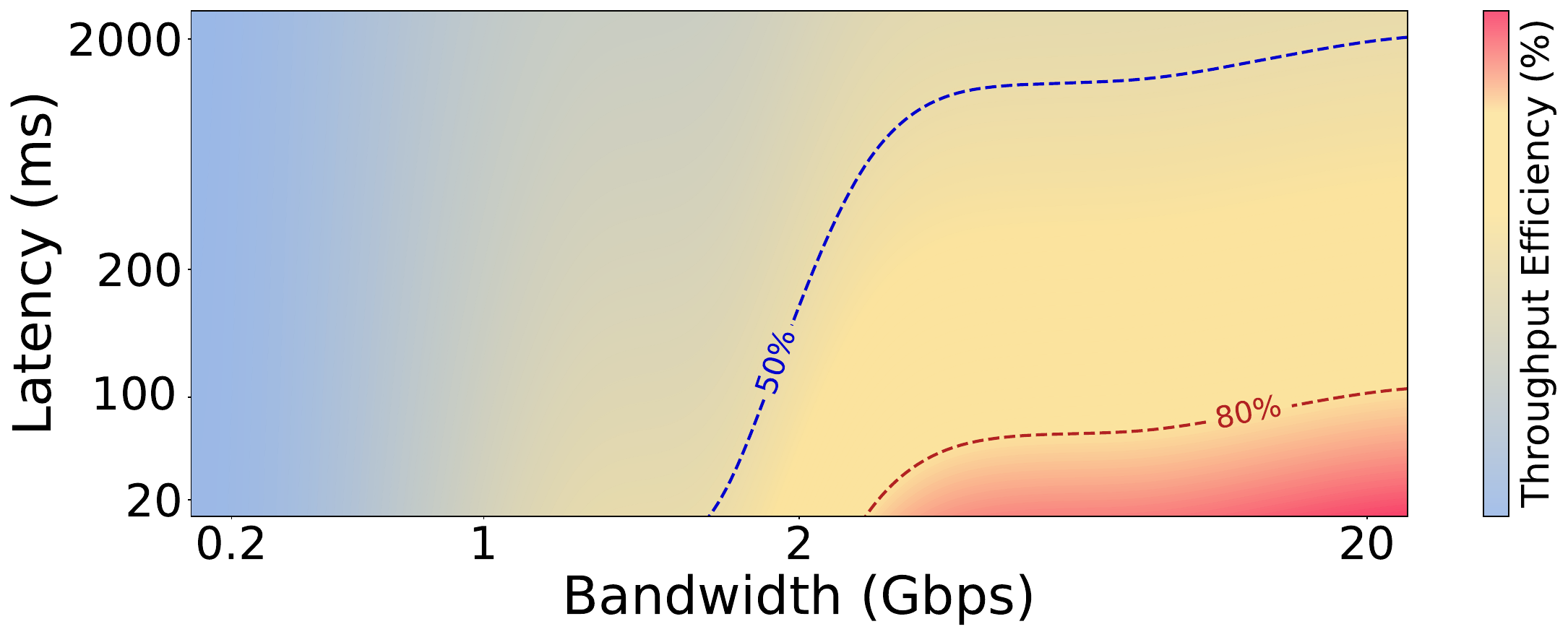}
    \caption{The heatmap of throughput efficiency ($\%$) for D-SGD with four nodes training GPT-2 \cite{gpt2}, under different latency and bandwidth conditions. Each node is equipped with one A$40$ GPU. The throughput efficiency at $(x, y)$ is defined as the throughput at $(x, y)$ divided by the maximum achievable throughput of the machines.}
    \label{throughput heatmap}
\end{figure}

To quantify the impact of network conditions, we conduct controlled experiments training GPT-2 under varying bandwidth and latency configurations \cite{amazon}. Fig.~\ref{throughput heatmap} presents the
resulting throughput efficiency, defined as end-to-end training throughput (\textit{i.e.}, the number of samples processed per second)
normalized by the maximum achievable throughput\footnote{This is determined by the compute capacity of the machines, without network constraints.}. \sr By mapping representative WAN configurations to this heatmap, we observe severe throughput degradation. For illustration, we consider two settings motivated by the cross-city deployment of DeepLink and the cross-continent deployment of Intellect-1: $(1{,}000\ \mathrm{Mb/s}, 100\ \mathrm{ms})$ and $(100\ \mathrm{Mb/s}, 500\ \mathrm{ms})$, respectively. The two settings fall into regions with throughput efficiencies of only about $35\%$ and $15\%$, respectively.\er\ Taken together, this inefficiency underscores that overcoming the communication bottleneck is not merely an optimization, but a prerequisite for unlocking the true scale of training across regions.

Gradient compression and delayed aggregation are common strategies to improve the throughput of D-SGD in a lossy manner, but joint usage introduces a complex three-way trade-off. Gradient compression reduces communication overhead by compressing gradient information before transmission, to address the low bandwidth constraints \cite{yan2022acJsac}.  Delayed aggregation \cite{DGA} applies updates using gradients computed several iterations earlier to alleviate the high latency problem.  While combining these strategies offers a promising path to overcome the problem of low-bandwidth and high-latency at the same time, it introduces intricate trade-offs in two key dimensions: (1) \textit{Iteration-to-accuracy}:
The impact of compressing delayed gradients on convergence remains uncertain. It is unclear to what extent aggregating delayed compressed gradients will affect model accuracy, due to the lack of theoretical analysis in prior work. (2) \textit{Time-to-iteration}: The effect of staleness and compression ratio on communication-computation overlap is not fully understood. Specifically, the existence of a threshold compression ratio, below which further compression no longer improves per-iteration time, remains unexplored. This threshold varies dynamically with the bandwidth and staleness. 

To minimize the wall-clock time required to reach the target accuracy, we analyze these two dimensions separately. First, in iteration-to-accuracy, we propose a new analytical framework called the Nested Virtual Sequence. This method builds on the original virtual sequence by constructing multiple virtual queues, decomposing the complex noise term into multiple analyzable noise terms. With this framework, \textit{we are the first to derive the convergence rate of distributed SGD with delayed aggregation and gradient compression (\algoname).} This convergence rate establishes the iteration-to-accuracy relationship and reveals that \textit{increasing staleness exponentially worsens the negative effect of the gradient compression on the model convergence}. Second, we use mathematical induction to model the relationship between the end-to-end time and the number of iterations in \algoname. Consequently, we establish the time-to-accuracy relationship and propose an adaptive algorithm that can select the optimal compression ratio and staleness in the challenging dynamic network conditions, thus achieving the balanced three-way trade-off during the whole training. 

In summary, the contributions of our work are as follows:

\begin{itemize}
    \item We propose the Nested Virtual Sequence, a theoretical framework to analyze the complex distributed SGD variant, by reducing it to a standard D-SGD process and several analytical noise terms. We are the first to derive the convergence rate of \algoname,  revealing that delayed aggregation exponentially worsens the negative effect of the compression on the training process.
    
    \item We mathematically establish an approximate relationship between end-to-end time and accuracy. Based on this, we propose \sysname, a joint optimization algorithm that adaptively determines the optimal compression ratio and staleness under varying training environments.
    
    \item We apply \sysname\, in challenging network environments (average bandwidth $\leq$ $1$ Gbps, latency $\geq$ $100$ ms). We conduct extensive experiments that cover diverse model architectures (ResNet, GPT, ViT), various network conditions, and system scalability. Experimental results demonstrate that our design achieves up to \sr$10.82\times$\er\ and \sr$1.36\times$\er\ speed-ups\ over D-SGD and the SOTA static strategy, respectively.
\end{itemize}

\section{Preliminaries}\label{preli}

In this work, we focus on distributed machine learning (DML) in data-parallel mode using \algoname \ as the optimizer.

\subsection{The Optimization Problem of DML}
In the data-parallel paradigm, the training dataset is partitioned across \( n \) worker nodes, each holding a subset of the data. The optimization problem of DML aims to minimize a global loss function \( f(\cdot) \) based on the model parameter \( \mathbf{x} \in \mathbb{R}^d \), and \( f(\cdot) \) is defined as the average of local loss functions of the $i$-th worker \( f_i(\cdot) \):
\begin{equation}
\underset{\mathbf{x}}{\min} \left[ f(\mathbf{x}) := \frac{1}{n} \sum_{i=1}^n f_i(\mathbf{x}) \right]. \nonumber
\end{equation}
\subsection{D-SGD and Its Variants}
\subsubsection{D-SGD}
At the \( t \)-th iteration, the $i$-th worker computes gradients \( \mathbf{g}_t^i \) based on its local data and model parameters at the $t$-th iteration \( \mathbf{x}_t \). \( \gamma \) is the stepsize and these local gradients are aggregated to update the global model, following $\mathbf{x}_{t+1} = \mathbf{x}_t - \frac{\gamma }{n} \sum_{i=1}^n \mathbf{g}_t^i$. As shown in Fig.~\ref{D-SGD difference fig}, standard D-SGD follows a strict sequential execution. The computation phase is blocked until the communication phase is fully completed, and communication includes end-to-end latency and transmission time.
\subsubsection{D-SGD with Gradient Compression}

Gradient compression is a typical communication optimization method to reduce the communication volume during the training process by compressing transferred gradients \cite{cui2021slashing}.
Existing strategies generally fall into three categories: sparsification, quantization, and low-rank compression. Quantization compression \cite{ lQuant2025jsac} reduces the communication volume by decreasing the number of bits of gradients. However, it struggles to achieve aggressive compression ratios (such as $1\%$ or less) and lacks fine-grained adaptivity (\textit{e.g.}, changing ratios from $12\%$ to $10\%$ or $4$-bit into $4.5$-bit quantization) due to its discrete value selection. Low-rank compression \cite{vogels2019powersgd} is tailored for the all-reduce communication primitive and well-suited for training within a single data center. However, it incurs higher compression costs compared to other compression strategies and cannot leverage its advantages in decentralized training, which prefers peer-to-peer communication under a delayed aggregation strategy \cite{DGA}. Sparsification compression transmits \cite{dagctmc} only parts of gradients, offering two advantages for decentralized distributed training: (1) It enables aggressive compression of $1\%$, with the help of the Error Feedback (EF) mechanism\footnote{In this work, the gradient compression algorithm carries vanilla EF \cite{stich2019error} by default. EF is a mechanism that collects and reuses the errors from the gradient compression to retain the model accuracy. Other EF strategies, such as EF21 \cite{EF21} or EControl \cite{Stich2023EControl} are out of this paper's scope.}; (2) Compression ratios are continuously selectable, facilitating easy adaptive adjustment. We use sparsification compression and incorporate EF. We denote $\mathbf{C}_{\delta}(\cdot)$ as the compression operator with the compression ratio $\delta$ and $\Delta_t^i$ as the compressed update at the $t$-th iteration in $i$-th worker. $\mathbf{e}_t^i$ represents the error term of the $i$-th worker at the $t$-th iteration, storing the discrepancy between the compressed updates and the true ones. The update rules for D-SGD with gradient compression are $\Delta_t^i = \mathbf{C}_\delta( \mathbf{e}_t^i + \gamma \mathbf{g}_t^i)$, $\mathbf{e}_{t+1}^i = \mathbf{e}_t^i + \gamma \mathbf{g}_t^i  - \Delta_t^i$ and $\mathbf{x}_{t+1} = \mathbf{x}_t - \frac{1}{n} \sum_{i=1}^n \Delta_t^i$.
As shown in Fig.~\ref{D-SGD difference fig}, D-SGD with gradient compression maintains a serial execution flow, but reduces transmission time compared to D-SGD.

\subsubsection{D-SGD with Delayed Aggregation}

\begin{figure}[!t]
    \centering
    \providecommand{\decoTimelineFigWidth}{1\linewidth}
    \includegraphics[width=\decoTimelineFigWidth]{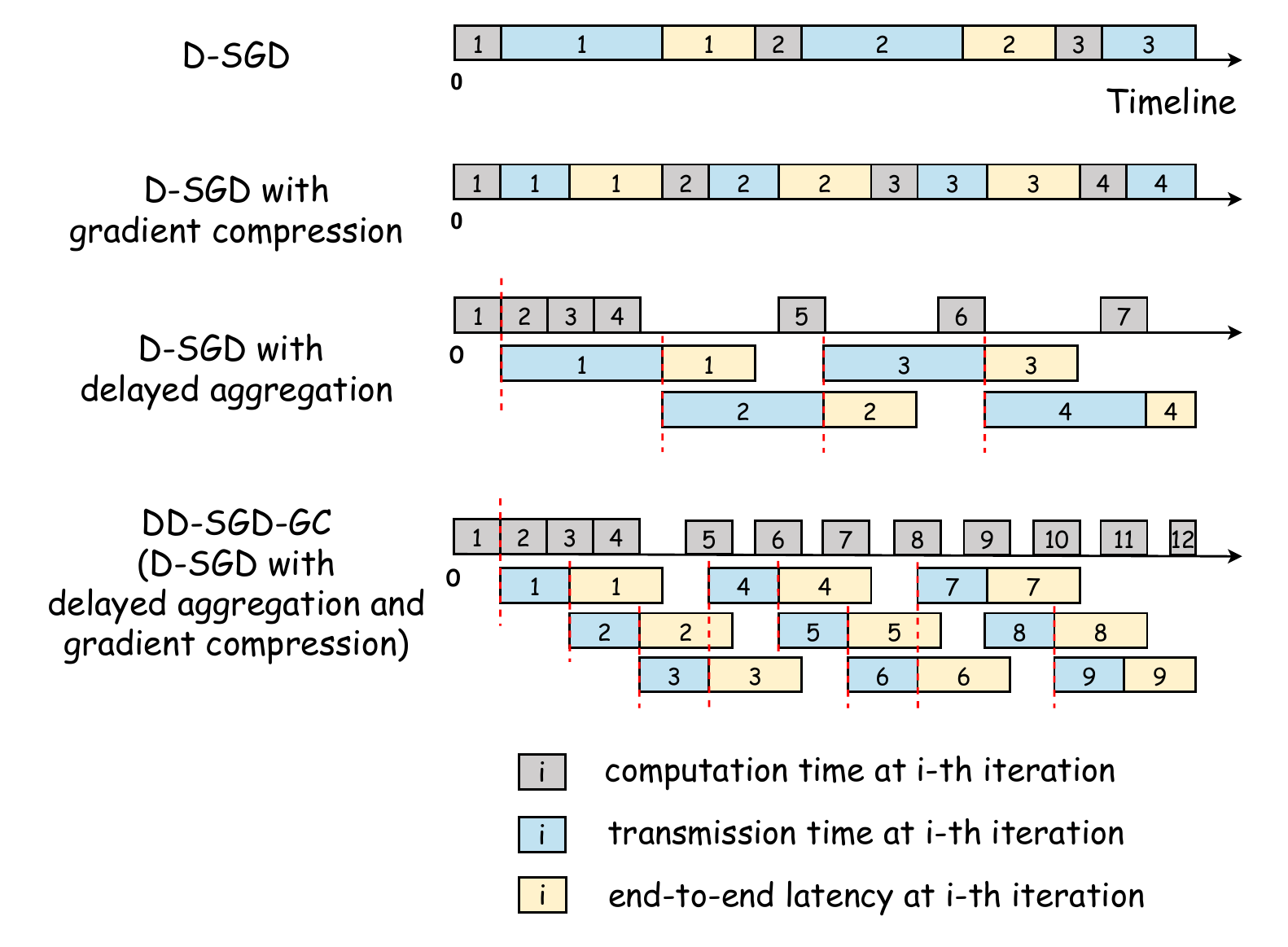}
    \caption{The running timelines for D-SGD and D-SGD with communication optimization strategies. 
    Both D-SGD and D-SGD with gradient compression are serial processes, with gradient compression reducing transmission time. D-SGD with delayed aggregation and \algoname \ operate in parallel.}
    \label{D-SGD difference fig}
\end{figure}

Delayed aggregation \cite{DGA} addresses communication delays by accounting for the delay staleness $\tau$ between the computation of gradients and their application to the model update \cite{stich2019error}. $\mathbf{g}_{t-\tau}^i$ represents the gradient computed by worker $i$ at the $(t-\tau)$-th iteration  and the update rule is given by
$\mathbf{x}_{t+1} = \mathbf{x}_t - \frac{\gamma }{n} \sum_{i=1}^n \mathbf{g}_{t-\tau}^i$. Unlike standard D-SGD, delayed aggregation enables pipelining: the system performs the $t$-th computation concurrently with the transmission of gradients from the $(t-\tau)$-th iteration (as shown in Fig.~\ref{D-SGD difference fig}). While the physical transmission duration remains unchanged, it is effectively hidden behind the computation. When further combined with gradient compression (\algoname), the transmission time is reduced.

\subsection{Virtual Sequence}

The Virtual Sequence method \cite{stich2018sparsified} is an analytical framework designed to simplify the convergence analysis of D-SGD variants like D-SGD with gradient compression. It defines an auxiliary sequence, $\tilde{\mathbf{x}}_t$, which evolves according to an idealized update rule and serves as a reference for the actual iterates, $\mathbf{x}_t$, accounting for noise introduced by compression and delay. We define the noise term as $B_t$, then we have $\tilde{\mathbf{x}}_t = \mathbf{x}_t - B_t$. We generalize updates as $\mathbf{v}_t$, (\textit{i.e.}, $\mathbf{x}_{t+1}=\mathbf{x}_t-\mathbf{v}_t$) and $B_t$ can be computed as  $B_{t+1} = B_{t} +  \frac{\gamma}{n} \sum_{i=1}^n \mathbf{g}_t^i -\mathbf{v}_t$. Consequently, we obtain $\tilde{\mathbf{x}}_{t+1}  = \mathbf{x}_{t+1} - B_{t+1}= \tilde{\mathbf{x}}_{t} - \frac{\gamma}{n} \sum_{i=1}^n \textbf{g}_t^i$. While effective for simple variants, this framework fails for complex coupled systems like \algoname. The failure arises because the noise term in \algoname \ involves a recursive coupling of compression errors and delay staleness. This dependency makes the noise term mathematically intractable under standard noise assumptions, necessitating a new analytical method to decouple these factors.

\section{Theoretical Analysis of \algoname}
 \label{section3}

 In this section, we establish the convergence guarantees for \algoname. The key notations are summarized in Table~\ref{table:notation}.
 The training procedure for \algoname \ is outlined in Algo.~\ref{DD-SGD-GC}.
Detailed proofs are provided in the Appendix (Sec.~\ref{sec:proof2T1}, ~\ref{proof2T2}).

\subsection{Regular Assumptions}

We adopt standard assumptions, following previous works \cite{cui2022infocom, stich2020communication}. 

\noindent \textbf{Assumption 1} ($L$-smoothness). We assume $L$-smoothness of $f_i,\ i\in [n]$, that is for all $\mathbf{x},\ \textbf{y} \in \mathbb{R}^d$:
\begin{equation}\label{L-smooth}
\lVert \nabla f_i(\textbf{y}) - \nabla f_i(\mathbf{x}) \rVert \leq L\lVert \textbf{y} - \mathbf{x} \rVert. 
\end{equation}

\noindent \textbf{Assumption 2} (Bounded gradient noise). We assume that we have access to stochastic gradient oracles $\mathbf{g}^i_t: \mathbb{R}^d \rightarrow  \mathbb{R}^d$ for each $f_i,\ i\in [n]$. We define the stochastic gradient noise of the $i$-th worker as $\boldsymbol{\xi}^i$.
Without loss of generality, we consider the case where $\boldsymbol{\xi}^i$ is uniformly bounded for all $\mathbf{x} \in \mathbb{R}^d, \ i \in [n]$:
\begin{equation}\label{xi}
\mathbf{g}^i_t=\nabla f_i(\mathbf{x}_t)+\boldsymbol{\xi} ^i,
\quad \mathbb{E}_{\boldsymbol{\xi} ^i}\boldsymbol{\xi} ^i = \textbf{0}_d, \quad \mathbb{E}_{\boldsymbol{\xi} ^ i} \lVert \boldsymbol{\xi} ^i \rVert^2 \leq \sigma^2.
\end{equation}

 \noindent \textbf{Assumption 3} (Measurement of data heterogeneity). We measure data dissimilarity using constants $\zeta^2 \geq 0$ and $Z^2\geq0$ that bound the variance across the $n$ nodes. We have:
\begin{equation}\label{zeta}
\frac{1}{n}\sum_{i\in[n]}\lVert \nabla f_i(\mathbf{x}) \rVert^2 \leq \zeta^2+Z^2\lVert\nabla f(\mathbf{x})\rVert^2,
\forall \mathbf{x} \in \mathbb{R}^d.
\end{equation}

For strongly convex objective functions, we additionally employ   \textbf{Assumption 4}.

\noindent \textbf{Assumption 4} ($\mu$-strongly convex). We assume $\mu$-strong convexity of $f_i, i\in [n]$, that is for all  $\mathbf{x},\textbf{y} \in \mathbb{R}^d$:
\begin{equation}\label{mu-convex}
\sr f_i(\mathbf{x}) - f_i(\textbf{y}) \geq  \langle \nabla f_i(\textbf{y}),\mathbf{x}-\textbf{y} \rangle + \frac{\mu}{2} \lVert \mathbf{x}-\textbf{y} \rVert^2.\er
\end{equation}

\begin{table}[!t]
\centering
\caption{Notation list.}
\scalebox{1}{
\begin{tabular}{ |c|c| } 
\hline
Notation & Description \\
\hline
$n$ & number of workers\\
\hline

\(f_i(\cdot)\) & the local loss function of worker $i$\\
\hline
\(f(\cdot)\) & the global loss function, \textit{i.e.}, $f(\mathbf{x})=\frac{1}{n} \sum_{i=1}^n f_i(\mathbf{x})$ \\
\hline
 $\mathbf{x}^i_t$ &  \makecell{the local model parameter in the $t$-th iteration \\ of the $i$-th worker}\\
\hline
 $\mathbf{x}_t$ &  the global model parameter in the $t$-th iteration \\
  \hline
 $\mathbf{g}^i_t$ & the stochastic gradient of $\textbf{x}_t$ of the worker $i$ \\
   \hline
 $\mathbf{g}_t$ & the average of $\mathbf{g}^i_t$, \textit{i.e.}, $\mathbf{g}_t=\frac{1}{n}\sum_{i=1}^n \mathbf{g}^i_t$\\
  \hline
  $\gamma$ &  the stepsize \\
\hline
\(\mathbf{T}\) & total number of iterations\\
\hline
$\delta$ & compression ratio (\(0 < \delta \leq 1\))\\
\hline
$\mathbf{C}_\delta(\cdot)$ & \makecell{the sparsification compressor with \\  compression ratio $\delta$}\\
 \hline
$\tau$ & the delayed staleness\\
\hline
$\mathbf{e}_t^i$ & \makecell{the local error term of the $i$-th worker \\ in the $t$-th iteration}\\

\hline
\(V_g\) & the volume of the gradient (bits) \\
 \hline
$E$ & the frequency to update  $\tau$ and $\delta$ with DeCo \\
\hline
\(\bar{a}\) &  \makecell{the average bandwidth (bits/s) of the slowest link \\ in previous $E$ iterations} \\
\hline
\(b\) & the network end-to-end latency (s) of the slowest link \\
 \hline
$T_{\text{comp}}$ & computation time per iteration (s)\\
\hline
$T_{\text{avg}}$ & average end-to-end training time per iteration (s)\\
\hline
$TS_t$ & end time of the $t$-th computation (s) \\
\hline
$TC_t$ &  end time of the $t$-th communication (s) \\
\hline
{$TM_t$} & {server-side message-ready time of the $t$-th aggregation (s)}\\
\hline
\end{tabular}
}
 \label{table:notation}
\end{table}

\begin{algorithm}[t]
\SetAlgoLined
\caption{\algoname}
\label{DD-SGD-GC}
\KwIn{$n$, $\gamma$, $\mathbf{T}$, $\mathbf{C}_\delta(\cdot)$, $\tau$}
\KwOut{$\textbf{x}_\mathbf{T}$}
Initialize $\mathbf{x}_0$, $\mathbf{e}_{0}^i = \mathbf{0}_d$\;
\For{$t\in[\mathbf{T}]$}{
    \tcc{Worker side}
    \For{$i\in[n]$}{ 
        Worker $i$ waits until it receives $\textbf{x}_{t}$\;
        Calculate $\mathbf{g}_t^i$ based on the global model $\mathbf{x}_t$\;
        $\Delta_t^i = \textbf{C}_{\delta} (\textbf{e}_t^i + \gamma\textbf{g}_t^i)$\;
        $\textbf{e}_{t+1}^i = \textbf{e}_{t}^i + \gamma \textbf{g}_{t}^i - \Delta_t^i$\;
        Send $\Delta_t^i$ to the server (non-blocking)\;
    }
    \tcc{Server side}
    Server waits until it receives $\Delta_{t-\tau}^i$ for all $i\in[n]$\;
    $\textbf{x}_{t+1} = \textbf{x}_t - \frac{1}{n}\sum_{i=1}^n \Delta_{t-\tau}^i$\;
    Broadcast $\textbf{x}_{t+1}$\;
}
\textbf{Return} $\textbf{x}_\mathbf{T}$\;
\end{algorithm}

\subsection{Theoretical Tool: Nested Virtual Sequences}

\begin{figure}[t]
    \centering
    \includegraphics[width=1\linewidth]{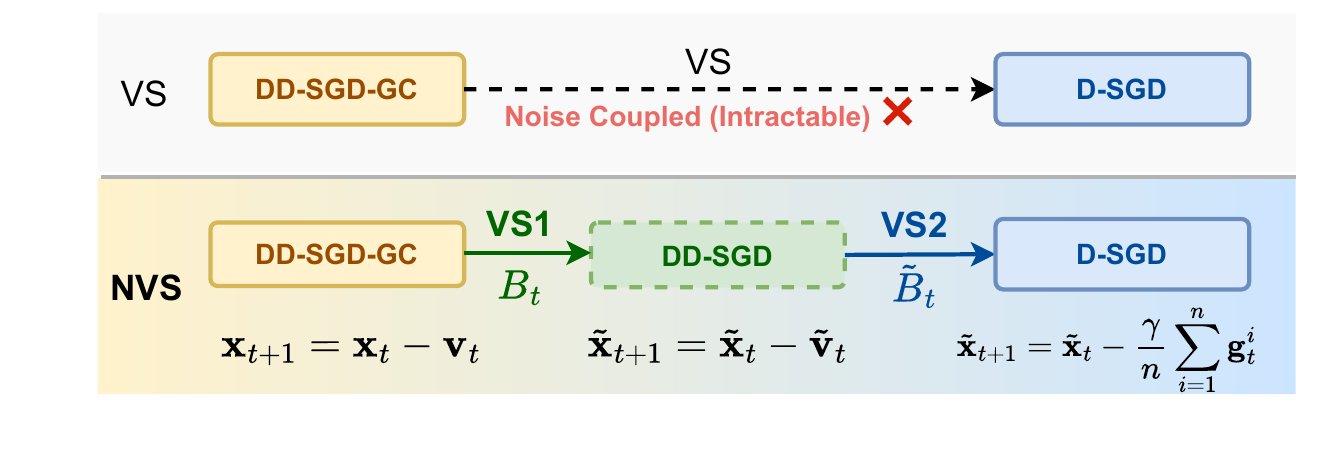}
    \caption{Comparison between VS and proposed NVS. NVS progressively simplifies complex D-SGD variants like \algoname \ to reach standard D-SGD. $B_t$ and $\tilde{B}_t$ are  analyzable noise terms decoupled from complex D-SGD variants. }
    \label{plt: NVS vs. VS}
\end{figure}

A major challenge in analyzing \algoname \ is the recursive coupling between compression error and delay staleness. Standard Virtual Sequence methods \cite{stich2018sparsified} fail to decouple these factors. To address this, we propose Nested Virtual Sequences (NVS). The core concept is to treat the actual update rule of the complex D-SGD variant, $\mathbf{x}_{t+1} = \mathbf{x}_t - \mathbf{v}_t$, as a process that can be progressively reduced to  standard D-SGD 
through hierarchical decoupling. 
We take DD-SGD-GC as an example and illustrate NVS-based analysis in Fig.~\ref{plt: NVS vs. VS}.

\noindent \textbf{Step 1: Decoupling D-SGD with delayed aggregation from \algoname.} We define the first virtual sequence $\tilde{\mathbf{x}}_{t}$ and the noise term $B_t$ following:
\begin{eqnarray}
    \tilde{\mathbf{x}}_{t} = \mathbf{x}_t - B_t, \quad B_{t+1} = B_{t}+\tilde{\mathbf{v}}_t-\mathbf{v}_t. \nonumber
\end{eqnarray}

Consequently, the update satisfies
\sr $\tilde{\mathbf{x}}_{t+1} = \mathbf{x}_{t+1} - B_{t+1} = \tilde{\mathbf{x}}_{t} - \mathbf{v}_t -\tilde{\mathbf{v}}_t + \mathbf{v}_t = \tilde{\mathbf{x}}_{t} - \tilde{\mathbf{v}}_t$\er. For \algoname, the actual update $\mathbf{v}_t$ and the cumulative compression noise $B_t$ follows:
\begin{align}
\label{NVS1}
\mathbf{v}_t &= \frac{1}{n}\sum_{i=1}^n \mathbf{C}_\delta(\gamma \mathbf{g}_{t-\tau}^i + \mathbf{e}_{t-\tau}^i), &
\sr B_t &= \frac{1}{n} \sum_{i=1}^n \mathbf{e}_{t-\tau}^i.\er
\end{align}

\noindent \textbf{Step 2: Decoupling D-SGD from D-SGD with delayed aggregation.}  Next, we simplify the sequence $\tilde{\mathbf{x}}_{t}$ (\textit{e.g.}, D-SGD with delayed aggregation) into D-SGD sequence $\hat{\mathbf{x}}_t$. We define the second noise term as $\tilde{B}_{t}$, having $\tilde{\mathbf{x}}_{t+1} = \tilde{\mathbf{x}}_{t} - \tilde{\mathbf{v}}_t$, $
   \tilde{B}_{t+1} = \tilde{B}_{t}+ \frac{\gamma}{n} \sum_{i=1}^n \mathbf{g}_t^i -\tilde{\mathbf{v}}_t$ and $
   \hat{\mathbf{x}}_{t} = \tilde{\mathbf{x}}_{t}-\tilde{B}_{t}$.
   It follows that $\hat{\mathbf{x}}_{t+1} = \hat{\mathbf{x}}_{t} - \frac{\gamma}{n} \sum_{i=1}^n \mathbf{g}_t^i$ and we have:
\begin{align}
\label{NVS2}
\tilde{\mathbf{v}}_t &= \frac{\gamma}{n} \sum_{i=1}^n \mathbf{g}_{t-\tau}^i, &
\tilde{B}_t &= \frac{\gamma}{n}\sum_{i=1}^n\sum_{j=1}^\tau  \mathbf{g}_{t-j}^i.
\end{align}
\noindent \textbf{Boundary conditions.}
For notational consistency, we define $\mathbf{g}_{t-\tau}^i = \mathbf{e}_{t-\tau}^i = \mathbf{0}^d$, for all $i \in [n]$, whenever \sr$t-\tau < 0$\er. 
This means that when \sr$t < \tau$\er, we have 
$\mathbf{v}_t = \tilde{\mathbf{v}}_t = B_t = \mathbf{0}^d, \quad  \quad 
\tilde{B}_t = \frac{\gamma}{n}\sum_{i=1}^n\sum_{j=1}^{t} \mathbf{g}_{t-j}^i$. Since $\tau$ is typically negligible compared to the total iterations $\mathbf{T}$ ($\tau \ll \mathbf{T}$), this initial phase has minimal impact on the overall convergence analysis and is omitted from the main discussion.

\subsection{Convergence Rate of \algoname}
\begin{table*}[t]
\centering
\scriptsize
\setlength{\tabcolsep}{4pt}
\renewcommand{\arraystretch}{1.25}
\caption{
Summary of the convergence rates for different algorithms under non-convex and  $\mu$-strongly convex cases. We define $\phi=\frac{1-\delta}{\delta\,(1-\tfrac{\delta}{2})^{\tau}}$, $F_0 \geq f(\mathbf{x}_0)-f^*$.
}
\resizebox{\textwidth}{!}{
\begin{tabular}{l@{\hskip 5pt}p{0.36\textwidth}@{\hskip 5pt}p{0.36\textwidth}}
\toprule
\textbf{Algorithm}
& \textbf{Non-convex}
& $\mu$-\textbf{strongly convex}\\
\midrule
\algoname
& 
$\mathcal{O}\!\left(
\frac{\sigma^2}{n\,\epsilon^2}
+\frac{\sqrt{\;\tfrac{\phi\,\zeta^2}{\delta} + \left(\phi+\tfrac{\tau}{n}\right)\sigma^2\;}}{\epsilon^{3/2}}
+\frac{\tau+Z\sqrt{\phi/\delta}}{\epsilon}
\right) \cdot LF_0$
& $\mathcal{O}\!\left(
\frac{\sigma^2}{n\,\mu\,\epsilon}
+\frac{\sqrt{\,L\!\left(\tfrac{\phi\,\zeta^2}{\delta}+\left(\phi+\tfrac{\tau}{n}\right)\sigma^2\right)\,}}{\mu\,\epsilon^{1/2}}
+\frac{L(\tau+Z\sqrt{\phi/\delta})}{\mu}
\right)$
\\[4pt]

\makecell[l]{D\text{-}SGD with\\ delayed aggregation}
& $ \mathcal{O}\!\left(
\frac{\sigma^2}{n\,\epsilon^2}
+\frac{\sqrt{\;\tfrac{\tau}{n}\,\sigma^2\;}}{\epsilon^{3/2}}
+\frac{\tau}{\epsilon}
\right) \cdot LF_0 $
& $ \mathcal{O}\!\left(
\frac{\sigma^2}{n\,\mu\,\epsilon}
+\frac{\sqrt{\,L\,\tfrac{\tau}{n}\,\sigma^2\,}}{\mu\,\epsilon^{1/2}}
+\frac{L\tau}{\mu}
\right)$
\\[4pt]

\makecell[l]{D\text{-}SGD with\\ gradient compression}
& $\mathcal{O}\!\left(
\frac{\sigma^2}{n\,\epsilon^2}
+\frac{\sqrt{\;\tfrac{\zeta^2}{\delta}+\sigma^2\;}}{\sqrt{\delta}\,\epsilon^{3/2}}
+\frac{Z}{\delta\,\epsilon}
\right) \cdot LF_0$
& $ \mathcal{O}\!\left(
\frac{\sigma^2}{n\,\mu\,\epsilon}
+\frac{\sqrt{\,L\!\left(\tfrac{\zeta^2}{\delta}+\sigma^2\right)\,}}{\sqrt{\delta}\,\mu\,\epsilon^{1/2}}
+\frac{LZ}{\mu\,\delta}
    \right)$
    \\ \midrule
\multicolumn{3}{p{\textwidth}}{\textit{Note:} The convergence analysis of D-SGD with delayed aggregation is shown in Appendix B.4 of \cite{deco_extended}. The convergence rate of D-SGD with gradient compression is shown in Table 1 of \cite{stich2020communication}}\\
\bottomrule
\end{tabular}}

\label{tab:degradation}
\end{table*}

\begin{theorem} [Non-convex convergence rate of \algoname]
Let $f: \mathbb{R}^d \rightarrow \mathbb{R}$ be $L$-smooth. There exists a stepsize $\gamma \leq \min \{ \frac{1}{4L\tau}, \frac{1}{4LZ\sqrt{\phi / \delta}}\}$, where  $\phi=\frac{1-\delta}{\delta (1-\frac{\delta}{2})^{\tau}}$, such that at most
\label{theorem:1}
\begin{equation*}
\begin{split}
\mathcal{O}\Bigg(
   &\frac{\sigma^2}{n\epsilon^2}
   + \frac{ \sqrt{\tfrac{\phi\zeta^2}{\delta} + \left(\phi+\tfrac{\tau}{n}\right)\sigma^2}}{\epsilon^{3/2}}
+ \frac{\tau}{\epsilon}
   + \frac{Z\sqrt{\phi}}{\sqrt{\delta}\,\epsilon}
\Bigg)    \\
   &\cdot L \,(f(\mathbf{x}_0) - f^*)
\end{split}
\end{equation*}
iterations of \algoname, it holds $\mathbb{E}\lVert \nabla f(\mathbf{x}_{out}) \rVert^2 \leq \epsilon $, and $\mathbf{x}_{out} = \mathbf{x}_t$ denotes an iterate $\mathbf{x}_t \in \left\{\mathbf{x}_0, \ldots, \mathbf{x}_{\mathbf{T}-1}\right\}$, where $\mathbf{T}$ denotes the total number of iterations, chosen at random uniformly.
\end{theorem}

\begin{remark}[$\phi$ determining the convergence rate in the non-degradation condition]\label{remark phi}
    In typical distributed training scenarios, the number of workers $n$ is large and  $\frac{\tau}{n} \leq 1$. As the training scale increases, $\tau$ tends to decrease due to the larger model requiring larger computation time, and $n$ tends to increase, resulting in an even smaller $\frac{\tau}{n}$. While to reduce the communication traffic, $\delta$ tends to be $10\%$ or less, common in training with gradient compression \cite{dgc, grace}, making the magnitude of $ \phi$ significantly larger than that of $\frac{\tau}{n}$. Then the convergence rate can be expressed as $\mathcal{O}(\frac{\sigma^2}{n\epsilon^2}+\frac{ \sqrt{\phi(\frac{\zeta^2}{\delta} +\sigma^2)}}{\epsilon^{3/2}}+\frac{Z\sqrt{\frac{\phi}{\delta}}+\tau}{\epsilon})$. Similar to previous work \cite{dagctmc}, the first term $\frac{\sigma^2}{n\epsilon^2}$ is independent of $\delta$ and $\tau$. The second term $\frac{ \sqrt{\phi(\frac{\zeta^2}{\delta} +\sigma^2)}}{\epsilon^{3/2}}$ has a greater impact on the overall convergence rate compared to the third term.
    
    Let us focus on $\frac{ \sqrt{\phi(\frac{\zeta^2}{\delta} +\sigma^2)}}{\epsilon^{3/2}}$, in LLM training scenarios (like pre-training or full-parameter fine-tuning), local datasets are typically allocated centrally, resulting in IID datasets ($\zeta\approx0$ ). Moreover, due to the small batch sizes commonly used in LLM training (limited by the constraint of GPU memory), $\sigma$ tends to be large. Under such circumstances, $\frac{\zeta^2}{\delta}\ll \sigma^2$ and $\phi$ captures the convergence rate. While in non-IID scenarios, (large $\zeta$) or training small models (small $\sigma^2$), the key factor is $\phi'=\frac{1-\delta}{\delta^2(1-\frac{\delta}{2})^{\tau}}$. In this work, we focus on the first scenario where $\phi$ is the key factor that determines the convergence of \algoname. For the analysis in non-IID scenarios (like Federated Learning), we use $\phi'$ as the key factor.

    For convenience, we define $\phi(\delta,\tau)=\frac{1-\delta}{\delta}(1-\delta/2)^{-\tau}$. $\phi(\delta,\tau)$ can be decomposed into a compression-only term $\frac{1-\delta}{\delta}$ and a delay-induced amplification factor $(1-\delta/2)^{-\tau}$ that grows with $\tau$. For instance, at $\delta=10\%$, increasing $\tau$ from $0$ to $20$/$40$ amplifies the compression-only term by $2.8\times$/$7.8\times$, showing that delay staleness further magnifies compression error. Therefore, the effects of $\delta$ and $\tau$ on the convergence rate are not independent. A heatmap of the delay-induced amplification factor over a wider $(\delta,\tau)$ range is provided in Appendix~\ref{appendix:phi-vis}.
\end{remark}

\begin{remark}[Special Cases]
    \algoname\ has two special cases, namely, without compression ($\delta =1$) and without delayed aggregation ($\tau =0$). The convergence rates for \algoname \ and its special cases are shown in Table~\ref{tab:degradation}.
    When $\delta =1$, $\phi = 0$ and \algoname \ reduces into D-SGD with delayed aggregation. Then the convergence rate is equal to $\mathcal{O}(\frac{\sigma^2}{n\epsilon^2}+\frac{ \sqrt{\frac{\tau}{n}\sigma^2}}{\epsilon^{3/2}} +\frac{\tau}{\epsilon})$, consistent with the convergence rate of D-SGD with delayed aggregation \cite{stich2022sharper, stich2019error}. When $\tau =0$, \algoname \ reduces to D-SGD with gradient compression, and $\phi = \frac{1-\delta}{\delta}$. Then the convergence rate becomes $\mathcal{O}(\frac{\sigma^2}{n\epsilon^2}+\frac{\zeta/\sqrt{\delta} + \sigma}{\sqrt{\delta}\epsilon^{3/2}} +\frac{Z}{\delta\epsilon})$, consistent with the convergence rate of D-SGD with gradient compression \cite{stich2020communication}.
\end{remark}

\begin{theorem} [Convex convergence rate of \algoname]
Let $f: \mathbb{R}^d \rightarrow \mathbb{R}$ be $L$-smooth and $\mu$-convex. Then there exists a stepsize $\gamma \leq \min \{ \sr\frac{1}{8\sqrt 2 L\tau}\er, \frac{1}{16LZ\sqrt{2\phi/\delta}}\}$, such that at most
\label{theorem:2}
\begin{equation*}
\begin{split}
    \mathcal{O}(&\frac{\sigma^2}{n\mu\epsilon}+\frac{ \sqrt{L(\frac{\phi\zeta^2}{\delta} + (\phi+\frac{\tau}{n})\sigma^2)}}{\mu\epsilon^{1/2}}+\frac{L\tau}{\mu}+\frac{LZ\sqrt{\phi/\delta}}{\mu})
\end{split}
\end{equation*}
iterations of \algoname, it holds $\mathbb{E}f(\mathbf{x}_{out})-f^* \leq \epsilon $, and $\mathbf{x}_{out} = \mathbf{x}_t$ denotes an iterate $\mathbf{x}_t \in \left\{\mathbf{x}_0, \ldots, \mathbf{x}_{\mathbf{T}-1}\right\}$, selected probabilistically based on $(1-\frac{\mu \gamma}{2})^{-t}$. 
\end{theorem}

\begin{remark}
Analogous to remarks in \textbf{Theorem~\ref{theorem:1}}, in the special cases, the convex convergence rate is the same as that of D-SGD with delayed aggregation (under $\delta=1$) and D-SGD with gradient compression (under $\tau=0$). In the non-degradation conditions, the convergence rate can be expressed as $\mathcal{O}(\frac{\sigma^2}{n\mu\epsilon }+\frac{ \sqrt{\phi(\frac{\zeta^2}{\delta} +\sigma^2)}}{\mu\epsilon^{1/2}}+\frac{L(\tau+Z\sqrt{\phi/\delta})}{\mu})$, $\phi$ dominates the convergence rate of \algoname \ in the convex case, too.
\end{remark}

\section{Mathematical Modeling Time to Accuracy in \algoname}
\label{section4}

We denote the bandwidth (bits/s) as a time-varying function $a(s)$, where $s$ represents continuous time, the average bandwidth over the previous $E$ iterations as $\bar{a}$, end-to-end latency (s) as $b$, and gradient size (bits) as $V_g$. We denote the computation time per iteration (s) and the average end-to-end training time per iteration (s) as $T_{\text{comp}}$ and $T_{\text{avg}}$, respectively. We use $\lceil \cdot \rceil$ \sr and $\lfloor \cdot \rfloor$\er\ to represent the ceiling \sr and floor\er\ functions. Detailed proofs are provided in Appendix B.5 of the extended version~\cite{deco_extended}.

\subsection{Iteration Timing Model of \algoname}
\label{subsec:timing_model}
Consider \algoname \ in a single-worker and a server setting. Each iteration consists of three stages: local computation $T_\text{comp}$; gradient transmission time $t_{\delta}$, where $\int_0^{t_{\delta}} a(s) \,\mathrm{d}s{=}\delta V_g$ and fixed end-to-end latency $b$. 

Let $t$ denote the iteration index ($t = 0, 1, 2, \dots$). Following the standard steady-state pipelined timing abstraction used in delayed aggregation \cite{DGA} and distributed pipeline training systems \cite{gpipe,pipedream}, the $(t{+}1)$-th computation starts after both the $t$-th computation and the $(t{-}\tau)$-th update are completed. \ Let $TS_t$ and $TC_t$ be the completion times of the $t$-th computation and communication, respectively. This leads to the following recursive relation:
\begin{equation}
TS_{t+1} = T_{\text{comp}} + \max\{\,TS_t,\; TC_{t-\tau}\,\}.
\end{equation}

\subsection{Modeling Assumptions}
To facilitate the time-to-iteration model in \algoname\, we introduce the following assumptions.

\noindent \textbf{Assumption 5} (Bottleneck equivalence). 
In synchronous parameter-server systems, the overall iteration time is dominated by the slowest link. This allows us to analyze the multi-worker system by focusing on a single worker-server link representing the bottleneck, without loss of generality for the analysis.

\noindent \textbf{Assumption 6} (Historical average bandwidth estimation).
Network measurement studies \cite{strauss2003measurement, li2004first} indicate that \sr network characteristics such as available bandwidth can be estimated and predicted with reasonable accuracy\er\ \cite{ng2002predicting}\sr; we therefore model the bandwidth as locally stable over short timescales.\er\ We define the window for recording bandwidth as $E$ iterations. Consequently, we approximate $a(s)=\bar{a}$ for the subsequent $E$ iterations, updating the bandwidth estimate every $E$ iterations. This implies the gradient transmission time is $t_{\delta}{=}\frac{\delta V_g}{\bar{a}}$.

\subsection{Time to Iteration in \algoname}\label{time2iteration}

\begin{theorem}[Estimation of $T_{\text{avg}}$ in \algoname]
\label{theorem:3}
Under \textbf{Assumptions 5} and  \textbf{6},
the average iteration time $T_{\text{avg}}$ satisfies the following approximation:           
\begin{equation}
\label{eq:T_avg}
T_{\text{avg}}\approx\max{\left \{  \frac{T_{\text{comp}}+b+\delta \cdot V_g/\bar{a}}{\tau+1} ,\frac{\delta \cdot V_g}{\bar{a}},T_{\text{comp}} \right \}},
\end{equation}
where the approximation error arises from finite-window pipeline boundary effects, which are amortized over the measurement window and become negligible when $E \gg \tau$.
\end{theorem}

\begin{remark}[The locally optimal compression ratio $\delta^*(\tau)$]\label{remark delta}
Our goal is to minimize the iteration time $T_{\text{avg}}$ without unnecessarily sacrificing model accuracy. According to Eq.~\eqref{eq:T_avg}, the lower bound of the average iteration time is the computation time $T_{\text{comp}}$. 
There exists a locally optimal compression ratio given a fixed $\tau$, denoted as $\delta^*(\tau)$, such that $T_{\text{avg}}(\tau, \delta^*(\tau)) = T_{\text{comp}}$. Choosing any $\delta\leq \delta^*$ yields no gain in $T_\text{avg}$, but degrades the model performance due to more information loss. Thus, for a $\tau$, $\delta^*(\tau)=\min\left\{\max\left(0, (\tau T_{\text{comp}} - b)\frac{\bar{a}}{V_g}\right),T_\text{comp}\frac{\bar{a}}{V_g},1\right\}$.
\end{remark}

\subsection{DeCo: Joint Optimization for Delay Staleness and Gradient Compression in DML}

\begin{figure*}[t]
    \centering
    \includegraphics[width=1\textwidth]{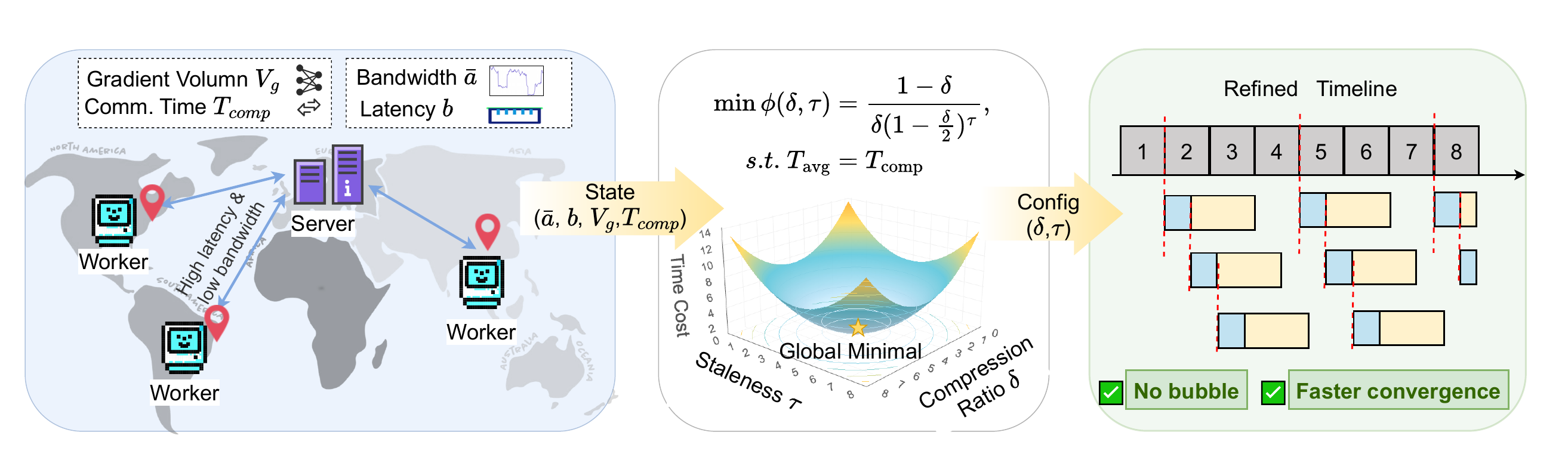}
    \caption{The high level design of \sysname. \sysname \ adaptively adjusts the compression ratio $\delta$ and delay staleness $\tau$ based on communication and computation conditions.}
    \label{DeCoFrame}
\end{figure*}

\begin{algorithm}[t]
\SetAlgoLined
\caption{\sysname}
\label{DeCoSGD}
\KwIn{$n$, $\gamma$,  $\mathbf{T}$, $\textbf{C}(\cdot)$, $V_g$, $T_{\text{comp}}$, $E$}
\KwOut{$\textbf{x}_\mathbf{T}$}
Initialize $\mathbf{x}_0$, $\mathbf{e}_{0}^i = \mathbf{0}_d$\;
Initialize $\tau$ and $\delta$ based on DeCo algorithm\;
\For{$t\in[\mathbf{T}]$}{
    \tcc{Worker side}
    \For{$i\in[n]$}{ 
        Worker $i$ waits until it receives $\textbf{x}_{t}$\;
        Calculate $\mathbf{g}_t^i$ based on the global model $\mathbf{x}_t$\;
        $\Delta_t^i = \textbf{C}_{\delta} (\textbf{e}_t^i + \gamma\textbf{g}_t^i)$\;
        $\textbf{e}_{t+1}^i = \textbf{e}_{t}^i + \gamma \textbf{g}_{t}^i - \Delta_t^i$\;
        Send $\Delta_t^i$ to the server\;
    }
    \tcc{Server side}
    Server waits until it receives $\Delta_{t-\tau}^i$ for all $i\in[n]$\;
    $\textbf{x}_{t+1} = \textbf{x}_t - \frac{1}{n}\sum_{i=1}^n \Delta_{t-\tau}^i$\;
    Broadcast $\textbf{x}_{t+1}$\;
    \tcc{Update $\tau$ and $\delta$ with DeCo}
    \If{$(t+1)\mod E==0$}{
    Get $\bar{a}$, $b$ from the network\;
    $\tau$, $\delta$ = DeCo($V_g$, $\bar{a}$, $b$, $T_{\text{comp}}$)\;
    Broadcast new $\tau$ and $\delta$ to workers\; 
    }  
}
\textbf{Return} $\textbf{x}_\mathbf{T}$\;
\end{algorithm}

\hfill

\begin{algorithm}[t]
\SetAlgoNoLine
\caption{DeCo}
\label{DeCo}
\KwIn{ $V_g$, $\bar{a}$, $b$, $T_{\text{comp}}$ }
\KwOut{Optimal $\tau$, $\delta$ with minimal $\phi$}

$\phi_{min} = \infty$ 

\For{$\tau \in \{ \lceil \frac{b+V_g/\bar{a}}{T_{\text{comp}}} \rceil ,\cdots,\sr\lfloor \frac{b}{T_{\text{comp}}} \rfloor + 1\er \}$ }{
    $\delta = \min\left\{(\tau T_{\text{comp}}-b)\frac{\bar{a}}{V_g},T_{\text{comp}}\frac{\bar{a}}{V_g},1\right\}$\;
    
    \tcc{Calculate the value of $\phi$}
    $\phi = \frac{1 - \delta}{\delta \cdot (1 - \frac{\delta}{2})^{\tau}}$\;
    
    \If{$\phi \le \phi_{min}$}{
        $\phi_{min} = \phi$\;
        $\tau^*, \delta^* = \tau, \delta$\;
    }
}
\textbf{Return} $\tau^*$, $\delta^*$\;
\end{algorithm}

Our design goal is to minimize the end-to-end time to achieve a specific accuracy, which is equivalent to minimizing the objective function $\phi$. Since $\tau$ is discrete, the minimum of $\phi$ cannot be directly solved by the partial derivatives. We use the critical condition $T_{\text{avg}} = T_{comp}$ (\textit{i.e.}, there are no redundant bubbles in the pipeline) to reduce the search space to a discrete set of candidate points. The mathematical description of this goal is 
\begin{equation}
\label{eq:ori_optimal_problem}
\begin{aligned}
    \min\ \phi(\delta,\tau)=\frac{1-\delta}{\delta(1-\frac{\delta}{2})^\tau},
    \textit{s.t.}\ T_{\text{avg}}{=}T_\text{comp} ,\tau \in \mathbb{N}, \delta \in \left ( 0,1 \right ].
\end{aligned}
\end{equation}

Combined with \textbf{Remark~\ref{remark delta}}, we can rewrite Eq.~\ref{eq:ori_optimal_problem} as:
\begin{equation}
\label{eq:equa_optimal_problem}
\begin{aligned}
    \min \ \phi(\delta^*(\tau),\tau)=\frac{1-\delta^*(\tau)}{\delta^*(\tau)(1-\frac{\delta^*(\tau)}{2})^\tau}\\
    \textit{s.t.} \ \tau\in\left[ \sr\lfloor \frac{b}{T_{comp}} \rfloor + 1\er,\cdots, \lceil \frac{b+\frac{V_g}{\bar{a}}}{T_{comp}} \rceil \right].
\end{aligned}
\end{equation}

We traverse the range of $\tau$ to obtain $\phi^*$ to achieve the fastest convergence rate and choose the smallest $\tau$ as $\tau^*$ . The procedure for obtaining $\tau^*$ and $\delta^*$ is referred to as the DeCo algorithm (shown in Algo.~\ref{DeCo}).

\subsection{Efficient Implementation of DeCo}
\label{subsec:implementation}

\sysname \ integrates DeCo into D-SGD (pseudo-code shown in Algo.~\ref{DeCoSGD}, design in Fig.~\ref{DeCoFrame}). We use a \textit{pre-computed Look-up Table (LUT) strategy to ensure $\mathcal{O}(1)$ runtime complexity for decision-making.} The implementation consists of two operational phases:
\subsubsection{Pre-computation}
To build the LUT, \sysname\ profiles the network during a warm-up phase to estimate the bandwidth range $[a_{min}, a_{max}]$. This range is then linearly interpolated into several sample points (\textit{e.g.}, $20$ points). With measured $b$, the server runs DeCo offline for each sample point to compute and store the optimal configuration.
\subsubsection{Runtime Lookup}
During training, the worker updates the average bandwidth $\bar{a}$ every $E$ iterations. The system then queries the LUT for the optimal configuration. If $\bar{a}$ is within the recorded range, the system retrieves $(\delta^*, \tau^*)$ by rounding $\bar{a}$ to the nearest discretized bandwidth in the LUT. This $O(1)$ lookup ensures minimal runtime cost. If $\bar{a}$ falls outside the recorded range $[a_{\min}, a_{\max}]$, \sysname\ falls back to online computation in the server to ensure stability and correctness.
The fallback recomputes $(\delta^*,\tau^*)$ using the latest measured $\bar{a}$, which prevents stale LUT entries under bandwidth drift. 

\section{Evaluation Experiments}
\label{section5}

\subsection{Experimental Environment}
The experiments are conducted on a distributed cluster comprising $32$ nodes. Each node is equipped with an Ubuntu 24.04 LTS system, an Intel Xeon Gold $6230$ CPU, and a single Nvidia A$40$ GPU with $48$GB memory. We utilize Python $3.10.16$, employing PyTorch $2.5.1$ with CUDA $12.4$ as the ML toolkit and Gloo as the communication backend. 

\subsection{Experimental Settings}

\noindent \textbf{Experiment tasks:}
To assess the efficacy of our proposed design, we evaluate three representative tasks:  CIFAR-$10$ with ResNet-34 \cite{resnet} (denoted as ResNet@CIFAR-$10$),
ImageNet \cite{imagenet} using ViT \cite{vit} (denoted as ViT@ImageNet) and Wikitext \cite{wikitext} on GPT \cite{gpt2} (denoted as GPT@Wikitext).
The ResNet architecture is ResNet-34, used widely in previous works \cite{dagctmc, stc}. 
The ViT base model comprises approximately $86$ million parameters, while the smallest GPT variant contains $124$ million parameters. For ResNet@CIFAR-$10$, the learning rate is $0.01$, and the batch size per node is $32$. 
For ViT@ImageNet and GPT@Wikitext, the learning rates are $0.01$ and $0.1$, respectively, with batch sizes of $40$ and $5$.
Regarding initialization, ResNet is trained from scratch, whereas ViT and GPT are initialized with pre-trained weights.

\noindent \textbf{Baselines:} 
We compare D-SGD, Accordion, DAGC, DC2, DGA, and CocktailSGD as baselines. D-SGD is the case $\tau=0, \delta=1$. DAGC \cite{dagctmc}
, Accordion \cite{accordion}, DC2 \cite{dc2} are related adaptive gradient compression algorithms. DAGC is the SOTA compressor in communication-constrained non-IID scenarios, assigning different compression ratios to workers with different local dataset volumes. The average compression ratio $\bar{\delta}$ of DAGC is set to match the initial $\delta$ of \sysname.
Accordion is the SOTA adaptive gradient compression algorithm, which dynamically switches between a high compression ratio (small $\delta$) for non-critical iterations and a low ratio (large $\delta$) otherwise. The large $\delta$ (as well as small $\delta$) is equal to the $\delta$ in \sysname \ under the constraint bandwidth (the half of the constraint bandwidth).
DC2 adapts $\delta$ based on the network condition. We use DA2 in DC2 and $k_0$ is equal to the initial $\delta$ of \sysname. The compressor
we use in all baselines is Top-$k$, which transmits the largest $k$ elements of gradients. We select Top-$k$ for its empirical robustness and widespread adoption in DML. DGA introduces delayed aggregation in FL to reduce the negative effect of the high latency. We configure DGA with $K{=}1$ (disabling infrequent communication) and $D{=}\tau$, equal to the initial $\tau$ of \sysname. CocktailSGD \cite{wang2023cocktailsgd} uses the fixed $\tau{=}1$ and the compressor composed of Random-$k$ with $k{=}20\%$, Top-$k$ with $k{=}10\%$ and using $4$-bit quantization, as stated in the main body. 

\begin{figure}[t]
    \centering
    \includegraphics[width=0.47\textwidth]{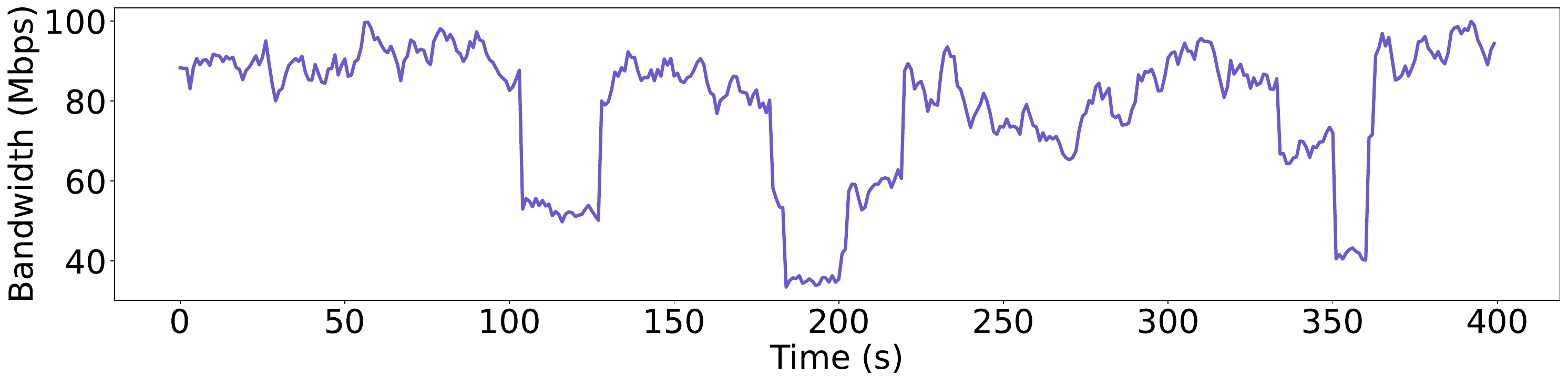}
    \caption{The network bandwidth over time with a constraint of $100$Mbps.}
    \label{fig:network100}
\end{figure}
\begin{figure}[t]
    \centering
    \includegraphics[width=1\linewidth]{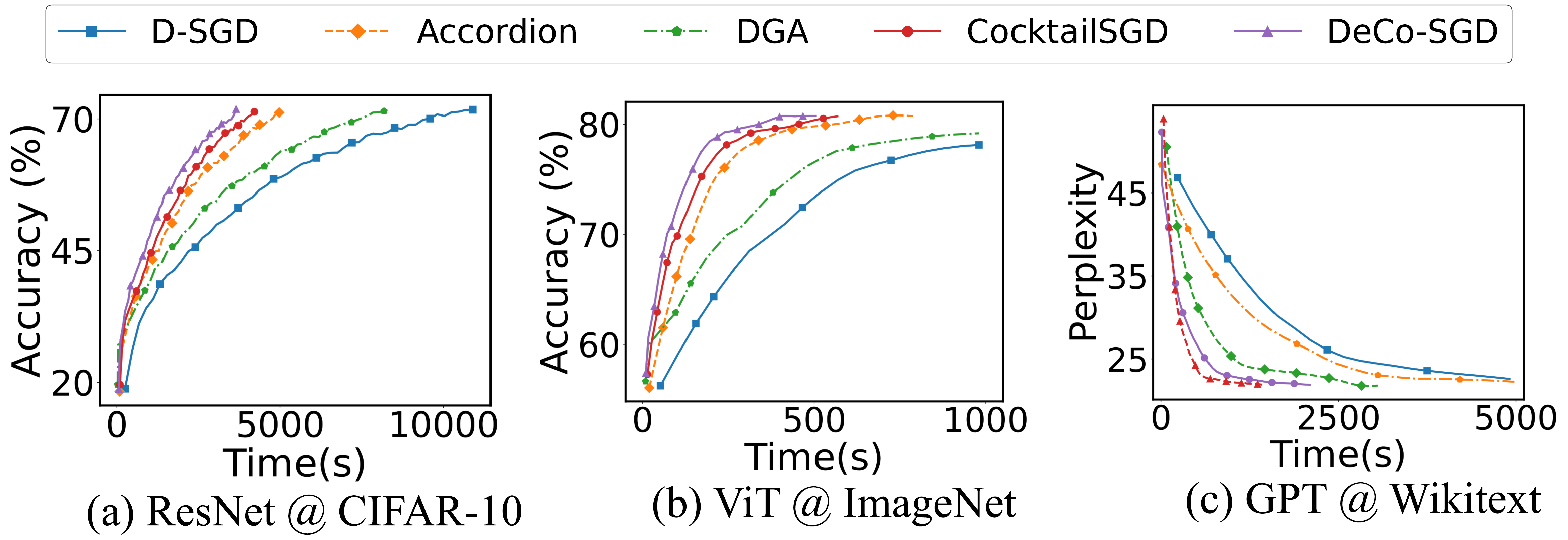}
    \caption{Training time comparison across different tasks.}
    \label{fig:1x4-models}
\end{figure}

\noindent \textbf{Network environment:} 
We simulate challenging network environments (static high latency $\ge$ $100$ms, dynamic low bandwidth,  constrained below $1$Gbps). We measure the performance of our design under multiple network conditions in Section~\ref{exp.4}. An example of
 dynamic bandwidth over time is shown in Fig.~\ref{fig:network100}.

\subsection{Comparison of Training Speed}\label{exp.2}
As illustrated in Fig.~\ref{fig:1x4-models}, \sysname{} demonstrates significant improvements in training speed compared to D-SGD, Accordion, DGA, and CocktailSGD in all tasks. All experiments were conducted with $4$ worker nodes, under a fixed latency of $200$ms, with bandwidth fluctuating around $100$Mbps.
Taking ViT@ImageNet as an example, when accuracy reaches 70\%, \sysname{} achieves a $4.99 \times$ speedup\footnote{Speed-up is defined as the ratio of the baseline method’s training time to that of \sysname.} over D-SGD and a $1.34 \times$ speedup over CocktailSGD. At 80\% accuracy, the speedup increases to $5.01 \times$ and $1.37 \times$, respectively. Compared with other baseline methods, the advantages of \sysname{} are pronounced, mainly for two reasons: (1) their parameter settings are suboptimal; (2) they cannot adapt to the dynamic network environment in real time. In contrast, \sysname{} utilizes theoretical guidance and runtime feedback to jointly optimize compression and latency, thereby achieving a more effective computation-communication trade-off throughout the training process.
To quantify the computation--communication trade-off, we provide profiling results under the same setting in Appendix~\ref{app:profiling-overhead}. The results show that the average DeCo decision time is below $10^{-6}$ s, and in cross-region distributed training, the communication benefit brought by compression is much larger than the cost of compression itself.

\subsection{Scalability Verification of \sysname}\label{exp.3}

\begin{figure*}[t]
    \centering
    \includegraphics[width=0.98\textwidth]{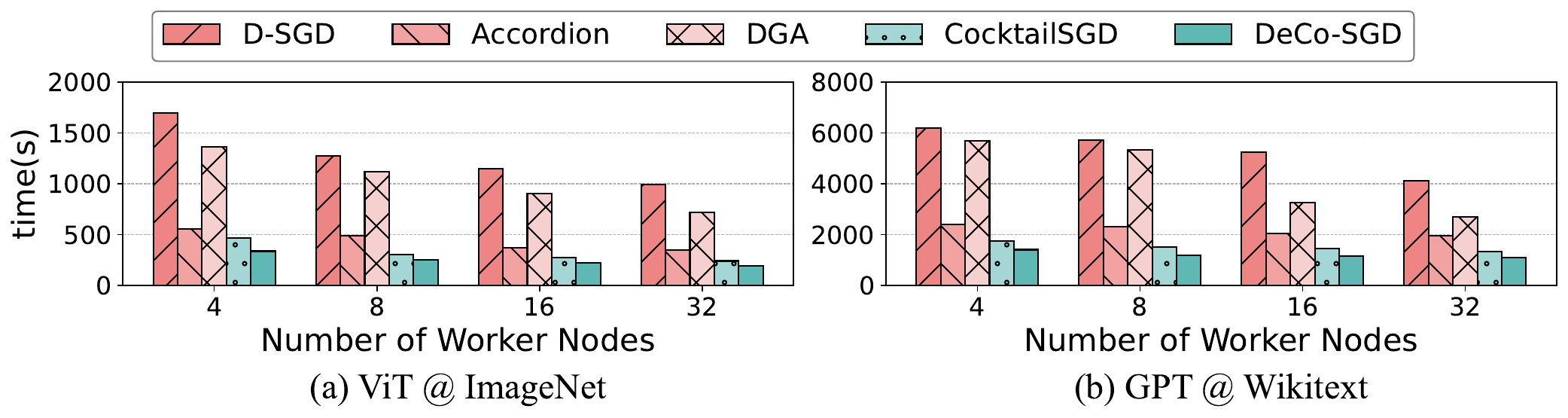}
    \caption{Training time comparison of different methods when the worker node scales. \sysname{} demonstrates superior scalability and efficiency compared to other methods across both settings.}
    \label{fig:bar}
\end{figure*}

\sysname \ is inherently scalable, as the computational complexity of \sysname{} in finding $\delta^*$ and $\tau^*$ is theoretically independent of the number of worker nodes. Fig.~\ref{fig:bar} illustrates the performance of \sysname \ across different numbers of worker nodes, scaling from $n{=}4$ to $n{=}32$. The results demonstrate that \sysname \ consistently maintains performance improvements as the number of nodes increases. The complete training curves are shown in Appendix C.2 of the extended version~\cite{deco_extended}. Detailed $32$-worker results under different network conditions are reported in Appendix~\ref{app:scalability32}. In these $32$-worker results, \sysname{} achieves up to $3.87\times$ and $1.22\times$ speed-ups over D-SGD and CocktailSGD, respectively, for GPT@Wikitext, and up to $5.08\times$ and $1.24\times$ speed-ups for ViT@ImageNet.

\subsection{\sysname \ under Different Bandwidths and Latency} \label{exp.4}

Table~\ref{tab:iid} presents the training times of different methods on GPT@Wikitext and ViT@ImageNet under varying network conditions, 
with all experiments conducted using $4$ worker nodes, using D-SGD, Accordion, DC2, DGA, CocktailSGD as baselines. Due to space limitations, we report the training time to reach a representative target metric for each task:  a perplexity of $22$ for GPT and $80$\% accuracy for ViT. Detailed convergence curves covering the entire training process are provided in Appendix C.3 of the extended version~\cite{deco_extended}, which verify that \sysname \ consistently outperforms other methods, not just the specific points reported here. The advantage increases further as bandwidth decreases and network conditions worsen (bandwidth drops, latency increases).
The results clearly demonstrate the robustness of our design against bandwidth and latency variations. The advantage of our design over baselines increases as network conditions worsen. We
achieve up to $5.02\times$ and $1.30\times$ speed-ups over D-SGD and CocktailSGD, respectively. Notably, in dynamic network environments, DC2 performs better than Accordion, but fails to mitigate the problem of high latency. This shows that gradient compression alone is insufficient to solve the problem.

\begin{table*}[!t]
\centering
\caption{Comparison of training time (s) under different methods and network conditions. $a$ is maximum bandwidth and $b$ is the end-to-end latency. The numbers in parentheses represent the speedup compared to \sysname.}
\label{tab:iid}
\scalebox{0.77}{
\begin{tabular}{@{}lcccccccc@{}}
\toprule
\makecell{\textbf{Model} \\ \textbf{@DataSet}} & \makecell{\textbf{a(Gbps)}, \\ \textbf{b(s)}} &  \textbf{D-SGD}  & \textbf{Accordion} & \textbf{DC2} & \textbf{DGA} & \textbf{CocktailSGD} & \textbf{\sysname} \\
\midrule
\multirow{4}{*}{\makecell{GPT\\@Wikitext}}
  & $0.1, 0.1$  &  $6396.95 \pm 32.36$ ($4.90\times$)  & $1955.28 \pm 13.59$ ($1.50\times$) & $1901.13 \pm 11.45$ (\sr$1.46\times$\er) & $5506.78 \pm 26.77$ ($4.22\times$) & $1535.21 \pm 12.91$ ($1.18\times$) & $\mathbf{1306.29 \pm 10.29}$ \\
  & $0.5, 0.1$  &  $1571.37 \pm 11.13$ ($2.44\times$) & $829.06 \pm 6.65$ ($1.29\times$) & $801.72 \pm 6.06$ (\sr$1.25\times$\er) & $1346.34 \pm 10.75$ ($2.09\times$) & $712.93 \pm 5.38$ ($1.11\times$) & $\mathbf{642.79 \pm 5.74}$ \\
  & $0.1, 1.0$ &  $7232.88 \pm 35.62$ ($5.02\times$)  & $2510.26 \pm 22.73$ ($1.74\times$) & $2469.74 \pm 22.98$ ($1.71\times$) & $5764.60 \pm 29.69$ ($4.00\times$) & $1683.11 \pm 14.66$ ($1.17\times$) & $\mathbf{1440.59 \pm 11.87}$ \\
  & $0.5, 1.0$ &  $2363.73 \pm 13.87$ ($2.37\times$)  & $1325.12 \pm 10.90$ ($1.33\times$) & $1321.25 \pm 10.31$ ($1.32\times$) & $1562.93 \pm 11.89$ ($1.57\times$) & $1132.63 \pm 9.91$ ($1.14\times$) & $\mathbf{997.25 \pm 7.92}$ \\
\midrule
\multirow{4}{*}{\makecell{ViT\\@ImageNet}} 
  & $0.1, 0.1$  &  $1373.84 \pm 10.58$ ($4.85\times$) & $440.14 \pm 3.28$ ($1.55\times$) & $415.69 \pm 3.14$ (\sr$1.47\times$\er) & $1183.68 \pm 7.85$ ($4.18\times$) & $364.77 \pm 1.02$ ($1.29\times$) & $\mathbf{283.06 \pm 1.80}$ \\
  & $0.5, 0.1$   &  $446.20 \pm 3.39$ ($2.64\times$)  & $274.60 \pm 2.21$ ($1.62\times$) & $253.36 \pm 2.17$ (\sr$1.50\times$\er) & $385.60 \pm 1.57$ ($2.28\times$) & $219.39 \pm 0.98$ ($1.30\times$) & $\mathbf{169.13 \pm 0.47}$ \\
  & $0.1, 1.0$  &  $1530.57 \pm 9.89$ ($3.85\times$)  & $604.67 \pm 2.79$ ($1.52\times$) & $579.03 \pm 2.53$ (\sr$1.46\times$\er) & $1287.03 \pm 10.70$ ($3.24\times$) & $505.91 \pm 2.56$ ($1.27\times$) & $\mathbf{397.16 \pm 1.39}$ \\
  & $0.5, 1.0$  &  $631.15 \pm 3.33$ ($2.67\times$)  & $373.79 \pm 1.92$ ($1.58\times$) & $368.29 \pm 1.83$ ($1.56\times$) & $444.38 \pm 2.18$ ($1.88\times$) & $278.57 \pm 1.35$ ($1.18\times$) & $\mathbf{235.95 \pm 0.88}$ \\
\bottomrule
\end{tabular}
}
\end{table*}

\subsection{\sysname \ under non-IID Scenarios} \label{exp:niid}

Federated learning or other data silo scenarios encounter non-IID problems, leading to accuracy degradation \cite{niid2020icml}. We compare \sysname \ with other algorithms in non-IID settings. To evaluate performance under these conditions, we simulate the label imbalance and the quantity imbalance using a Dirichlet distribution with a concentration parameter of $0.5$ \cite{niidbench}. Notably, for these experiments, we replace Accordion with DAGC in contrast to Section~\ref{exp.4}, as DAGC is a robust baseline specifically designed for non-IID scenarios. Following Remark \ref{remark phi}, we substitute
$\phi$ with $\phi'=\phi/\delta$.
Despite the increased training uncertainty and instability inherent in non-IID data, \sysname \ demonstrates exceptional adaptability. It achieves speed-ups ranging from \sr$1.31\times$\er\ to \sr$2.07\times$\er\ over the strong baseline DAGC, significantly outperforming the gains observed in IID settings. In non-IID scenarios, our design achieves up to $5.13\times$ and $1.36\times$ speed-ups over D-SGD and CocktailSGD. The complete training curves are shown in Appendix C.4 of the extended version~\cite{deco_extended}.

\begin{table*}[!t]
\centering
\caption{Comparison of training time (s) under different methods and network conditions in non-IID  scenarios. $a$ is the maximum bandwidth and $b$ is the end-to-end latency.}
\label{tab:non-iid}
\scalebox{0.77}{
\begin{tabular}{@{}lcccccccc@{}}
\toprule
\makecell{\textbf{Model} \\ \textbf{@DataSet}} & \makecell{\textbf{a(Gbps)}, \\ \textbf{b(s)}}  & \textbf{D-SGD} & \textbf{DAGC} & \textbf{DC2} & \textbf{DGA} & \textbf{CocktailSGD} & \textbf{\sysname} \\
\midrule
\multirow{4}{*}{\makecell{GPT\\@Wikitext}}
  & $0.1, 0.1$  & $7276.34 \pm 49.33$ ($5.13\times$) & $ 2283.53 \pm 12.55$ ($1.61\times$)& $ 2244.73 \pm 12.21$ ($1.58\times$)& $ 6108.13 \pm 35.23$ ($4.31\times$)  & $1732.25 \pm 12.58$ ($1.22\times$) & $\mathbf{1417.55 \pm 11.23}$ \\
  & $0.5, 0.1$   & $1785.64 \pm 12.55$ ($2.41\times$) & $969.11 \pm 7.12$ (\sr$1.31\times$\er)  &   $917.54 \pm 7.42$ (\sr$1.24\times$\er)  & $1515.61 \pm 12.48$ ($2.04\times$) & $815.52 \pm 6.45$ ($1.10\times$) & $\mathbf{741.35 \pm 5.73}$ \\
  & $0.1, 1.0$  & $8179.46 \pm 54.32$ ($4.76\times$) & $2841.85 \pm 20.07$ ($1.65\times$) & $2827.36 \pm 20.91$ (\sr$1.65\times$\er) & $6517.52 \pm 34.66$ ($3.79\times$) & $ 2018.44 \pm 14.65$ ($1.17\times$) & $\mathbf{1718.41 \pm 13.46}$ \\
  & $0.5, 1.0$  & $2736.48 \pm 19.64$ ($2.31\times$) & $1576.29 \pm 12.98$ (\sr$1.33\times$\er) & $1570.48 \pm 12.07$ ($1.32\times$) & $1775.52 \pm 13.29$ ($1.50\times$) & $1459.16 \pm 11.63$ ($1.23\times$) & $\mathbf{1186.33 \pm 9.27}$ \\
\midrule
\multirow{4}{*}{\makecell{ViT\\@ImageNet}} 
  & $0.1, 0.1$   & $1273.93 \pm 9.32$ ($4.64\times$) & $567.73 \pm 3.73$ (\sr$2.07\times$\er)  & $537.85 \pm 3.14$ (\sr$1.96\times$\er) & $1324.88 \pm 10.26$ ($4.82\times$) & $368.39 \pm 1.82$ ($1.34\times$) & $\mathbf{274.76 \pm 1.45}$\\
  & $0.5, 0.1$   & $434.16 \pm 2.15$ ($2.34\times$) & $364.22 \pm 1.80$ (\sr$1.96\times$\er) & $353.70 \pm 1.79$ (\sr$1.90\times$\er) & $467.97 \pm 2.10$ ($2.52\times$) & $224.09 \pm 1.38$ ($1.21\times$) & $\mathbf{185.90 \pm 0.68}$\\
  & $0.1, 1.0$   & $1724.74 \pm 10.47$ ($4.22\times$)  & $672.44 \pm 3.21$ ($1.64\times$) & $644.02 \pm 3.14$ ($1.57\times$) & $1367.39 \pm 8.25$ ($3.34\times$) & $555.26 \pm 3.11$ ($1.36\times$) & $\mathbf{409.18 \pm 2.48}$\\
  & $0.5, 1.0$   & $677.23 \pm 4.21$ ($2.78\times$) & $489.41 \pm 2.21$ (\sr$2.01\times$\er) &  $478.93 \pm 1.85$ (\sr$1.97\times$\er) & $599.93 \pm 2.83$ ($2.46\times$) & $283.89 \pm 1.62$ ($1.17\times$) & $\mathbf{243.52 \pm 1.02}$\\
\bottomrule
\end{tabular}
}
\end{table*}

\subsection{Comparison with Local-Update Baselines}\label{exp:local-update}

To further compare with WAN-oriented training strategies, we evaluate two local-update baselines, $\gamma$-FedHT~\cite{lu2025fedht} and DiLoCo~\cite{douillard2023diloco}, under the same network settings as Table~\ref{tab:iid}. $\gamma$-FedHT is a strong FedAVG-based local-update method with stepsize-aware hard-threshold compression, where workers perform several local training iterations and upload compressed accumulated updates with error feedback. DiLoCo is a representative low-communication distributed training method proposed by DeepMind, where workers perform multiple local Adam-based optimizer steps before periodic global synchronization. For both baselines, we set the communication interval to $5$; for $\gamma$-FedHT, the initial hard-threshold is chosen so that its initial compression ratio matches the initial $\delta$ of \sysname{}. For DiLoCo, we use Adam as the local optimizer with the common setting $\beta_1=0.9$, $\beta_2=0.999$, and $\epsilon=10^{-8}$. As shown in Table~\ref{tab:local-update}, \sysname{} consistently achieves shorter time-to-target than both baselines, because these local-update methods mainly reduce communication frequency but still suffer from end-to-end latency, while \sysname{} uses delayed aggregation to overlap communication with computation.

\begin{table}[!t]
\centering
\caption{Comparison with local-update baselines. $a$ is the maximum bandwidth and $b$ is the end-to-end latency.}
\label{tab:local-update}
\resizebox{\columnwidth}{!}{
\begin{tabular}{@{}lcccc@{}}
\toprule
\makecell{\textbf{Model} \\ \textbf{@DataSet}} & \makecell{\textbf{a(Gbps)}, \\ \textbf{b(s)}} & \textbf{$\gamma$-FedHT} & \textbf{DiLoCo} & \textbf{\sysname} \\
\midrule
\multirow{4}{*}{\makecell{GPT\\@Wikitext}}
  & $0.1, 0.1$ & $1898.81 \pm 13.93$ ($1.45\times$) & $5091.46 \pm 36.18$ ($3.90\times$) & $\mathbf{1306.29 \pm 10.29}$ \\
  & $0.5, 0.1$ & $868.18 \pm 6.00$ ($1.35\times$) & $1245.74 \pm 11.77$ ($1.94\times$) & $\mathbf{642.79 \pm 5.74}$ \\
  & $0.1, 1.0$ & $2481.77 \pm 19.40$ ($1.72\times$) & $5289.67 \pm 30.64$ ($3.67\times$) & $\mathbf{1440.59 \pm 11.87}$ \\
  & $0.5, 1.0$ & $1259.37 \pm 9.66$ ($1.26\times$) & $1461.31 \pm 11.71$ ($1.47\times$) & $\mathbf{997.25 \pm 7.92}$ \\
\midrule
\multirow{4}{*}{\makecell{ViT\\@ImageNet}}
  & $0.1, 0.1$ & $428.43 \pm 2.29$ ($1.51\times$) & $953.44 \pm 8.24$ ($3.37\times$) & $\mathbf{283.06 \pm 1.80}$ \\
  & $0.5, 0.1$ & $243.45 \pm 1.73$ ($1.44\times$) & $346.85 \pm 2.40$ ($2.05\times$) & $\mathbf{169.13 \pm 0.47}$ \\
  & $0.1, 1.0$ & $584.95 \pm 3.16$ ($1.47\times$) & $999.14 \pm 7.81$ ($2.52\times$) & $\mathbf{397.16 \pm 1.39}$ \\
  & $0.5, 1.0$ & $363.94 \pm 1.95$ ($1.54\times$) & $413.49 \pm 2.87$ ($1.75\times$) & $\mathbf{235.95 \pm 0.88}$ \\
\bottomrule
\end{tabular}
}
\end{table}

\subsection{Real Geo-Distributed Deployments}\label{exp:real-geo}

We further conduct real geo-distributed parameter-server deployments with four workers. For Shenzhen--Beijing, the server and two workers are placed in Shenzhen, and the other two workers are placed in Beijing. For China--U.S., we use the same four-worker setting across the two regions. The average bandwidth / end-to-end latency of the bottleneck worker-server path is $86.623$ Mbps / $45.008$ ms for Shenzhen--Beijing and $12.567$ Mbps / $161.822$ ms for China--U.S. As shown in Table~\ref{tab:real-geo}, \sysname{} remains effective under practical cross-region deployments. Standard deviations for \sysname{} and CocktailSGD are reported over multiple runs; the other baselines are evaluated with a single run under the same deployment.

\begin{table*}[!t]
\centering
\caption{Real geo-distributed parameter-server deployment results.}
\label{tab:real-geo}
\resizebox{\textwidth}{!}{
\begin{tabular}{@{}llcccccc@{}}
\toprule
\textbf{Deployment} & \makecell{\textbf{Model} \\ \textbf{@DataSet}} & \textbf{D-SGD} & \textbf{Accordion} & \textbf{DC2} & \textbf{DGA} & \textbf{CocktailSGD} & \textbf{\sysname} \\
\midrule
\multirow{2}{*}{Shenzhen--Beijing} & GPT@Wikitext & $9380.12$ ($6.11\times$) & $2159.20$ ($1.41\times$) & $2184.83$ ($1.42\times$) & $8013.96$ ($5.22\times$) & $1960.36 \pm 13.22$ ($1.28\times$) & $\mathbf{1535.98 \pm 10.35}$ \\
 & ViT@ImageNet & $1840.09$ ($5.78\times$) & $407.69$ ($1.28\times$) & $423.95$ ($1.33\times$) & $1548.21$ ($4.86\times$) & $374.22 \pm 1.97$ ($1.17\times$) & $\mathbf{318.63 \pm 1.94}$ \\
\midrule
\multirow{2}{*}{China--U.S.} & GPT@Wikitext & $60883.32$ ($10.82\times$) & $10292.28$ ($1.83\times$) & $11325.73$ ($2.01\times$) & $51869.97$ ($9.22\times$) & $6788.38 \pm 49.73$ ($1.21\times$) & $\mathbf{5628.56 \pm 36.67}$ \\
 & ViT@ImageNet & $9335.99$ ($6.42\times$) & $2074.27$ ($1.43\times$) & $2188.06$ ($1.51\times$) & $8201.98$ ($5.64\times$) & $1796.52 \pm 10.39$ (\sr$1.24\times$\er) & $\mathbf{1453.75 \pm 10.48}$ \\
\bottomrule
\end{tabular}
}
\end{table*}

\subsection{Network Robustness and Sensitivity}\label{exp:robustness}

We further evaluate the robustness of \sysname{} to packet loss, update interval, and bandwidth-estimation error under the base network setting $a{=}0.5$ Gbps and $b{=}0.1$ s. Table~\ref{tab:packet-loss} reports the full comparison under a packet loss rate of $0.01\%$. Packet loss substantially reduces the effective bandwidth, with the measured bandwidth dropping below $20$ Mbps in this setting, and therefore slows down all methods; nevertheless, \sysname{} still outperforms all baselines. Table~\ref{tab:sensitivity} further isolates the update interval $E$ and bandwidth-estimation error, which only affect \sysname{}'s online decision process. We use $E{=}10$ without estimation error as the reference setting. Reducing $E$ to $5$ leads to a moderate performance drop, consistent with the larger finite-window boundary effect predicted by the $\mathcal{O}(1/E)$ gap in Theorem~\ref{theorem:3}. The $\pm20\%$ bandwidth-estimation error directly perturbs the measured bandwidth. In both cases, \sysname{} remains faster than CocktailSGD. The other settings follow Table~\ref{tab:iid}.

\begin{table*}[!t]
\centering
\caption{Robustness results under packet loss.}
\label{tab:packet-loss}
\resizebox{\textwidth}{!}{
\begin{tabular}{@{}lcccccc@{}}
\toprule
\makecell{\textbf{Model} \\ \textbf{@DataSet}} & \textbf{D-SGD} & \textbf{Accordion} & \textbf{DC2} & \textbf{DGA} & \textbf{CocktailSGD} & \textbf{\sysname} \\
\midrule
GPT@Wikitext & $33862.05 \pm 249.83$ ($7.33\times$) & $7794.24 \pm 63.55$ (\sr$1.69\times$\er) & $7703.11 \pm 65.19$ ($1.67\times$) & $28104.67 \pm 24.14$ ($6.09\times$) & $5538.69 \pm 38.19$ ($1.20\times$) & $\mathbf{4617.217 \pm 30.75}$ \\
ViT@ImageNet & $6689.18 \pm 52.28$ ($6.21\times$) & $1620.53 \pm 11.64$ ($1.50\times$) & $1591.28 \pm 11.23$ (\sr$1.48\times$\er) & $5728.15 \pm 40.27$ ($5.32\times$) & $1276.29 \pm 10.82$ ($1.18\times$) & $\mathbf{1077.270 \pm 7.23}$ \\
\bottomrule
\end{tabular}
}
\end{table*}

\begin{table}[!t]
\centering
\caption{Sensitivity to update interval and bandwidth-estimation error. $E{=}10$, no estimation error is the reference setting.}
\label{tab:sensitivity}
\resizebox{\columnwidth}{!}{
\begin{tabular}{@{}llcc@{}}
\toprule
\makecell{\textbf{Model} \\ \textbf{@DataSet}} & \textbf{Setting} & \textbf{CocktailSGD} & \textbf{\sysname} \\
\midrule
\multirow{3}{*}{\makecell{GPT\\@Wikitext}}
  & $E{=}10$, no estimation error & $712.93 \pm 5.38$ ($1.11\times$) & $\mathbf{642.79 \pm 5.74}$ \\
  & $E{=}5$, no estimation error & $712.93 \pm 5.38$ ($1.07\times$) & $\mathbf{666.50 \pm 6.14}$ \\
  & $E{=}10$, bandwidth error $\pm20\%$ & $712.93 \pm 5.38$ ($1.05\times$) & $\mathbf{676.80 \pm 8.30}$ \\
\midrule
\multirow{3}{*}{\makecell{ViT\\@ImageNet}}
  & $E{=}10$, no estimation error & $219.39 \pm 0.98$ ($1.30\times$) & $\mathbf{169.13 \pm 0.47}$ \\
  & $E{=}5$, no estimation error & $219.39 \pm 0.98$ (\sr$1.22\times$\er) & $\mathbf{180.50 \pm 0.65}$ \\
  & $E{=}10$, bandwidth error $\pm20\%$ & $219.39 \pm 0.98$ ($1.16\times$) & $\mathbf{189.60 \pm 1.45}$ \\
\bottomrule
\end{tabular}
}
\end{table}

\section{Related Work}
\noindent \textbf{Communication-efficient and adaptive training.} To alleviate the communication bottlenecks in distributed training, gradient compression has become a standard practice. While early works focus on static compression strategies (\textit{e.g.}, fixed quantization bits or sparsity ratios), recent research has shifted towards dynamic and adaptive strategies to enhance robustness against system heterogeneity or network dynamics.
For instance, DAGC \cite{lu2023dagc} assigns different compression ratios based on the local data volume of each worker to address bottlenecks in communication-constrained non-IID scenarios. FedCG \cite{jiang2023heterogeneity} selects representative clients and adapts compression ratios according to their heterogeneity and capacity changes to make the system robust to device heterogeneity. 
HGC \cite{hu2024practical} compresses both uplink and downlink communications in federated learning, removing redundancy in the training process to greatly reduce data transfer while maintaining model performance. DC2 \cite{dc2} adaptively controls the gradient compression ratio based on network delays, effectively managing network variations to strike a better balance between accelerating distributed training and preserving model accuracy. These methods, however, are largely designed for bandwidth efficiency and remain insensitive to the fixed latency. This limitation leaves model performance vulnerable in challenging network environments where high latency coexists with limited bandwidth.

\noindent \textbf{Latency-tolerant training.} To enhance the tolerance of training to high latency, relaxation of the strict synchronization barrier is a common approach. Fully asynchronous methods are primarily designed to address the straggler effect caused by heterogeneous device capabilities.
For example, FedAC \cite{zang2024fedac} incorporates look-ahead momentum to correct updates in buffered aggregation, while FedFa \cite{xu2024fedfa} utilizes an update buffer to enable fully asynchronous updates. To address computational heterogeneity, Shadowheart SGD \cite{tyurinshadowheart} dynamically adjusts local update and communication frequencies, and P2P-AFL \cite{sad2025towards} proposes a decentralized model fusion mechanism. Although these approaches effectively reduce waiting times, the fully asynchronous strategy may introduce excessive staleness, leading to model accuracy degradation. It is also unsuitable for WAN-based training, which has relatively fixed latency. To address this, DGA \cite{DGA} introduces delayed aggregation into FL to mitigate the fixed end-to-end latency issue in FL. However, it does not account for communication bottlenecks. CO2 \cite{sun2024co2} introduces local update and delayed aggregation with $\tau{=}1$ to achieve complete overlap between communication and computing. However, it focuses on high-speed interconnects and is not suitable for adverse network conditions, which require adaptive adjustment of the staleness to ensure effective overlap.

\noindent \textbf{Distributed training in WANs.} WAN-based training faces the challenges of high latency and low, fluctuating bandwidth. DiLoCo \cite{douillard2023diloco} and OpenDiLoCo \cite{jaghouar2024opendiloco} propose a low-communication paradigm by synchronizing globally only after hundreds of local steps, effectively bypassing WAN latency but potentially introducing local model overfitting. DTFM \cite{yuan2022decentralized} and SWARM \cite{ryabinin2023swarm} introduce graph-based optimization and dynamic pipeline routing to handle heterogeneous and unstable connections. However, topology optimization and sparse synchronization alone cannot fully resolve the dynamic trade-off between communication bottlenecks and staleness noise. StellaTrain \cite{SIGCOMM24StellaTrain} integrates cache-aware compression for geo-distributed clusters but relies on empirical tuning and lacks rigorous theoretical guidance. Consequently, existing works fail to maintain optimal convergence rates under the extreme and fluctuating network conditions typical of WANs.

\noindent \textbf{Future Trends: Organizational Decentralization.} Looking forward, cross-datacenter training is evolving from simple resource pooling to \textit{organizational decentralization} \cite{dong25emnlpsurvey, llama}, where global schedulers like MAST \cite{choudhury2024mast} coordinate thousands of GPUs across regions. A key frontier lies in establishing a \textit{scaling law for decentralized training}. Recent work by Charles et al. \cite{scalinglaw2Diloco} has pioneered this direction by deriving scaling laws for DiLoCo, confirming the immense potential of decentralized paradigms to scale robustly with model size. While they validate the feasibility of this direction, our work focuses on addressing the specific throughput bottlenecks in cross-region training through lossy strategies. We provide a theoretical foundation for leveraging gradient compression and delayed aggregation to enable future large-scale distributed training.

\section{Conclusion}

In this work, we address the challenge of efficient distributed training in high-latency, low-bandwidth, and varying bandwidth networks to leverage islands of resources. Gradient compression and delayed aggregation are two common strategies to mitigate network constraints, but jointly employing them introduces a complex trade-off. We mathematically model the joint configuration problem from two perspectives: time-to-iteration and iteration-to-accuracy. For the former, we derive the threshold compression ratio that maximizes computation-communication overlap under the given staleness and network conditions. For the latter, we introduce the nested virtual sequence to derive the convergence rate of \algoname. We are the first to reveal that the staleness exponentially amplifies the negative impact of gradient compression in \algoname. Based on these insights, we propose \sysname, an adaptive optimizer that dynamically adjusts the compression ratio and staleness according to the communication and computation conditions. Our design achieves up to \sr$10.82\times$\er\ speed-up over D-SGD and \sr$1.36\times$\er\ over the SOTA method.

\bibliographystyle{IEEEtran}
\bibliography{main}

\newpage
\appendix
\addtocontents{toc}{\protect\setcounter{tocdepth}{2}}
\section{Notation List}

%

\begin{table}[!t]
\centering
\caption{Notation list (extended).}
\footnotesize
\renewcommand{\arraystretch}{0.92}
\scalebox{1}{
\begin{tabular}{ |c|c| }
\hline
Notation & Description \\
\hline
$n$ & number of workers\\
\hline
\(f_i(\cdot)\) & the local loss function of worker $i$\\
\hline
\(f(\cdot)\) & the global loss function, \textit{i.e.}, $f(\mathbf{x})=\frac{1}{n} \sum_{i=1}^n f_i(\mathbf{x})$ \\
\hline
$\mathbf{x}^i_t$ &  \makecell{the local model parameter in the $t$-th iteration \\ of the $i$-th worker}\\
\hline
$\mathbf{x}_t$ &  the global model parameter in the $t$-th iteration \\
\hline
$\mathbf{g}^i_t$ & the stochastic gradient of $\mathbf{x}_t$ of the worker $i$ \\
\hline
$\mathbf{g}_t$ & the average of $\mathbf{g}^i_t$, \textit{i.e.}, $\mathbf{g}_t=\frac{1}{n}\sum_{i=1}^n \mathbf{g}^i_t$\\
\hline
$\gamma$ & the stepsize \\
\hline
$\mathbf{T}$ & total number of iterations\\
\hline
$\delta$ & compression ratio (\(0 < \delta \leq 1\))\\
\hline
$\mathbf{C}_\delta(\cdot)$ & \makecell{the sparsification compressor with \\ compression ratio $\delta$}\\
\hline
$\tau$ & the delayed staleness\\
\hline
$\mathbf{e}_t^i$ & \makecell{the local error term of the $i$-th worker \\ in the $t$-th iteration}\\
\hline
$V_g$ & the volume of the gradient (bits) \\
\hline
$E$ & the frequency to update $\tau$ and $\delta$ with DeCo \\
\hline
$\bar{a}$ & \makecell{the average bandwidth (bits/s) of the slowest link \\ in previous $E$ iterations} \\
\hline
$b$ & the network end-to-end latency (s) of the slowest link \\
\hline
$T_{\text{comp}}$ & computation time per iteration (s)\\
\hline
$T_{\text{avg}}$ & average end-to-end training time per iteration (s)\\
\hline
$TS_t$ & end time of the $t$-th computation (s) \\
\hline
$TC_t$ & end time of the $t$-th communication (s) \\
\hline
$TM_t$ & \makecell{server-side message-ready time of the $t$-th \\ aggregation, i.e., $TC_t = TM_t + b$ (s)} \\
\hline
$\hat{\mathbf{x}}_t$ & \makecell{NVS hat sequence: virtual iterate after \\ removing compression noise} \\
\hline
$\tilde{\mathbf{x}}_t$ & \makecell{NVS tilde sequence: virtual iterate after \\ further removing delay-aggregation noise} \\
\hline
$B_t$ & \makecell{NVS noise term capturing the staled-compression \\ residual at iteration $t$} \\
\hline
$\tilde{B}_t$ & \makecell{NVS noise term capturing the delay-aggregation \\ residual at iteration $t$} \\
\hline
$\mathbf{x}^*$ & a minimizer of the global loss $f$ \\
\hline
$f^*$ & the optimal value $f(\mathbf{x}^*)$ \\
\hline
$L$ & Lipschitz-smoothness constant of $f$ \\
\hline
$\mu$ & strong-convexity constant of $f$ (when applicable) \\
\hline
$\sigma^2$ & uniform upper bound on stochastic-gradient noise variance \\
\hline
$\zeta^2$ & \makecell{uniform upper bound on inter-worker data \\ heterogeneity (additive part)} \\
\hline
$Z^2$ & \makecell{uniform upper bound on inter-worker data \\ heterogeneity (gradient-scaling part)} \\
\hline
$\phi(\delta,\tau)$ & \makecell{coupling term $\frac{1-\delta}{\delta}(1-\delta/2)^{-\tau}$ that \\ characterizes the joint compression--delay effect} \\
\hline
\end{tabular}
}
\label{table:notation_extended}
\end{table}

\section{Detailed proof}
\label{appendixB}

\subsection{Technical Results}
We list lemmas derived from other works here to help us complete the whole proof. Detailed proof of these lemmas can be found from the reference and we do not write here. For notational consistency, we define $\mathbb{E}\lVert\nabla f(\mathbf{x}_k)\rVert^2 = F_k = 0$ for all $k<0$.

\begin{lemma}\label{lemma_js}
    If $c_1, c_2 \in \mathbb{R}^d$ then the Jensen's inequality is: For all $\rho > 0$, we have
    \begin{equation}\label{jensen_1}
        \lVert c_1+c_2\rVert^2 \leq (1+\rho) \lVert c_1 \rVert^2 + (1+\rho^{-1}) \lVert c_2 \rVert^2.
    \end{equation}
    This can be written as:
    \begin{equation}\label{jensen_2}
        2\langle c_1,c_2\rangle \leq \rho \lVert c_1 \rVert^2 + \rho^{-1} \lVert c_2 \rVert^2.
    \end{equation}    
\end{lemma}

\begin{lemma}[The nature of Top-$k$]\label{lemma_topk}
 By definition, the Top-$k$ compressor $\mathbf{C}_{\delta}$ is a mapping that has the property $\mathbb{R}^d \rightarrow \mathbb{R}^d$:
\begin{equation}\label{top-k}
\mathbb{E}_{\mathbf{C}_{\delta}}\lVert \mathbf{C}_{\delta}(\mathbf{x}) - \mathbf{x} \rVert^2 \leq (1-\delta) \lVert \mathbf{x} \rVert^2.
\end{equation}
\end{lemma}

\begin{lemma}[Lemma 27 of the work \cite{stich2020communication}]\label{O_nonconvex}
\label{Lemma27}
Let $(r_t)_{t \geq 0}$ and $(s_t)_{t \geq 0}$ be sequences of positive numbers satisfying
\[
r_{t+1} \leq r_t - B \gamma s_t + C \gamma^2 + D \gamma^3,
\]
for some positive constants $B > 0$, $C, D \geq 0$ and step-sizes $0 < \gamma \leq \frac{1}{E}$, for $E \geq 0$. Then there exists a constant stepsize $\gamma \leq \frac{1}{E}$ such that
\begin{equation}
\label{B/T+1}
    \frac{B}{T+1} \sum_{t=0}^T s_t \leq \frac{E r_0}{T+1} + 2 D^{1/3} \left( \frac{r_0}{T+1} \right)^{2/3} + 2 \left( \frac{C r_0}{T+1} \right)^{1/2}.
\end{equation}

\end{lemma}

\begin{remark}
\textit{To ensure that the right hand side in Eq.~\ref{B/T+1} is less than $\epsilon > 0$,}
\[
T = \mathcal{O}\left( \frac{C}{\epsilon^2} + \frac{\sqrt{D}}{\epsilon^{3/2}} + \frac{E}{\epsilon} \right) \cdot r_0
\]
steps are sufficient.
\end{remark}

\begin{lemma}[Lemma 25 of the work \cite{stich2020communication}]\label{O_convex}
Let $(r_t)_{t \geq 0}$ and $(s_t)_{t \geq 0}$ be sequences of positive numbers satisfying
\begin{equation*}
    r_{t+1} \leq (1 - \gamma A) r_t - B \gamma s_t + C \gamma^2 + D \gamma^3,
\end{equation*}
for some positive constants $A, B > 0$, $C, D \geq 0$, and for constant step-sizes $0 < \gamma \leq \frac{1}{E}$, for $E \geq 0$. Then we have 
\begin{align*}
    \frac{B}{W_T} \sum_{t=0}^{T} w_t s_t + A r_{T+1}
     \leq r_0  E
    \exp\left[
        -  \frac{A}{E}(T + 1)
    \right] + \frac{2 C \ln \tau}{A(T+1)} + \frac{D \ln^2 \tau}{A^2(T+1)^2},
\end{align*}
for $w_t := (1 - \gamma A)^{-(t+1)}$, $W_T := \sum_{t=0}^{T} w_t$ and $\tau = \max \left\{ \exp[1], \min \left\{
        \frac{A^2 r_0 (T + 1)^2}{C}, \frac{A^3 r_0 (T + 1)^3}{D}
    \right\} \right\}$.
\end{lemma}

\begin{remark}
\textit{Lemma ~\ref{O_convex}} establishes a bound of the order
\[
\tilde{\mathcal{O}}\left( r_0 E  \exp\left[ - \frac{A}{E} \cdot T \right]
+ \frac{C}{A T} + \frac{D}{A^2 T^2} \right),
\]
that decreases with $T$. To ensure that this expression is less than $\epsilon$,
\[
T = \tilde{\mathcal{O}}\left( \frac{C}{A \epsilon} + \frac{\sqrt{D}}{A \sqrt{\epsilon}} +  \frac{E}{A} \log \frac{1}{\epsilon} \right)
= \tilde{\mathcal{O}}\left( \frac{C}{A \epsilon} + \frac{\sqrt{D}}{A \sqrt{\epsilon}} + \frac{E}{A} \right)
\]
steps are sufficient.
\end{remark}

\subsection{Proof of Theorem 1}\label{sec:proof2T1}
\begin{lemma}\label{lemma1}
        Let $\left\{\mathbf{x}_t, \tilde{\mathbf{x}}_t,
        \hat{\mathbf{x}}_t,
        B_t, \tilde{B}_t, \mathbf{v}_t, \tilde{\mathbf{v}}_t\right\}_{t \geq 0}$ be defined as in \algoname\ using NVS, with gradient oracle $\left\{\mathbf{g}_t\right\}_{t \geq 0}$ and let $f: \mathbb{R}^d \rightarrow \mathbb{R}$  be  $L$-smooth. If $\gamma \leq \frac{1}{4L}$, then we have

\begin{equation*}
\begin{aligned}
\mathbb{E} f\left(\hat{\mathbf{x}}_{t+1}\right) 
\leq  \mathbb{E} f\left(\hat{\mathbf{x}}_{t}\right)  - \frac{\gamma}{2} \mathbb{E} \left\|\nabla f\left(\mathrm{x}_t\right)\right\|^2  + \gamma L^2 (2\mathbb{E} \left\|\tilde{B}_t\right\|^2 +  \mathbb{E} \left\|B_t\right\|^2) +\frac{\gamma^2 L \sigma^2}{2 n}. 
\end{aligned}
\end{equation*}
\end{lemma}

\textit{Proof.} We begin with the definition of $\hat{\mathbf{x}}_{t+1}$ and have
\begin{equation}
\begin{aligned}
\label{x_hat}
 \mathbb{E} f\left(\hat{\mathbf{x}}_{t+1}\right)
 &= \mathbb{E} f\left(\hat{\mathbf{x}}_{t} - \frac{\gamma}{n} \sum_{i=1}^n \mathbf{g}_t^i \right) \\
&\leq \mathbb{E} f\left(\hat{\mathbf{x}}_t\right)-
 \gamma \mathbb{E}\left\langle\nabla f\left(\hat{\mathbf{x}}_t\right), \mathbf{g}_t\right \rangle +\frac{L}{2}\mathbb{E} \left\| \frac{\gamma}{n}\sum_{i=1}^n \mathbf{g}_t^i\right\|^2 \\
& \leq  \mathbb{E} f\left(\hat{\mathbf{x}}_t\right)-
\underbrace{ \gamma \mathbb{E}\left\langle\nabla f\left(\hat{\mathbf{x}}_t\right), \mathbf{g}_t\right\rangle}_{:= A_1} +\frac{L\gamma^2}{2}\mathbb{E}\left\| \nabla f (\mathbf{x}_t) \right\|^2 + \frac{L \gamma^2 \sigma^2}{2 n}.
\end{aligned}
\end{equation}

The first line is due to $\hat{\mathbf{x}}_{t+1} = \hat{\mathbf{x}}_{t} - \frac{\gamma}{n} \sum_{i=1}^n \mathbf{g}_t^i$. The second line is due to Eq.~\ref{L-smooth} and the third line is due to Eq.~\ref{xi}. We estimate $A_1$ as 
\begin{equation}
\begin{aligned}
\label{A-1}
& -A_1
 = \gamma \mathbb{E} \langle-\nabla f\left(\hat{\mathbf{x}}_t\right), \mathbf{g}_t\rangle \\
&=  \gamma ( \mathbb{E}\left\langle  \nabla f\left(\tilde{\mathbf{x}}_t\right) -\nabla f\left(\hat{\mathbf{x}}_t\right)  , \mathbf{g}_t\right\rangle + \mathbb{E}\left\langle  \nabla f\left(\mathbf{x}_t\right) - \nabla f \left(\tilde{\mathbf{x}}_t\right)  , \mathbf{g}_t\right\rangle -  \mathbb{E} \langle\nabla f\left(\mathbf{x}_t\right), \mathbf{g}_t\rangle ) \\
&\leq  \frac{c_1\gamma L^2}{2} \mathbb{E} \lVert \tilde{B}_t \rVert ^2 + \frac{\gamma}{2 c_1} \mathbb{E} \lVert \nabla f(\mathbf{x}_t) \rVert ^2 +\frac{c_2 \gamma L^2}{2} \mathbb{E} \lVert B_t \rVert^2 + \frac{\gamma}{2 c_2}  \mathbb{E} \lVert \nabla f(\mathbf{x}_t) \rVert ^2 -\gamma \mathbb{E}  \lVert \nabla f(\mathbf{x}_t) \rVert ^2 \\
& \leq -\frac{5\gamma}{8} \mathbb{E} \lVert \nabla f(\mathbf{x}_t) \rVert ^2 + \gamma L^2 (2\mathbb{E} \lVert \tilde{B}_t \rVert^2 + \mathbb{E} \lVert B_t \rVert^2).
\end{aligned}
\end{equation}

The third line is due to Jensen's inequality (\textit{i.e.}, Eq.~\ref{jensen_2}). In the fourth line, we let $c_1=4$, $c_2=2$. 
Bringing Eq.~\ref{A-1} into Eq.~\ref{x_hat}, we complete the proof.

\begin{lemma}\label{lemma2}

Define $\{ \tilde{B}_t\}_{t\geq 0}$ as in Eq.~\ref{NVS2}. Then with $\gamma \leq \frac{1}{4L\tau}$, we have

\begin{equation*}
\begin{aligned}
\sum_{t=0}^T \mathbb{E} \lVert
\tilde{B}_t \rVert ^2 \leq 
\frac{1}{16L^2} \sum_{t=0}^T \mathbb{E}\lVert \nabla f (\mathbf{x}_t) \rVert ^2 + \frac{\tau \gamma^2 (1+T) \sigma^2}{n}.
\end{aligned}
\end{equation*}
\end{lemma}

\textit{Proof.}
\begin{equation*}
\begin{aligned}
\mathbb{E} \lVert
\tilde{B}_t\rVert^2 &= \gamma^2 \mathbb{E}\lVert
\sum_{j=1}^\tau(\nabla f(\mathbf{x}_{t-j})+\frac{1}{n}\sum_{i=1}^n \xi^i )
\rVert ^2 
\\&\leq \gamma^2(\tau \sum_{j=1}^\tau \mathbb{E}\lVert \nabla f (\mathbf{x}_{t-j}) \rVert ^2 + \frac{1}{n^2}\cdot n \tau \sigma^2).
\end{aligned}
\end{equation*}

The inequality is due to Jensen's inequality and Eq.~\ref{xi}. Then we have 
\begin{equation}\label{eqInLemma6}
\begin{aligned}
\sum_{t=0}^T \mathbb{E} \lVert
\tilde{B}_t\rVert^2 & \leq \gamma^2 \tau \sum_{t=0}^T \sum_{j=1}^\tau  \mathbb{E}\lVert\nabla f(\mathbf{x}_{t-j}) \rVert^2 + \frac{\gamma^2  \tau \sigma^2(1+T)}{n} \\
&\leq \gamma^2 \tau \cdot \tau \sum_{t=0}^{T-\tau} \mathbb{E}\lVert \nabla f(\mathbf{x}_{t}) \rVert^2 +  \frac{\gamma^2  \tau \sigma^2(1+T)}{n} \\
&\leq \gamma^2 \tau^2 \sum_{t=0}^{T} \mathbb{E}\lVert \nabla f (\mathbf{x}_t) \rVert ^2 + \frac{\gamma^2  \tau \sigma^2(1+T)}{n}.
\end{aligned}
\end{equation}

The second line is due to the fact that the term $\mathbb{E}\lVert \nabla f (\mathbf{x}_t) \rVert ^2$ appears at most $\tau$ times. Let $\gamma \leq \frac{1}{4L\tau}$, we complete the proof.


\begin{lemma}\label{lemma3}
Define $\{ B_t\}_{t\geq 0}$ as in Eq.~\ref{NVS1}. Then by using the relative compressor, with $\gamma \leq \frac{1}{4LZ\sqrt{\phi / \delta}}$, where $\phi=\frac{1-\delta}{\delta (1-\frac{\delta}{2})^{\tau}}$, we have
\begin{equation*}
\begin{aligned}
\sum_{t=0}^T \mathbb{E}\lVert B_t \rVert ^2 \leq \frac{1}{4L^2} \sum_{t=0}^T\mathbb{E}  \lVert \nabla f (\mathbf{x}_t) \rVert ^2+ 2 \gamma^2 \phi (T+1) (\frac{2}{\delta} \zeta^2 + \sigma^2). 
\end{aligned}
\end{equation*}
\end{lemma}

\textit{Proof.} 
\begin{equation}
\begin{aligned}
\label{EF_iter}
\frac{1}{n} \sum_{i=1}^n \mathbb{E}\lVert e_{t-\tau+1}^i \rVert^2 
&\leq \frac{(1-\delta)(1+\rho)}{n} \sum_{i=1}^n \mathbb{E}\lVert e_{t-\tau}^i \rVert^2  \\& \quad +\gamma^2 (1-\delta) ((1+\rho^{-1})Z^2 \mathbb{E} \lVert \nabla f (\mathbf{x}_{t-\tau}) \rVert^2 + (1+\rho^{-1})\zeta^2 + \sigma^2)
\\& \leq \frac{1}{n} (1-\frac{\delta}{2}) \sum_{i=1}^n \mathbb{E}\lVert e_{t-\tau}^i \rVert^2 + \gamma^2 (1-\delta)(\frac{2Z^2}{\delta}\mathbb{E} \lVert \nabla f (\mathbf{x}_{t-\tau}) \rVert^2 + \frac{2}{\delta}\zeta^2 + \sigma^2 ).
\end{aligned}
\end{equation}

The first inequality is due to Lemma~\ref{lemma_topk}, Eq.~\ref{zeta}, \ref{xi} and \ref{jensen_1}. In the second line, we let $\rho=\frac{\delta}{2-\delta}$, having $1+\rho^{-1}=\frac{2}{\delta}$ and $(1-\delta)(1+\rho)=1-\frac{\delta}{2-\delta}<1-\frac{\delta}{2}$.

By unrolling the recurrence, we can derive the upper bound:

\begin{equation*}
\begin{aligned}
\frac{1}{n} \sum_{i=1}^n \mathbb{E}\lVert e_{t-\tau+1}^i \rVert^2  &\leq \gamma^2 \frac{(1-\delta)}{(1-\frac{\delta}{2})^{\tau}} \sum_{i=0}^{t-\tau} (1-\frac{\delta}{2})^{t-i} (\frac{2Z^2}{\delta} \mathbb{E} \lVert \nabla f(\mathbf{x}_i) \rVert^2 + \frac{2}{\delta} \zeta^2 + \sigma^2) \\
& \leq \frac{\gamma^2 (1-\delta)}{(1-\frac{\delta}{2})^{\tau}}  \sum_{i=0}^{t} (1-\frac{\delta}{2})^{t-i} (\frac{2Z^2}{\delta} \mathbb{E} \lVert \nabla f(\mathbf{x}_i) \rVert^2 + \frac{2}{\delta}\zeta^2 + \sigma^2), \\
\sum_{t=0}^T [\frac{1}{n} \sum_{i=1}^n \mathbb{E}\lVert e_{t-\tau+1}^i \rVert^2] 
&\leq \frac{\gamma^2 (1-\delta)}{(1-\frac{\delta}{2})^{\tau}} \sum_{j=0}^\infty 
(1-\frac{\delta}{2})^j \sum_{i=0}^{T} 
(\frac{2Z^2}{\delta} \mathbb{E} \lVert \nabla f(\mathbf{x}_i)
\rVert^2 + \frac{2}{\delta}\zeta^2 + \sigma^2) \\
&\leq \frac{ 2 \gamma^2 (1-\delta)}{\delta (1-\frac{\delta}{2})^{\tau}} \sum_{i=0}^{T}(\frac{2Z^2}{\delta} \mathbb{E} \lVert \nabla f(\mathbf{x}_i) \rVert^2 + \frac{2}{\delta} \zeta^2 + \sigma^2).
\end{aligned}
\end{equation*}

Choosing $\frac{ 4 \gamma^2 (1-\delta)Z^2}{\delta^2 (1-\frac{\delta}{2})^{\tau}} \leq \frac{1}{4L^2} $, we have 
\begin{equation*}
\begin{aligned}
&\sum_{t=0}^T [\frac{1}{n} \sum_{i=1}^n \mathbb{E}\lVert e_{t-\tau+1}^i \rVert^2]
 \leq \frac{1}{4L^2} \sum_{i=0}^T \mathbb{E} \lVert \nabla f(\mathbf{x}_i) \rVert^2 + \frac{ 2 \gamma^2 (1-\delta) (1+T)}{\delta (1-\frac{\delta}{2})^{\tau}} ( \frac{2}{\delta}\zeta^2 + \sigma^2).
\end{aligned}
\end{equation*}

\noindent\textbf{Restatement of Theorem 1 (Non-convex convergence rate of \algoname).}
Let $f: \mathbb{R}^d \rightarrow \mathbb{R}$ be $L$-smooth.  There exists a stepsize $\gamma \leq \min \{\frac{1}{4L\tau}, \frac{1}{4LZ\sqrt{\phi / \delta}}\}$, where  $\phi=\frac{1-\delta}{\delta (1-\frac{\delta}{2})^{\tau}}$, such that at most
\begin{equation*}
\begin{split}
    \mathcal{O}(&\frac{\sigma^2}{n\epsilon^2}+\frac{ \sqrt{(\frac{\phi\zeta^2}{\delta} + (\phi+\frac{\tau}{n})\sigma^2)}}{\epsilon^{3/2}}+\frac{\tau}{\epsilon}+\frac{Z\sqrt{\phi}}{\sqrt{\delta} \epsilon})\cdot L (f(\mathbf{x}_0) - f^*)
\end{split}
\end{equation*}
iterations of \algoname, it holds $\mathbb{E}\lVert \nabla f(\mathbf{x}_{out}) \rVert^2 \leq \epsilon $, and $\mathbf{x}_{out} = \mathbf{x}_t$ denotes an iterate $\mathbf{x}_t \in \left\{\mathbf{x}_0, \ldots, \mathbf{x}_{\mathbf{T}-1}\right\}$, where $\mathbf{T}$ denotes the total iterations, chosen at random uniformly.

\textit{Proof.}
Let $G_t:=\sum_{i=0}^t \mathbb{E} \left\|\nabla f\left(\mathrm{x}_i\right)\right\|^2$. According to above lemmas (Lemma~\ref{lemma1}, \ref{lemma2}, \ref{lemma3}), we have
\begin{equation*}
\begin{aligned}
  \frac{\gamma}{2}G_T  &\leq \mathbb{E} f(\hat{\mathbf{x}}_0) - \mathbb{E} f(\hat{\mathbf{x}}_{T+1}) +\gamma L^2 (2\sum_{i=0}^T \lVert \tilde{B}_t \rVert^2 + \sum_{i=0}^T \lVert B_t \rVert^2) 
  +\frac{\gamma^2 L (T+1) \sigma^2}{2n}
  \\ 
  &\leq  \mathbb{E} f(\mathbf{x}_0) - f^* + \frac{\gamma}{4}G_T + \frac{\gamma}{8}G_T  + \frac{\gamma^2 L (T+1) \sigma^2 }{2 n} 
  + \gamma^3 L^2 (T+1) (2\phi \frac{2}{\delta}\zeta^2 + 2\phi \sigma^2 + \frac{2\tau \sigma^2}{n}).
\end{aligned}
\end{equation*}

Divide both sides of the inequality by $4\gamma (1+T)$, and we have:
\begin{equation*}
\begin{aligned}
  & \frac{1}{8(1+T)} G_T \leq \frac{\mathbb{E} f(\mathbf{x}_0) - f^*}{\gamma (T+1)} + \gamma \frac{L\sigma^2}{2 n}  + 2\gamma^2 L^2 (\frac{2\phi}{\delta}\zeta^2 + \phi \sigma^2 + \frac{\tau \sigma^2}{n} ).
\end{aligned}
\end{equation*}

We do not believe that \algoname{} will degrade into D-SGD, meaning that $\tau=0$ and $\delta=1$ cannot coexist at the same time. In this way, we have $\gamma\leq \min \{ \frac{1}{4L}, \frac{1}{4L\tau},\frac{1}{4LZ\sqrt{\phi/\delta}} \} \leq \min\{ \frac{1}{4L\tau},\frac{1}{4LZ\sqrt{\phi/\delta}}\}$.

According to Lemma~\ref{O_nonconvex}, we let $B=1/8$, $C=\frac{L\sigma^2}{2 n}$, $D=L^2 (\frac{2\phi}{\delta}\zeta^2 + \phi \sigma^2 + \frac{\tau \sigma^2}{n} )$, $E= \max\{4L\tau,4LZ\sqrt{\phi/\delta}\}$, $r_0=\mathbb{E} f(\mathbf{x} _0) - f^*$. To ensure the right hand side of the above inequality is less than $\epsilon>0$, $T=\mathcal{O}(\frac{C}{\epsilon^2}+\frac{\sqrt{D}}{\epsilon^{3/2}}+\frac{E}{\epsilon})\cdot r_0$ iterations are sufficient, which completes the proof.

\subsection{Proof of Theorem 2}\label{proof2T2}

For further discussion, let's define  $F_t:=\mathbb{E} f(\mathbf{x}_t) - f^*$ and  $X_t:=\mathbb{E} \lVert \hat{\mathbf{x}}_t - \mathbf{x}^* \rVert ^ 2 $. 


\begin{lemma}
Let $\left\{\mathbf{x}_t, \tilde{\mathbf{x}}_t,
        \hat{\mathbf{x}}_t,
        B_t, \tilde{B}_t, \mathbf{v}_t, \tilde{\mathbf{v}}_t\right\}_{t \geq 0}$ be defined as in \algoname\ using NVS, with gradient oracle $\left\{\mathbf{g}_t\right\}_{t \geq 0}$ and let $f: \mathbb{R}^d \rightarrow \mathbb{R}$  be $L$-smooth and $\mu$-strongly convex. If $\gamma \leq \frac{1}{4L}$, then we have
        
\begin{equation*}
\begin{aligned}
    X_{t+1}&\leq (1-\frac{\gamma \mu}{2})X_t - \frac{\gamma}{2}F_t + \gamma^2 \sigma^2/n
     + 6 \gamma L (\mathbb{E} \lVert
\tilde{B}_t \rVert ^2 +\mathbb{E}\lVert  B_t \rVert ^2).
\end{aligned}
\end{equation*}
\end{lemma}

\textit{Proof.}

\begin{equation*}
\begin{aligned}
    X_{t+1}& = X_{t} + \gamma^2 \lVert \frac{1}{n} \sum_{i=1}^n \mathbf{g}_t^i \rVert^2 - \underbrace{ 2 \gamma \mathbb{E}\left\langle \hat{\mathbf{x}}_t -\mathbf{x}^*, \mathbf{g}_t\right\rangle}_{:= A_2}.
\end{aligned}
\end{equation*}

We estimate the term $A_2$ as 
\begin{equation*}
\begin{aligned}
-A_2&
 =  - 2\gamma (\mathbb{E}\left\langle  \nabla f\left(\mathbf{x}_t\right), \mathbf{x}_t-\mathbf{x}^*\right\rangle  - \mathbb{E}\left\langle  \nabla f\left(\mathbf{x}_t\right), \tilde{\mathbf{x}}_t-\hat{\mathbf{x}}_t\right\rangle   - \mathbb{E}\left\langle  \nabla f\left(\mathbf{x}_t\right), \mathbf{x}_t-\tilde{\mathbf{x}}_t\right\rangle ) \\
&\leq -\gamma \mu \mathbb{E} \lVert \mathbf{x}_t - \mathbf{x}^* \rVert ^ 2 - 2\gamma F_t + \gamma F_t  + 4\gamma L \mathbb{E} \lVert
\tilde{B}_t \rVert ^2  + 4\gamma L \mathbb{E}\lVert  B_t \rVert ^2 \\
& \leq -\frac{\gamma \mu}{2} X_t -\gamma F_t + \gamma(4L+2\mu) (\mathbb{E} \lVert
\tilde{B}_t \rVert ^2  + \mathbb{E}\lVert  B_t \rVert ^2).
\end{aligned}
\end{equation*}

The second line is due to the fact that $f: \mathbb{R}^d \rightarrow \mathbb{R}$ is $L$-smooth and $\mu$-strongly convex. Then choosing $\gamma \leq \frac{1}{4L}, \mu \leq L$, we have 
\begin{equation*}
\begin{aligned}
    X_{t+1}& \leq  (1-\frac{\gamma \mu }{2})X_{t} + (2L\gamma^2 - \gamma)F_t  + \gamma(4L+2\mu)(\mathbb{E} \lVert
\tilde{B}_t \rVert ^2  + \mathbb{E}\lVert  B_t \rVert ^2)+\gamma ^2 \sigma^2/n  \nonumber \\
    & \leq  (1-\frac{\gamma \mu }{2})X_{t} - \frac{\gamma}{2} F_t + 6L\gamma (\mathbb{E} \lVert
\tilde{B}_t \rVert ^2  + \mathbb{E}\lVert  B_t \rVert ^2) +\gamma ^2 \sigma^2/n.
\end{aligned}
\end{equation*}

This completes the proof.

\begin{lemma}\label{tildeB_nonconvex}
Let $w_t = (1-\frac{\gamma \mu }{2})^{-(t+1)}$ and define $\tilde{B}_t$ as in Eq.~\ref{NVS2}. Then with $\gamma \leq \frac{1}{8\sqrt{2}\ L\tau}$, we have

\begin{equation*}
\begin{aligned}
\sum_{t=0}^T w_t\mathbb{E} \lVert
\tilde{B}_t \rVert ^2 \leq 
\frac{1}{32L} \sum_{t=0}^T w_t F_t + \frac{W_T \tau \gamma^2 \sigma^2}{n},
\end{aligned}
\end{equation*}
where $W_t = \sum_{i=0}^t w_i$.
\end{lemma}

\textit{Proof.}  
Based on Eq.~\ref{eqInLemma6}, when $\gamma \leq \frac{1}{8\sqrt{2}L\tau}$, we have
\begin{equation*}
\begin{aligned}
\mathbb{E} \lVert
\tilde{B}_t \rVert ^2 &\leq 
\frac{1}{128 L^2\tau} \sum_{j=1}^{\tau} \mathbb{E}\lVert \nabla f (\mathbf{x}_{t-j}) \rVert ^2 + \frac{ \tau \gamma^2 \sigma^2}{n} \\
&\leq \frac{1}{64 L\tau} \sum_{j=1}^{\tau} F_{t-j} + \frac{ \tau \gamma^2 \sigma^2}{n}.
\end{aligned}
\end{equation*}

The second inequality is due to the assumption of L-smoothness $\mathbb{E}\lVert \nabla f (\mathbf{x}_{t}) \rVert^2 \leq 2L F_t$.

We observe that when $\tau\geq 1$, we have $w_t=(1-\frac{\mu\gamma}{2})^{-(t+1)}$, $1-\frac{\mu\gamma}{2}\geq 1-\frac{\mu}{8\sqrt{2}\ L\tau}\geq 1-\frac{1}{8\sqrt{2}\ \tau}\geq(1+\frac{1}{2\tau})^{-1}$, thus having $w_t=w_{t-j}(1-\frac{\mu\gamma}{2})^{-j}\leq w_{t-j}(1+\frac{1}{2\tau})^j\leq w_{t-j}(1+\frac{1}{2\tau})^\tau \leq w_{t-j}\exp[1/2]\leq  2w_{t-j}$. Based on this inequality, we have 

\begin{equation*}
\begin{aligned}
\sum_{t=0}^Tw_t\mathbb{E} \lVert
\tilde{B}_t \rVert ^2 &\leq 
\frac{1}{32L}\sum_{t=0}^T\frac{w_t}{2\tau} \sum_{j=1}^{\tau} F_{t-j} + \frac{W_T \tau \gamma^2 \sigma^2}{n}\\
&\leq \frac{1}{32L} \sum_{t=0}^T \frac{1}{\tau}\sum_{j=1}^{\tau}w_{t-j} F_{t-j} + \frac{W_T \tau \gamma^2 \sigma^2}{n} \\
&\leq \frac{1}{32L}\sum_{t=0}^T w_t F_t + \frac{W_T \tau \gamma^2 \sigma^2}{n}.
\end{aligned}
\end{equation*}

\begin{lemma}
Let $w_t = (1-\frac{\gamma \mu }{2})^{-(t+1)}$ and define $B_t$ as in Eq.~\ref{NVS1}. Then by using the relative compressor, with $\gamma \leq \frac{1}{16LZ\sqrt{2\phi/\delta}}$, where $\phi=\frac{1-\delta}{\delta (1-\frac{\delta}{2})^{\tau}}$ we have
\begin{equation*}
\begin{aligned}
\sum_{t=0}^T w_t \mathbb{E}\lVert  B_t \rVert ^2 & \leq \frac{1}{32L} \sum_{t=0}^T w_t F_t + 2 \gamma^2 \phi W_T (\frac{2}{\delta} \zeta^2 + \sigma^2), 
\end{aligned}
\end{equation*}
where $W_t = \sum_{i=0}^t w_i$.
\end{lemma}

\textit{Proof.} We denote $E_t = \frac{1}{n} \sum_{i=1}^n \mathbb{E}\lVert e_{t-\tau+1}^i \rVert^2$, and based on Eq.~\ref{EF_iter}, then we have:

\begin{equation}
\begin{aligned}
 w_t E_t 
& \leq \gamma^2 \frac{(1-\delta)}{(1-\frac{\delta}{2})^{\tau}} \sum_{i=0}^{t-\tau} w_t (1-\frac{\delta}{2})^{t-i} (\frac{2Z^2}{\delta} \mathbb{E} \lVert \nabla f(\mathbf{x}_i) \rVert^2 + \frac{2}{\delta} \zeta^2 + \sigma^2) \\
& \leq \frac{\gamma^2 (1-\delta)}{(1-\frac{\delta}{2})^{\tau}}  \sum_{i=0}^{t} w_t (1-\frac{\delta}{2})^{t-i} (\frac{2Z^2}{\delta} \mathbb{E} \lVert \nabla f(\mathbf{x}_i) \rVert^2 + \frac{2}{\delta}\zeta^2 + \sigma^2) \\
& \leq \frac{\gamma^2 (1-\delta)}{(1-\frac{\delta}{2})^{\tau}}  \sum_{i=0}^{t} (1-\frac{\delta}{2})^{t-i}  w_i (1+\frac{\delta}{4})^{t-i} (\frac{2Z^2}{\delta} \mathbb{E} \lVert \nabla f(\mathbf{x}_i) \rVert^2 ) \\
& \quad + \frac{\gamma^2 (1-\delta)}{(1-\frac{\delta}{2})^{\tau}}  \sum_{i=0}^{t} w_t (\frac{2}{\delta}\zeta^2 + \sigma^2)(1-\frac{\delta}{2})^{t-i} \\
&\leq  \frac{\gamma^2 (1-\delta)}{(1-\frac{\delta}{2})^{\tau}} [\sum_{i=0}^{t} (1-\frac{\delta}{4})^{t-i}  w_i \frac{2Z^2}{\delta} \mathbb{E} \lVert \nabla f(\mathbf{x}_i) \rVert^2 + \sum_{i=0}^{t} w_t (\frac{2}{\delta}\zeta^2 + \sigma^2)(1-\frac{\delta}{2})^{t-i} ], \nonumber \\
\end{aligned}
\end{equation}

\begin{equation}
\begin{aligned}
\sum_{t=0}^T w_t E_t 
&\leq \frac{\gamma^2 (1-\delta)}{(1-\frac{\delta}{2})^{\tau}} [ \sum_{j=0}^\infty 
(1-\frac{\delta}{4})^j \sum_{i=0}^{T} w_i 
\frac{2Z^2}{\delta} \mathbb{E} \lVert \nabla f(\mathbf{x}_i) \rVert^2  + \sum_{j=0}^\infty 
(1-\frac{\delta}{2})^j W_T (\frac{2}{\delta}\zeta^2 + \sigma^2)] \\
& \leq \frac{ 2 \gamma^2 (1-\delta)}{\delta (1-\frac{\delta}{2})^{\tau}} [ \frac{4Z^2}{\delta} \sum_{i=0}^{T} w_i \mathbb{E} \lVert \nabla f(\mathbf{x}_i) \rVert^2 + W_T(\frac{2}{\delta} \zeta^2 + \sigma^2)]
\\& \leq \frac{ 2 \gamma^2 (1-\delta)}{\delta (1-\frac{\delta}{2})^{\tau}} [ \frac{8LZ^2}{\delta} \sum_{i=0}^{T} w_i F_i + W_T(\frac{2}{\delta} \zeta^2 + \sigma^2)]
. \nonumber
\end{aligned}
\end{equation}

Choosing $\frac{ 16LZ^2 \gamma^2 (1-\delta)}{\delta^2 (1-\frac{\delta}{2})^{\tau}} \leq \frac{1}{32L} $, we complete the proof.

\noindent\textbf{Restatement of Theorem 2 (Convex convergence rate of \algoname).}
Let $f: \mathbb{R}^d \rightarrow \mathbb{R}$ be $L$-smooth and $\mu$-convex. Then there exists a stepsize $\gamma \leq \min \{  \frac{1}{8\sqrt{2}\ L\tau }, \frac{1}{16LZ\sqrt{2\phi/\delta}}\}$, such that at most
\begin{equation*}
\begin{split}
    \mathcal{O}(&\frac{\sigma^2}{n\mu\epsilon}+\frac{ \sqrt{L(\frac{\phi\zeta^2}{\delta} + (\phi+\frac{\tau}{n})\sigma^2)}}{\mu \epsilon^{1/2}}+\frac{L(\tau+Z\sqrt{\phi/\delta})}{\mu})
\end{split}
\end{equation*}
iterations of \algoname, it holds $\mathbb{E}f(\mathbf{x}_{out})-f^* \leq \epsilon $, and $\mathbf{x}_{out} = \mathbf{x}_t$ denotes an iterate $\mathbf{x}_t \in \left\{\mathbf{x}_0, \ldots, \mathbf{x}_{T-1}\right\}$,  selected probabilistically based on $(1-\frac{\mu \gamma}{2})^{-t}$.

\textit{Proof.}
By substituting Lemma 5, 6 into Lemma 4, we can get 
\begin{equation*}
\begin{aligned}
& w_T X_{T+1} \leq X_0 - \frac{\gamma}{2}\sum_{t=0}^T w_t F_t + \frac{W_T \gamma^2 \sigma^2}{n}  + 6\gamma L (\sum_{t=0}^T w_t \mathbb{E} \lVert
\tilde{B}_t \rVert ^2  + \sum_{t=0}^T w_t \mathbb{E}\lVert  B_t \rVert ^2) \\
&\leq X_0 - \frac{\gamma}{8}\sum_{t=0}^T w_t F_t + \frac{W_T \gamma^2 \sigma^2}{n}  + 6LW_T\gamma^3(\frac{\tau \sigma^2}{n} + \frac{4\phi\zeta^2}{\delta}+2\phi\sigma^2).
\end{aligned}
\end{equation*}
According to Lemma~\ref{O_convex}, we complete the proof.

\subsection{Proof of Convergence rate of D-SGD with delayed aggregation}\label{supp:proofddsgd}
\setcounter{theorem}{3}

The update rule of D-SGD with delayed aggregation is: $\mathbf{x}_{t+1}=\mathbf{x}_t - \frac{\gamma}{n} \sum_{i=1}^n \mathbf{g}_{t-\tau}^i$. Based on the virtual sequence, we have 
\begin{equation}\label{dd-sgd}
\begin{aligned}
\tilde{B}_t=\frac{\gamma}{n}\sum_{i=1}^n\sum_{j=1}^\tau  \mathbf{g}_{t-\tau}^i{,} \quad \hat{\mathbf{x}}_{t+1}=\hat{\mathbf{x}}_t - \frac{\gamma}{n}\mathbf{g}_{t}^i.
\end{aligned} 
\end{equation}

Similar to Section~\ref{sec:proof2T1}, we have the following lemmas.


\begin{lemma}\label{lemma:dd-sgd}
        Let $\left\{\mathbf{x}_t, \hat{\mathbf{x}}_t,
        \tilde{B}_t\right\}_{t \geq 0}$ be defined in D-SGD with delayed aggregation (Eq.~\ref{dd-sgd}), with gradient oracle $\left\{\mathbf{g}_t\right\}_{t \geq 0}$ and let $f: \mathbb{R}^d \rightarrow \mathbb{R}$  be  $L$-smooth. If $\gamma \leq \frac{3}{2L}$, then we have

\begin{equation*}
\begin{aligned}
\mathbb{E} f\left(\hat{\mathbf{x}}_{t+1}\right) 
\leq  \mathbb{E} f\left(\hat{\mathbf{x}}_{t}\right)  - \frac{\gamma}{2} \mathbb{E} \left\|\nabla f\left(\mathrm{x}_t\right)\right\|^2  + 2\gamma L^2  \mathbb{E} \left\|\tilde{B}_t\right\|^2 +\frac{\gamma^2 L \sigma^2}{2 n}. 
\end{aligned}
\end{equation*}
\end{lemma}

\textit{Proof.} We begin with the definition of $\hat{\mathbf{x}}_{t+1}$ and have
\begin{equation}
\begin{aligned}
\label{DD-SGD:x_hat}
 \mathbb{E} f\left(\hat{\mathbf{x}}_{t+1}\right)
 &= \mathbb{E} f\left(\hat{\mathbf{x}}_{t} - \frac{\gamma}{n} \sum_{i=1}^n \mathbf{g}_t^i \right)  \\
& \leq  \mathbb{E} f\left(\hat{\mathbf{x}}_t\right)-
\underbrace{ \gamma \mathbb{E}\left\langle\nabla f\left(\hat{\mathbf{x}}_t\right), \mathbf{g}_t\right\rangle}_{:= A_1} +\frac{L\gamma^2}{2}\mathbb{E}\left\| \nabla f (\mathbf{x}_t) \right\|^2 + \frac{L \gamma^2 \sigma^2}{2 n}.
\end{aligned}
\end{equation}

We estimate $A_1$ as 
\begin{equation}
\begin{aligned}
-A_1
&=  \gamma ( \mathbb{E}\left\langle  \nabla f\left(\mathbf{x}_t\right) -\nabla f\left(\hat{\mathbf{x}}_t\right)  , \mathbf{g}_t\right\rangle -  \mathbb{E} \langle\nabla f\left(\mathbf{x}_t\right), \mathbf{g}_t\rangle ) \\
&\leq  \frac{c_1\gamma L^2}{2} \mathbb{E} \lVert \tilde{B}_t \rVert ^2 + \frac{\gamma}{2 c_1} \mathbb{E} \lVert \nabla f(\mathbf{x}_t) \rVert ^2 - \gamma \mathbb{E}  \lVert \nabla f(\mathbf{x}_t) \rVert ^2 \\
& \leq -\frac{7\gamma}{8} \mathbb{E} \lVert \nabla f(\mathbf{x}_t) \rVert ^2 + 2\gamma L^2 \mathbb{E} \lVert \tilde{B}_t \rVert^2.
\end{aligned}
\end{equation}

The second line is due to Jensen's inequality (\textit{i.e.}, Eq.~\ref{jensen_2}). We let $c_1=4$ in the third line. We bring  $A_1$ into  Eq.~\ref{DD-SGD:x_hat} and complete the proof.

\begin{theorem} [Non-convex convergence rate of D-SGD with delayed aggregation]
Let $f: \mathbb{R}^d \rightarrow \mathbb{R}$ be $L$-smooth. Then there exists a stepsize $\gamma \leq \frac{1}{4L\tau}$, such that at most
\begin{equation*}
\begin{split}
    \mathcal{O}(&\frac{\sigma^2}{n\epsilon^2}+\frac{ \sqrt{\frac{\tau}{n}\sigma^2}}{\epsilon^{3/2}}+\frac{\tau}{\epsilon})\cdot L (f(\mathbf{x}_0) - f^*)
\end{split}
\end{equation*}
iterations of D-SGD with delayed aggregation, it holds $\mathbb{E}\lVert \nabla f(\mathbf{x}_{out}) \rVert^2 \leq \epsilon $, and $\mathbf{x}_{out} = \mathbf{x}_t$ denotes an iterate $\mathbf{x}_t \in \left\{\mathbf{x}_0, \ldots, \mathbf{x}_{\mathbf{T}-1}\right\}$, where $\mathbf{T}$ denotes the total iterations, chosen at random uniformly.
\end{theorem}

\textit{Proof.}
Let $G_t:=\sum_{i=0}^t \mathbb{E} \left\|\nabla f\left(\mathrm{x}_i\right)\right\|^2$. According to above lemmas (Lemma~\ref{lemma:dd-sgd}, \ref{lemma2}), we have
\begin{equation*}
\begin{aligned}
  \frac{3\gamma}{8}G_T  \leq  \mathbb{E} f(\mathbf{x}_0) - f^* + \frac{\gamma^2 L (T+1) \sigma^2 }{2 n} 
  + \gamma^3 L^2 (T+1)  \frac{2\tau \sigma^2}{n}.
\end{aligned}
\end{equation*}

In this way, we have $\gamma\leq \min \{ \frac{3}{2L}, \frac{1}{4L\tau} \} \leq  \frac{1}{4L\tau}$. According to Lemma~\ref{O_nonconvex}, we let $B=3/8$, $C=\frac{L\sigma^2}{2 n}$, $D=\frac{L^2\tau\sigma^2}{n}$, $E= 4L\tau$, $r_0=\mathbb{E} f(\mathbf{x} _0) - f^*$, completing the proof.

\begin{lemma}\label{lemma:ddsgd2}
Let $\left\{\mathbf{x}_t, \hat{\mathbf{x}}_t,
        \tilde{B}_t\right\}_{t \geq 0}$ be defined in D-SGD with delayed aggregation (Eq.~\ref{dd-sgd}), with gradient oracle $\left\{\mathbf{g}_t\right\}_{t \geq 0}$ and let $f: \mathbb{R}^d \rightarrow \mathbb{R}$  be  $L$-smooth and $\mu$-strongly convex. If $\gamma \leq   \frac{1}{2L}$, then we have
        
\begin{equation*}
\begin{aligned}
    X_{t+1}&\leq (1-\frac{\gamma \mu}{2})X_t - \frac{\gamma}{2}F_t + \gamma^2 \sigma^2/n
     + 5 \gamma L \mathbb{E} \lVert
\tilde{B}_t \rVert ^2.
\end{aligned}
\end{equation*}
\end{lemma}

\textit{Proof.}

\begin{equation*}
\begin{aligned}
    X_{t+1}& = X_{t} + \gamma^2 \lVert \frac{1}{n} \sum_{i=1}^n \mathbf{g}_t^i \rVert^2 - \underbrace{ 2 \gamma \mathbb{E}\left\langle \hat{\mathbf{x}}_t -\mathbf{x}^*, \mathbf{g}_t\right\rangle}_{:= A_2}.
\end{aligned}
\end{equation*}

We estimate the term $A_2$ as 
\begin{equation*}
\begin{aligned}
\label{A-2}
-A_2&
 =  - 2\gamma (\mathbb{E}\left\langle  \nabla f\left(\mathbf{x}_t\right), \mathbf{x}_t-\mathbf{x}^*\right\rangle  - \ \mathbb{E}\left\langle  \nabla f\left(\mathbf{x}_t\right), \mathbf{x}_t-\hat{\mathbf{x}}_t\right\rangle ) \\
&\leq -\gamma \mu \mathbb{E} \lVert \mathbf{x}_t - \mathbf{x}^* \rVert ^ 2 - 2\gamma F_t + \frac{1}{2}\gamma F_t  + 4\gamma L \mathbb{E} \lVert
\tilde{B}_t \rVert ^2 \\
& \leq -\frac{\gamma \mu}{2} X_t -\frac{3}{2}\gamma F_t + \gamma(4L+\mu) \mathbb{E} \lVert
\tilde{B}_t \rVert ^2.
\end{aligned}
\end{equation*}

The second line is due to the fact that $f: \mathbb{R}^d \rightarrow \mathbb{R}$ is $L$-smooth and $\mu$-strongly convex. Then choosing $\gamma \leq \frac{1}{2L}, \mu \leq L$, we have 
\begin{equation*}
\begin{aligned}
    X_{t+1}\leq  (1-\frac{\gamma \mu }{2})X_{t} - \frac{\gamma}{2} F_t + 5L\gamma \mathbb{E} \lVert
\tilde{B}_t \rVert ^2 + \gamma ^2 \sigma^2/n.
\end{aligned}
\end{equation*}

This completes the proof.

\begin{theorem} [Convex convergence rate of D-SGD with delayed aggregation]
Let $f: \mathbb{R}^d \rightarrow \mathbb{R}$ be $L$-smooth and $\mu$-strongly convex. Then there exists a stepsize $\gamma \leq   \frac{1}{8\sqrt{2}\ L\tau }$, such that at most
\begin{equation*}
\begin{split}
    \mathcal{O}(&\frac{\sigma^2}{n\mu\epsilon}+\frac{ \sqrt{L\frac{\tau}{n}\sigma^2}}{\mu \epsilon^{1/2}}+\frac{L\tau}{\mu})
\end{split}
\end{equation*}
iterations of D-SGD with delayed aggregation, it holds $\mathbb{E}f(\mathbf{x}_{out})-f^* \leq \epsilon $, and $\mathbf{x}_{out} = \mathbf{x}_t$ denotes an iterate $\mathbf{x}_t \in \left\{\mathbf{x}_0, \ldots, \mathbf{x}_{T-1}\right\}$,  selected probabilistically based on $(1-\frac{\mu \gamma}{2})^{-t}$.
\end{theorem}

\textit{Proof.}
By substituting Lemma~\ref{tildeB_nonconvex} into Lemma~\ref{lemma:ddsgd2}, we can get 
\begin{equation*}
\begin{aligned}
& w_T X_{T+1} \leq X_0 - \frac{\gamma}{2}\sum_{t=0}^T w_t F_t + \frac{W_T \gamma^2 \sigma^2}{n}  + 5\gamma L \sum_{t=0}^T w_t \mathbb{E} \lVert
\tilde{B}_t \rVert ^2 \\
&\leq X_0 - \frac{11\gamma}{32}\sum_{t=0}^T w_t F_t + \frac{W_T \gamma^2 \sigma^2}{n}  + 5LW_T\gamma^3\frac{\tau \sigma^2}{n}.
\end{aligned}
\end{equation*}

According to Lemma~\ref{O_convex}, we complete the proof.

\subsection{Proof of Theorem 3}

\noindent\textbf{Restatement of Theorem 3 (Estimation of $T_{\text{avg}}$ in \algoname).}
Under \textbf{Assumptions 5} and  \textbf{6},
the average iteration time $T_{\text{avg}}$ satisfies the following approximation:           
\begin{equation}
T_{\text{avg}}\approx\max{\left \{  \frac{T_{\text{comp}}+b+\delta \cdot V_g/\bar{a}}{\tau+1} ,\frac{\delta \cdot V_g}{\bar{a}},T_{\text{comp}} \right \}},
\end{equation}
where the approximation error arises from pipeline boundary effects and can be neglected when $E \gg \tau$.

\textit{Proof.}
Based on the assumption,  the iterative formula of $TS_k$ derived from the relationship between computation and communication can be written as :
\begin{equation}
\label{eq:itera_formula}
    \left\{\begin{matrix} 
  TC_k=TM_k+b \\  
 TS_{k+1}=T_{\text{comp}}+max\left\{ TC_{k-\tau},TS_k \right\} \\
 TM_{k+1}=\frac{\delta V_g}{\bar{a}}+max\left\{ TM_k,TS_{k+1} \right\}
\end{matrix}\right. \quad , k=0,1,2... \quad .
\end{equation}  
with initial values $TS_0 = TM_0 = 0$ and $TC_k = 0$ for all $k \leq 0$.
The analysis is divided into four cases, each corresponding to a specific regime determined by the relative magnitudes of $T_{\text{comp}}$, $\frac{\delta V_g}{\bar{a}}$, and $\tau$.

\subsection*{Case 1: Computation-Dominated Regime}
Consider the regime where $T_{\text{comp}} > \frac{\delta V_g}{\bar{a}}$ and $\tau T_{\text{comp}} > \frac{\delta V_g}{\bar{a}} + b$. We claim that:
\[
\forall k \in \mathbb{N}^+, \quad TS_k = k T_{\text{comp}}.
\]

\noindent To verify this, we define the inductive hypothesis:
\[
P(k): \quad \left\{
\begin{aligned}
TS_i &= i T_{\text{comp}}, \\
TM_i &= \frac{\delta V_g}{\bar{a}} + i T_{\text{comp}},
\end{aligned}
\right. \quad \forall i = 1, \ldots, k.
\]

\noindent The base case $P(1)$ can be directly verified from the recurrence relations. Now assume $P(k)$ holds for some $k \in \mathbb{N}^+$. We analyze the $(k+1)$-th step:

\begin{itemize}
    \item If $k \le \tau$, then $TC_{k-\tau} = 0 \le TS_k$;
    \item Otherwise, by the inductive assumption, we have:
    \[
    TC_{k-\tau} = (k - \tau) T_{\text{comp}} + \frac{\delta V_g}{\bar{a}} + b \le k T_{\text{comp}} = TS_k.
    \]
\end{itemize}

\noindent Thus, it follows that:
\[
TS_{k+1} = T_{\text{comp}} + \max\{TC_{k-\tau}, TS_k\} = T_{\text{comp}} + TS_k = (k + 1) T_{\text{comp}}.
\]

\noindent Moreover,
\[
TM_{k+1} = \frac{\delta V_g}{\bar{a}} + \max\{TM_k, TS_{k+1}\} = \frac{\delta V_g}{\bar{a}} + TS_{k+1},
\]
since $TM_k \le TS_{k+1}$ by the inductive assumption. Therefore, $P(k+1)$ holds, completing the induction.

\noindent It follows that:
\[
TC_E = TM_E + b = \frac{\delta V_g}{\bar{a}} + E \cdot T_{\text{comp}} + b,
\]
and thus the approximation error satisfies:
\[
\left| TC_E - E \cdot T_{\text{comp}} \right| = \frac{\delta V_g}{\bar{a}} + b \le b + \min\left\{T_{\text{comp}}, \frac{\delta V_g}{\bar{a}}\right\}.
\]

\subsection*{Case 2: Communication-Dominated Regime}
Now consider the case where $\frac{\delta V_g}{\bar{a}} > T_{\text{comp}}$ and $\tau \cdot \frac{\delta V_g}{\bar{a}} > T_{\text{comp}} + b$. We claim that there exists $N_1 \in \mathbb{N}^+$ such that:
\[
\forall k > N_1, \quad TS_{k+1} - TS_k = \frac{\delta V_g}{\bar{a}}.
\]

\noindent To establish this, we define the inductive statement:
\[
P(k): \quad
\left\{
\begin{aligned}
TM_i &= T_{\text{comp}} + i \cdot \frac{\delta V_g}{\bar{a}}, \\
TM_i &\ge TS_{i+1},
\end{aligned}
\right.
\quad \forall i = 1, \ldots, k.
\]

\noindent For $i = 1, \ldots, \tau+1$, direct computation gives:
\[
TS_i = i T_{\text{comp}}, \quad TM_i = T_{\text{comp}} + i \cdot \frac{\delta V_g}{\bar{a}},
\]
hence $P(\tau)$ holds. Assume $P(k)$ is true for some $k \ge \tau$.

\noindent Then:
\[
TM_{k+1} = \frac{\delta V_g}{\bar{a}} + TM_k = T_{\text{comp}} + (k+1) \cdot \frac{\delta V_g}{\bar{a}}.
\]

\noindent For the next computation step:
\[
\begin{aligned}
TC_{k+1-\tau} + T_{\text{comp}} &= (k+1 - \tau) \cdot \frac{\delta V_g}{\bar{a}} + 2 T_{\text{comp}} + b \\
&\le (k+1) \cdot \frac{\delta V_g}{\bar{a}} + T_{\text{comp}} \le TM_{k+1},
\end{aligned}
\]
\[
TS_{k+1} + T_{\text{comp}} \le TM_k + T_{\text{comp}} \le TM_{k+1}.
\]
Therefore,
\[
TS_{k+2} = T_{\text{comp}} + \max\left\{TC_{k+1 - \tau}, TS_{k+1} \right\} \le TM_{k+1},
\]
which implies $P(k+1)$ holds. By induction, $P(k)$ is valid for all $k \ge \tau$.

\noindent Consequently, for all sufficiently large $k$, the recurrence stabilizes and we obtain:
\[
TS_k = T_{\text{comp}} + TC_{k-\tau} = 2 T_{\text{comp}} + b + (k - \tau) \cdot \frac{\delta V_g}{\bar{a}},
\]
with:
\[
\left| TC_E - E \cdot \frac{\delta V_g}{\bar{a}} \right| = T_{\text{comp}} + b \le b + \min\left\{T_{\text{comp}}, \frac{\delta V_g}{\bar{a}} \right\}.
\]

\noindent Finally, we verify the existence of $N_1 > \tau$ such that $TC_{N_1 - \tau} \ge TS_{N_1}$. If such $N_1$ does not exist, then both $\{TC_{k - \tau}\}$ and $\{TS_k\}$ would form arithmetic sequences with conflicting common differences, leading to a contradiction.

\subsection*{Case 3: Periodic Structure under Intermediate Delays}

Consider the case where the computation time satisfies $T_{\text{comp}} > \frac{\delta V_g}{\bar{a}}$ while the aggregated period satisfies $\tau T_{\text{comp}} \le \frac{\delta V_g}{\bar{a}} + b$. Under this condition, we can establish the existence of a positive integer $N$ such that the sequence $\left\{TS_{k+1} - TS_k\right\}_{k=N}^{T-1}$ becomes $(\tau+1)$-periodic. Specifically, for all $k \ge N$ and any $d \in \{1, \dots, \tau+1\}$, we have:
\begin{equation*}
    TS_{(k+1)(\tau+1)+d} - TS_{k(\tau+1)+d} = T_{\text{comp}} + b + \frac{\delta V_g}{\bar{a}}.
\end{equation*}

This periodic property can be verified by explicitly analyzing the first $(2\tau+2)$ terms of the sequence $\{(TS_k, TM_k)\}$. We have:
\begin{equation}
\label{eq:case3_terms}
\begin{aligned}
    &\left\{
        \begin{aligned}
            TS_i &= i T_{\text{comp}}, \\
            TM_i &= \frac{\delta V_g}{\bar{a}} + i T_{\text{comp}},
        \end{aligned}
    \right. \quad \text{for } i = 1, \dots, \tau+1; \\
    &\\
    &\left\{
        \begin{aligned}
            TS_i &= \frac{\delta V_g}{\bar{a}} + b + (i - \tau) T_{\text{comp}}, \\
            TM_i &= 2 \frac{\delta V_g}{\bar{a}} + b + (i - \tau) T_{\text{comp}},
        \end{aligned}
    \right. \quad \text{for } i = \tau+2, \dots, 2\tau+2.
\end{aligned}
\end{equation}

Let us define the index group:
\begin{equation*}
\begin{aligned}
    A_k &:= \left\{(TM_{k(\tau+1)+d}, TS_{k(\tau+1)+d}) \mid d = 1, \dots, \tau+1 \right\}, \\
    A_k + a &:= \left\{(TM_{k(\tau+1)+d} + a, TS_{k(\tau+1)+d} + a) \mid d = 1, \dots, \tau+1 \right\}, \quad \forall a \in \mathbb{R}.
\end{aligned}
\end{equation*}

By rewriting the iterative update rule in the compact form $A_k = F(A_{k-1})$, it follows from translation invariance that:
\[
    A_k + a = F(A_{k-1} + a).
\]

Now define the shifted sequence $A_k' := A_{k+1} - \left(T_{\text{comp}} + b + \frac{\delta V_g}{\bar{a}} \right)$. Using Eq.~\eqref{eq:case3_terms}, we obtain:
\begin{equation*}
\begin{aligned}
    A_0' &= A_1 - \left(T_{\text{comp}} + b + \frac{\delta V_g}{\bar{a}}\right) = A_0, \\
    A_1' &= F(A_0') = A_1.
\end{aligned}
\end{equation*}
Therefore, we conclude:
\begin{equation*}
\begin{aligned}
    A_2 &= A_1' + \left(T_{\text{comp}} + b + \frac{\delta V_g}{\bar{a}}\right) = A_1 + \left(T_{\text{comp}} + b + \frac{\delta V_g}{\bar{a}}\right),
\end{aligned}
\end{equation*}
and recursively,
\begin{equation*}
    A_{k+1} = A_k + \left(T_{\text{comp}} + b + \frac{\delta V_g}{\bar{a}}\right), \quad \forall k \in \mathbb{N}^*.
\end{equation*}

Let $E = m(\tau+1) + d$ for $d = 1, \dots, \tau+1$, then the cumulative time $TC_E$ satisfies:
\begin{equation*}
\begin{aligned}
    TC_E &= m \left(T_{\text{comp}} + b + \frac{\delta V_g}{\bar{a}} \right) + d T_{\text{comp}} + \frac{\delta V_g}{\bar{a}} + b, \\
    \left| TC_E - E \left( \frac{T_{\text{comp}} + b + \frac{\delta V_g}{\bar{a}}}{\tau+1} \right) \right|
    &= d \left( T_{\text{comp}} - \frac{T_{\text{comp}} + b + \frac{\delta V_g}{\bar{a}}}{\tau+1} \right) + \frac{\delta V_g}{\bar{a}} + b \\
    &\le \tau T_{\text{comp}} \le b + \min\left\{T_{\text{comp}}, \frac{\delta V_g}{\bar{a}} \right\}.
\end{aligned}
\end{equation*}
This confirms that $TC_E$ deviates from its average rate by at most a bounded constant, further supporting the regularity and predictability of the update pattern in this intermediate-delay regime.

\subsection*{Case 4: Periodic Structure under Computation-Dominated Regime}

In this case, we consider the regime where $T_{\text{comp}} < \frac{\delta V_g}{\bar{a}}$ and $\tau \frac{\delta V_g}{\bar{a}} \le T_{\text{comp}} + b$. Under these conditions, we can establish that there exists a positive integer $N$ such that the update interval sequence $\left\{TS_{k+1} - TS_k\right\}_{k=N}^{T-1}$ exhibits $(\tau+1)$-periodicity. In particular, for all $k \ge N$ and for all $d \in \{1, \dots, \tau+1\}$, the following recurrence holds:
\begin{equation*}
    TS_{(k+1)(\tau+1)+d} - TS_{k(\tau+1)+d} = T_{\text{comp}} + b + \frac{\delta V_g}{\bar{a}}.
\end{equation*}

Analogously to Case 3, we examine the first $(3\tau + 3)$ elements of the sequence $\left\{(TS_k, TM_k)\right\}$ to validate the periodic structure:
\begin{equation*}
\begin{aligned}
    &\left\{
        \begin{aligned}
            TS_i &= i T_{\text{comp}}, \\
            TM_i &= i \frac{\delta V_g}{\bar{a}} + T_{\text{comp}},
        \end{aligned}
    \right. \quad &i = 1, \dots, \tau+1; \\
    &\\
    &\left\{
        \begin{aligned}
            TS_i &= (b + T_{\text{comp}} + \tfrac{\delta V_g}{\bar{a}}) + T_{\text{comp}} + (i - \tau - 2) \tfrac{\delta V_g}{\bar{a}}, \\
            TM_i &= (b + T_{\text{comp}} + \tfrac{\delta V_g}{\bar{a}}) + T_{\text{comp}} + (i - \tau - 1) \tfrac{\delta V_g}{\bar{a}},
        \end{aligned}
    \right. \quad &i = \tau+2, \dots, 2\tau+2; \\
    &\\
    &\left\{
        \begin{aligned}
            TS_i &= 2(b + T_{\text{comp}} + \tfrac{\delta V_g}{\bar{a}}) + T_{\text{comp}} + (i - 2\tau - 3) \tfrac{\delta V_g}{\bar{a}}, \\
            TM_i &= 2(b + T_{\text{comp}} + \tfrac{\delta V_g}{\bar{a}}) + T_{\text{comp}} + (i - 2\tau - 2) \tfrac{\delta V_g}{\bar{a}},
        \end{aligned}
    \right. \quad &i = 2\tau+3, \dots, 3\tau+3.
\end{aligned}
\end{equation*}

We define the grouping $A_k$ as in Case 3. From the recurrence observed above, we find that:
\[
    A_2 = A_1 + \left( T_{\text{comp}} + \frac{\delta V_g}{\bar{a}} + b \right),
\]
and by translation invariance of the update dynamics, it follows that:
\[
    A_{k+2} = A_{k+1} + \left( T_{\text{comp}} + b + \frac{\delta V_g}{\bar{a}} \right), \quad \forall k \in \mathbb{N}^*.
\]
Thus, the $(\tau+1)$-periodicity is established, completing the proof for Case 4.

Now, let $E = m(\tau + 1) + d$ with $d \in \{1, \dots, \tau + 1\}$. The cumulative computation time $TC_E$ satisfies:
\begin{equation*}
\begin{aligned}
    TC_E &= m \left(T_{\text{comp}} + b + \frac{\delta V_g}{\bar{a}} \right) + d \cdot \frac{\delta V_g}{\bar{a}} + T_{\text{comp}} + b, \\
    \left| TC_E - E \cdot \frac{T_{\text{comp}} + b + \frac{\delta V_g}{\bar{a}}}{\tau + 1} \right|
    &= d \left( \frac{\delta V_g}{\bar{a}} - \frac{T_{\text{comp}} + b + \frac{\delta V_g}{\bar{a}}}{\tau + 1} \right) + T_{\text{comp}} + b \\
    &\le \tau \cdot \frac{\delta V_g}{\bar{a}} \le b + \min\left\{ T_{\text{comp}}, \frac{\delta V_g}{\bar{a}} \right\}.
\end{aligned}
\end{equation*}

\subsection*{Summary of All Cases and Average Iteration Time}

By integrating the results from all four cases, we obtain a unified expression for the asymptotic average iteration time:
\begin{equation*}
\begin{aligned}
    T'_{\text{avg}} &=
    \begin{cases}
        \dfrac{T_{\text{comp}} + \frac{\delta V_g}{\bar{a}} + b}{\tau + 1}, & \tau < \tau_{\text{thres}}, \\
        \max\left\{ T_{\text{comp}}, \frac{\delta V_g}{\bar{a}} \right\}, & \tau \ge \tau_{\text{thres}},
    \end{cases} \quad
    \text{with}  \quad \left| TC_E - E\cdot  T'_{\text{avg}} \right| \le b + \min\left\{ T_{\text{comp}}, \frac{\delta V_g}{\bar{a}} \right\},
\end{aligned}
\end{equation*}
where $T'_{\text{avg}} := \lim\limits_{k \to \infty} \frac{TS_k}{k}$ approximates the true average iteration time $T_{\text{avg}}$, and
\begin{equation*}
    \tau_{\text{thres}} = \min \left\{ \frac{\frac{\delta V_g}{\bar{a}} + b}{T_{\text{comp}}}, \frac{T_{\text{comp}} + b}{\frac{\delta V_g}{\bar{a}}} \right\}.
\end{equation*}

As a final conclusion, we estimate the average iteration time by:
\begin{equation*}
    T_{\text{avg}} \approx T'_{\text{avg}} = \max \left\{ \frac{T_{\text{comp}} + b + \frac{\delta V_g}{\bar{a}}}{\tau + 1}, \frac{\delta V_g}{\bar{a}}, T_{\text{comp}} \right\},
\end{equation*}
with the absolute error bounded by:
\begin{equation*}
    \left| T_{\text{avg}} - T'_{\text{avg}} \right| \le \frac{b + \min\left\{ T_{\text{comp}}, \frac{\delta V_g}{\bar{a}} \right\}}{E} = O\left( \frac{1}{E} \right).
\end{equation*}

\section{Addendum to Evaluation Experiments}
\label{appendixC}

\subsection{Compression Ratio and Staleness Curves in Section~\ref{exp.2}}
\label{appendixD}

\begin{figure}[h]
    \centering

    \begin{minipage}{0.32\linewidth}
        \centering
        \includegraphics[width=\linewidth]{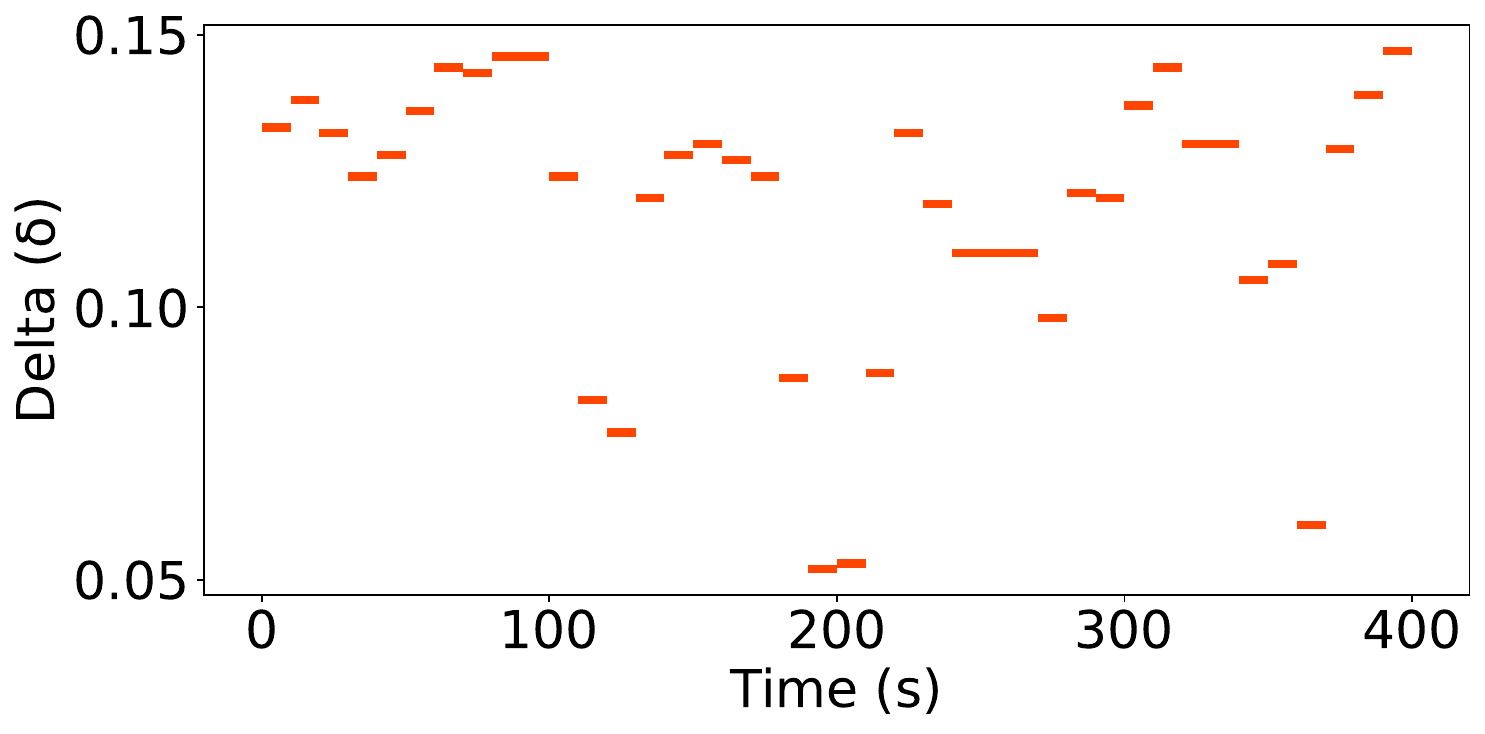}
        \caption*{(a) ResNet@CIFAR-10}
    \end{minipage}
    \hfill
    \begin{minipage}{0.32\linewidth}
        \centering
        \includegraphics[width=\linewidth]{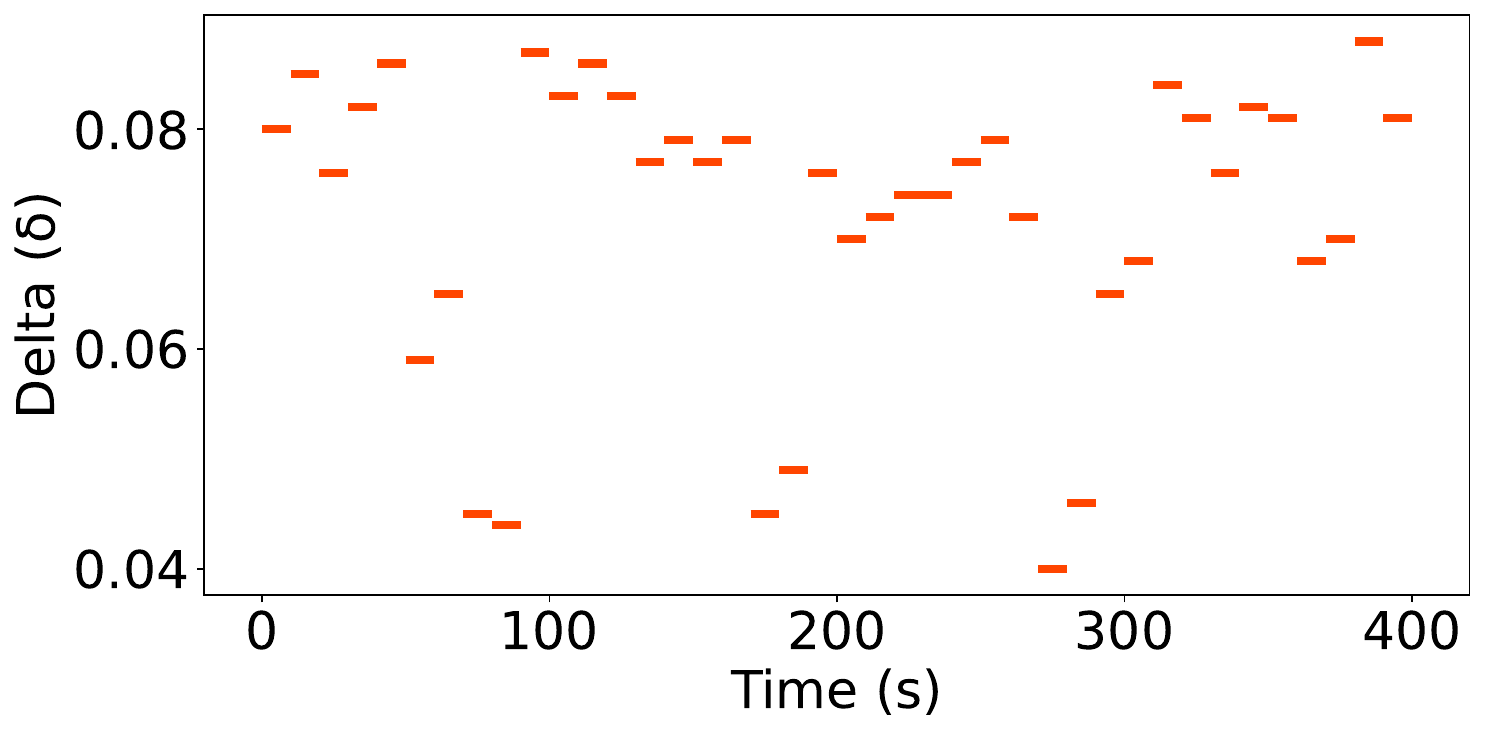}
        \caption*{(b) ViT@ImageNet}
    \end{minipage}
    \hfill
    \begin{minipage}{0.32\linewidth}
        \centering
        \includegraphics[width=\linewidth]{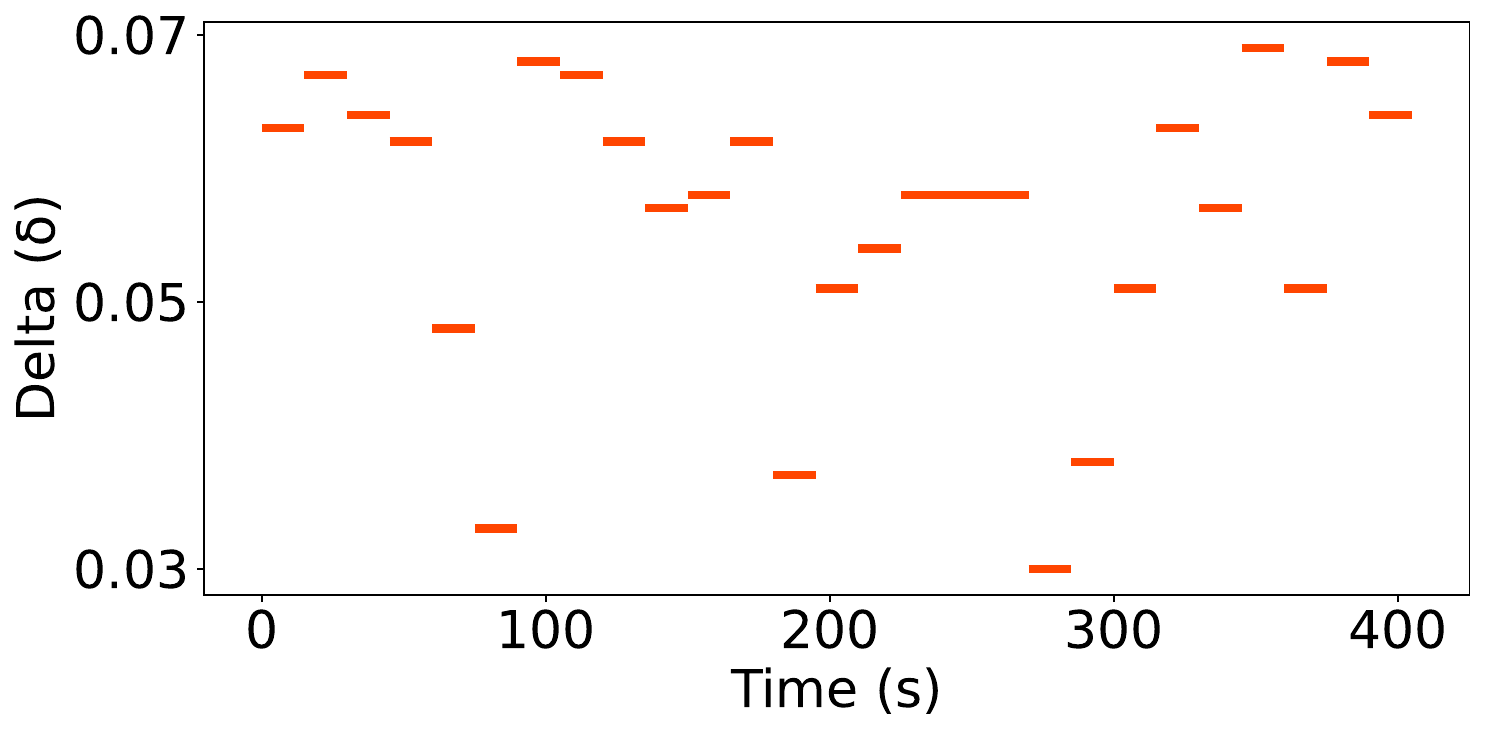}
        \caption*{(c) GPT@Wikitext}
    \end{minipage}

    \caption{$\delta$ varying over time for tasks: ResNet@CIFAR-10, ViT@ImageNet, and GPT@Wikitext, with $\tau=2$.}
    \label{fig:network}
\end{figure}

As shown in Fig.~\ref{fig:network}, the network conditions exhibit dynamic bandwidth fluctuations across all tasks (CNN@FMNIST, CNN@CIFAR-10, ViT@ImageNet, and GPT@Wikitext), and the compression ratio $\delta$ is adaptively adjusted over time under a fixed 200\,ms transmission delay with $\tau = 2$.

\subsection{Detailed Curves of Section~\ref{exp.3} to Verify the Scalability of \sysname}

\begin{figure}[h]
\centering
\begin{minipage}{0.23\linewidth}
    \centering
    \includegraphics[width=\linewidth]{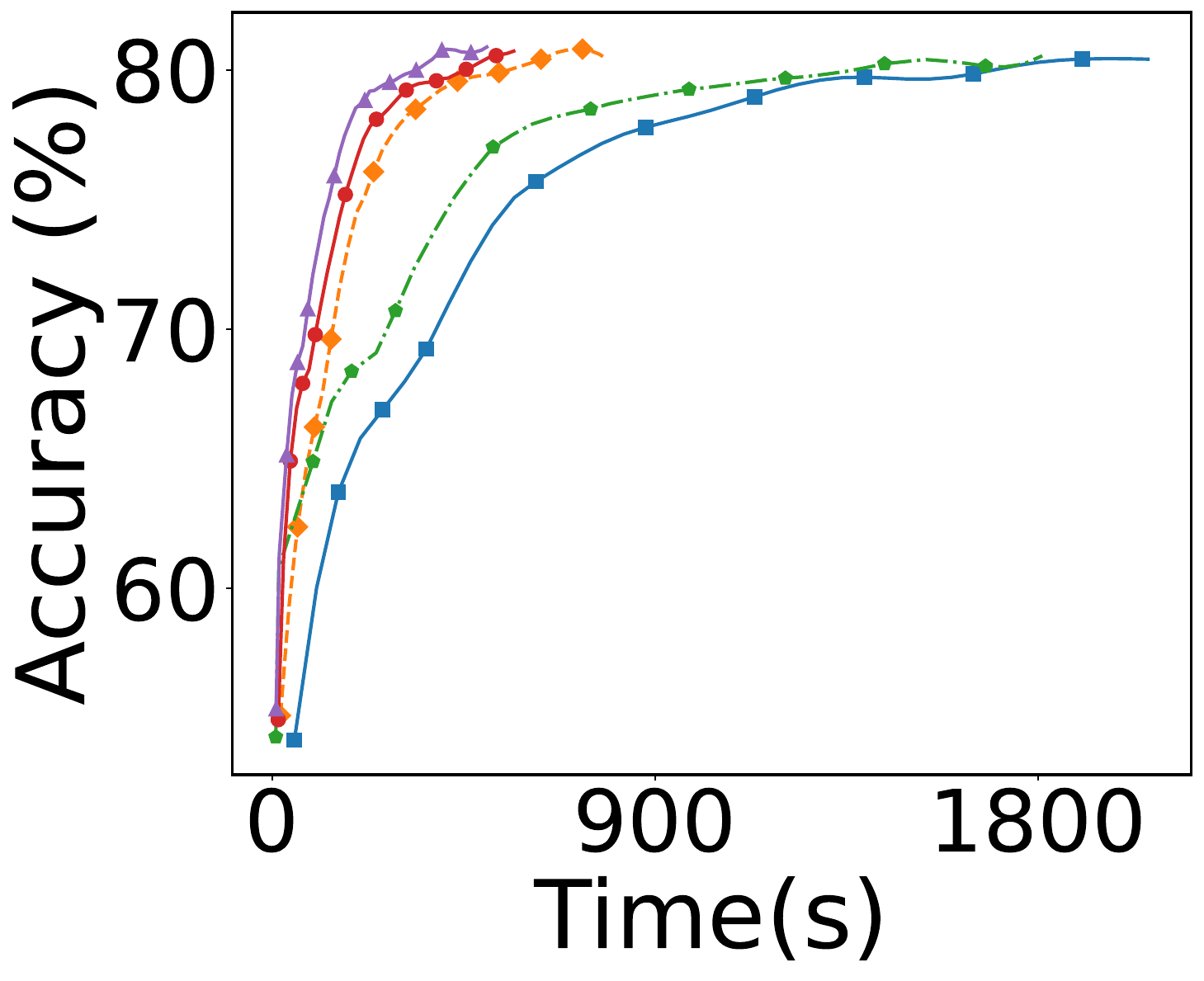}
    \caption*{\shortstack{(a) ViT@ImageNet \\ $n=4$}}
\end{minipage}
\hfill
\begin{minipage}{0.23\linewidth}
    \centering
    \includegraphics[width=\linewidth]{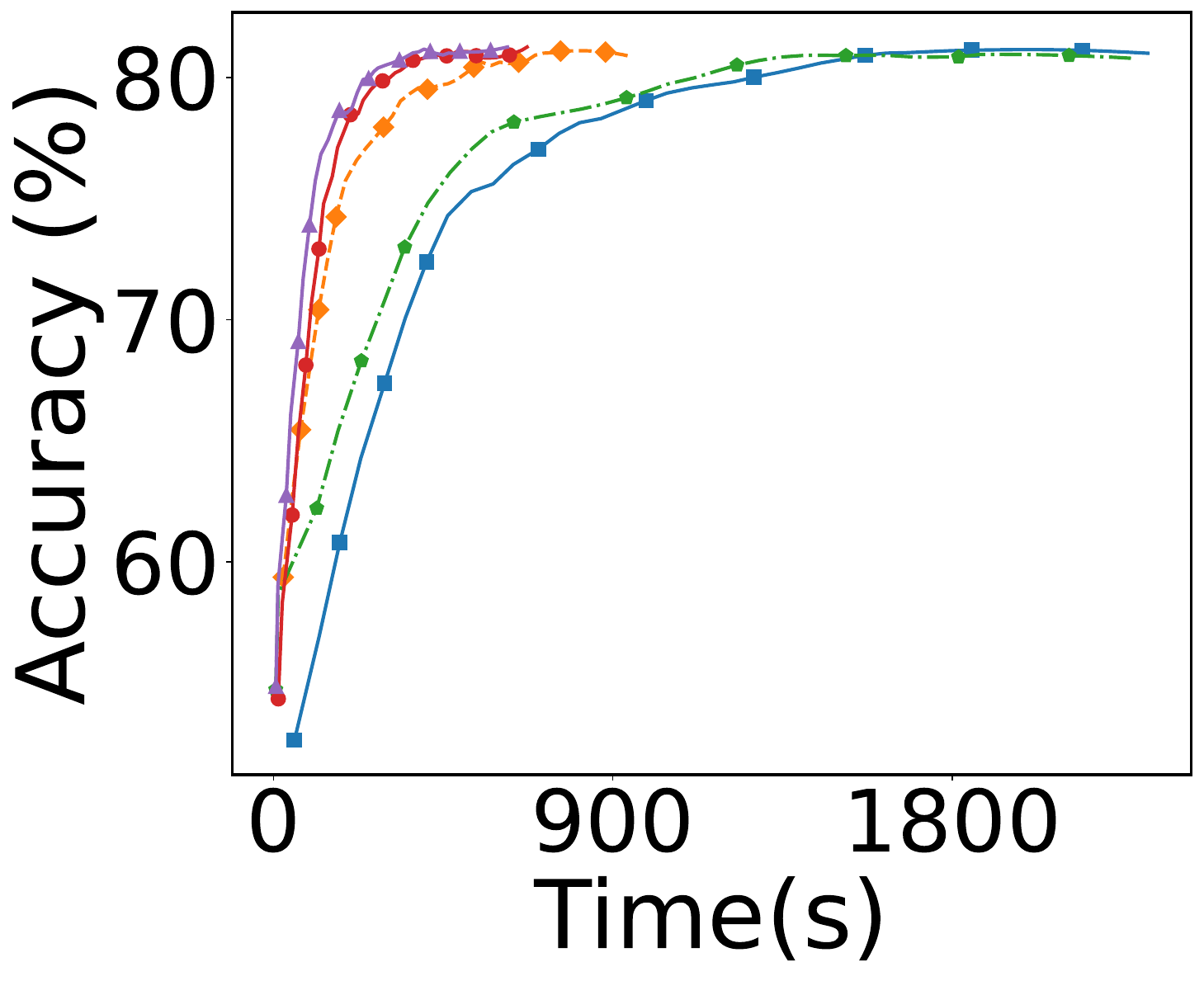}
    \caption*{\shortstack{(b) ViT@ImageNet \\ $n=8$}}
\end{minipage}
\hfill
\begin{minipage}{0.23\linewidth}
    \centering
    \includegraphics[width=\linewidth]{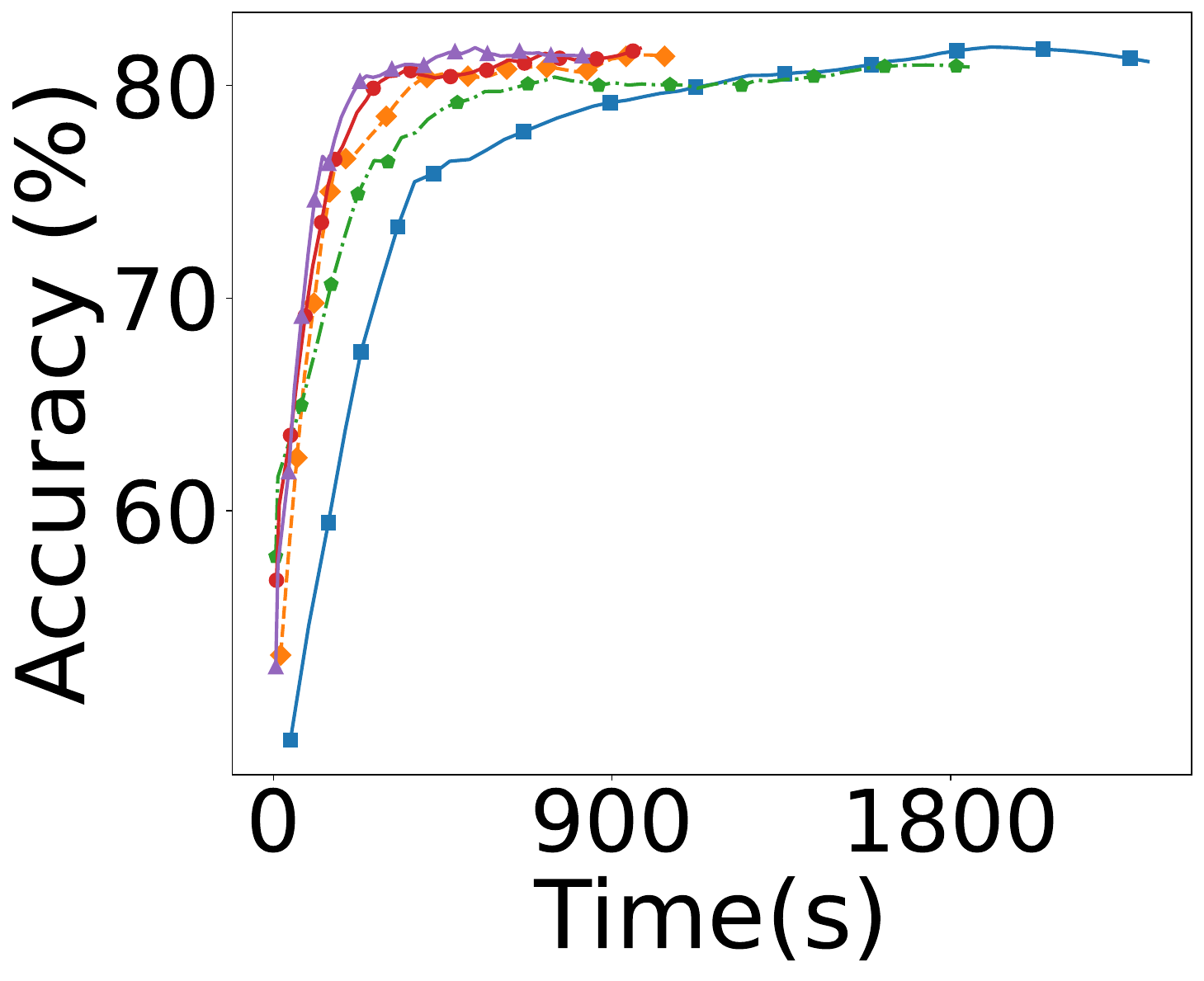}
    \caption*{\shortstack{(c) ViT@ImageNet \\ $n=16$}}
\end{minipage}
\hfill
\begin{minipage}{0.23\linewidth}
    \centering
    \includegraphics[width=\linewidth]{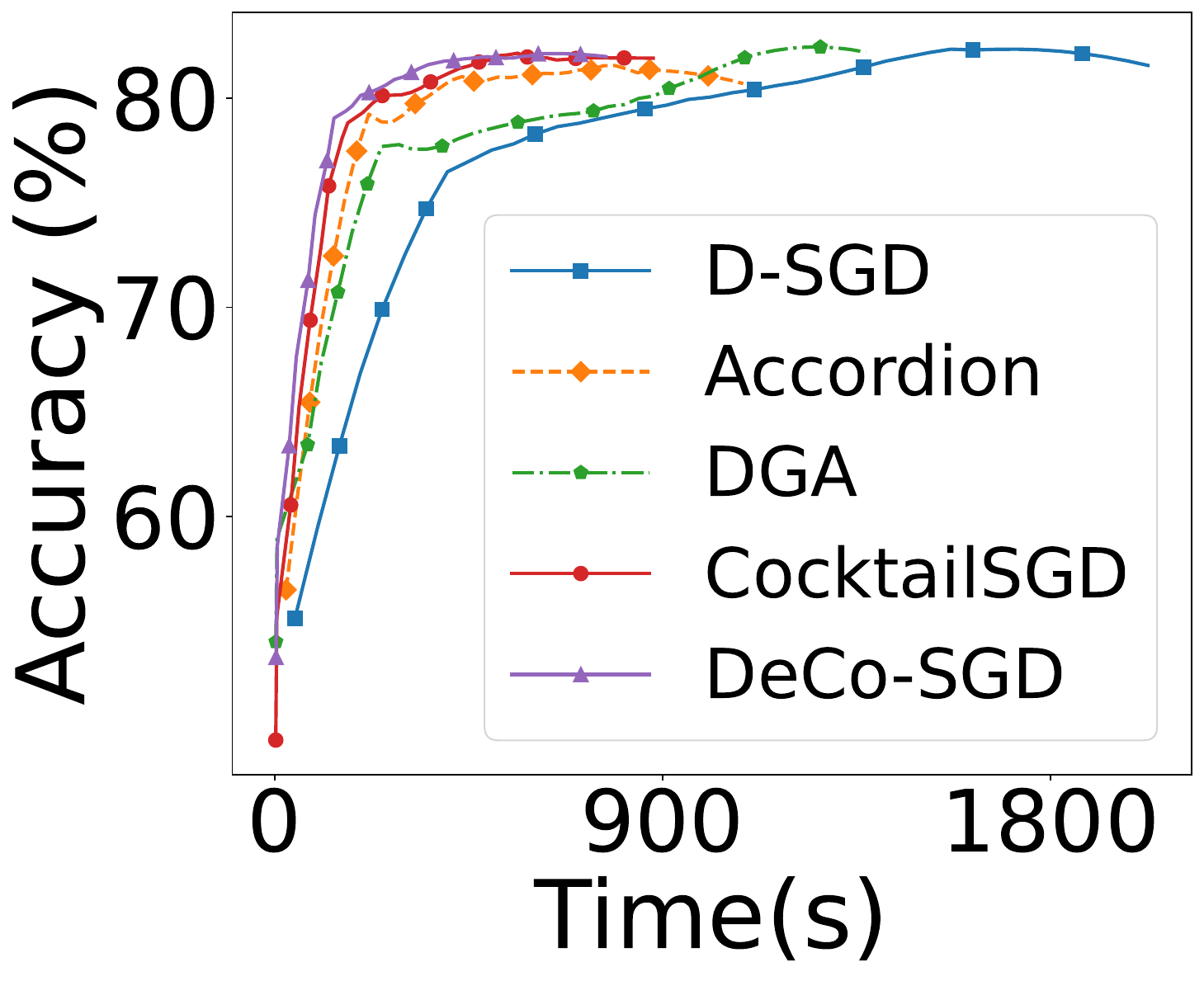}
    \caption*{\shortstack{(d) ViT@ImageNet \\ $n=32$}}
\end{minipage}

\vspace{2mm} 

\begin{minipage}{0.23\linewidth}
    \centering
    \includegraphics[width=\linewidth]{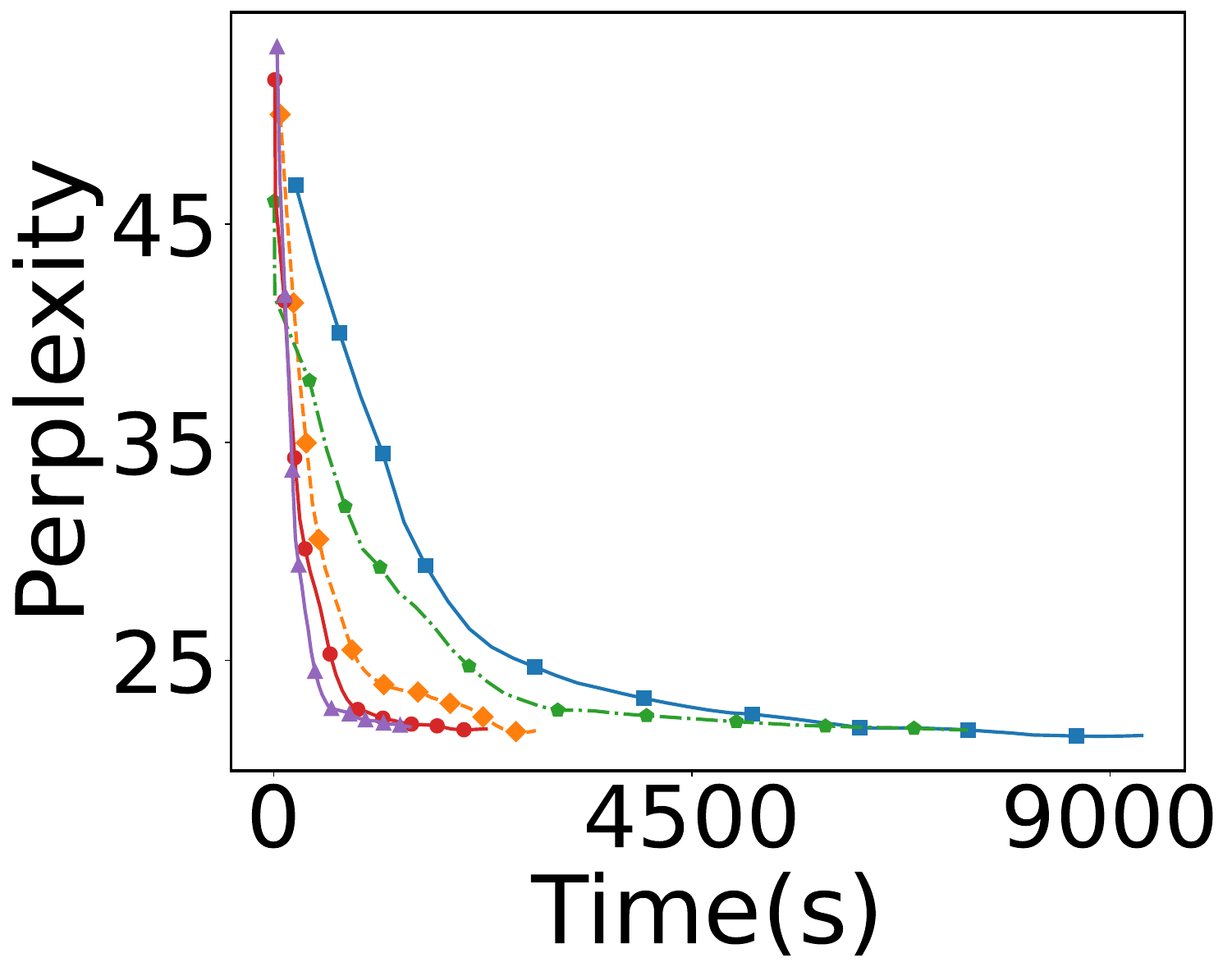}
    \caption*{\shortstack{(e) GPT@Wikitext \\ $n=4$}}
\end{minipage}
\hfill
\begin{minipage}{0.23\linewidth}
    \centering
    \includegraphics[width=\linewidth]{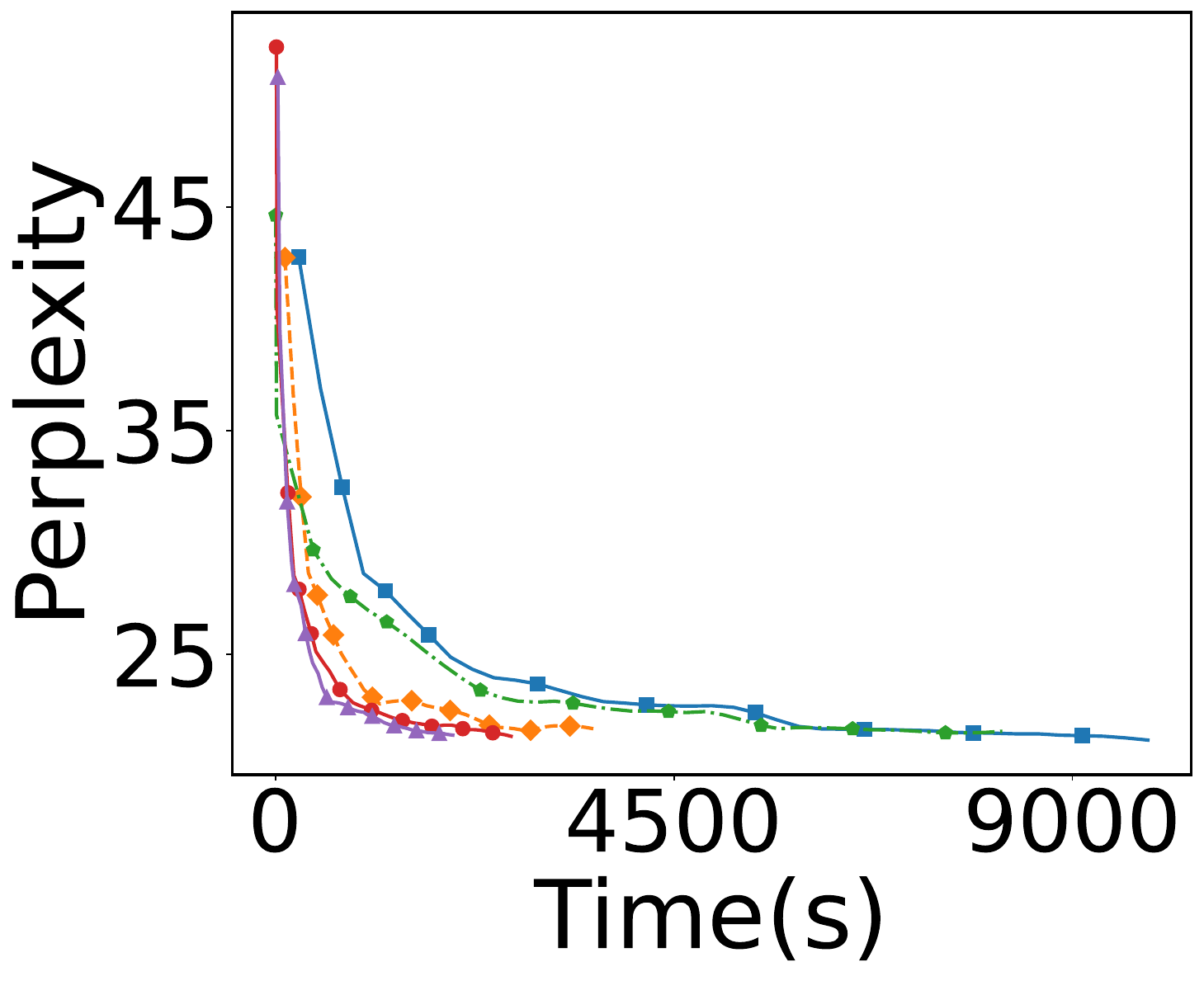}
    \caption*{\shortstack{(f) GPT@Wikitext \\ $n=8$}}
\end{minipage}
\hfill
\begin{minipage}{0.23\linewidth}
    \centering
    \includegraphics[width=\linewidth]{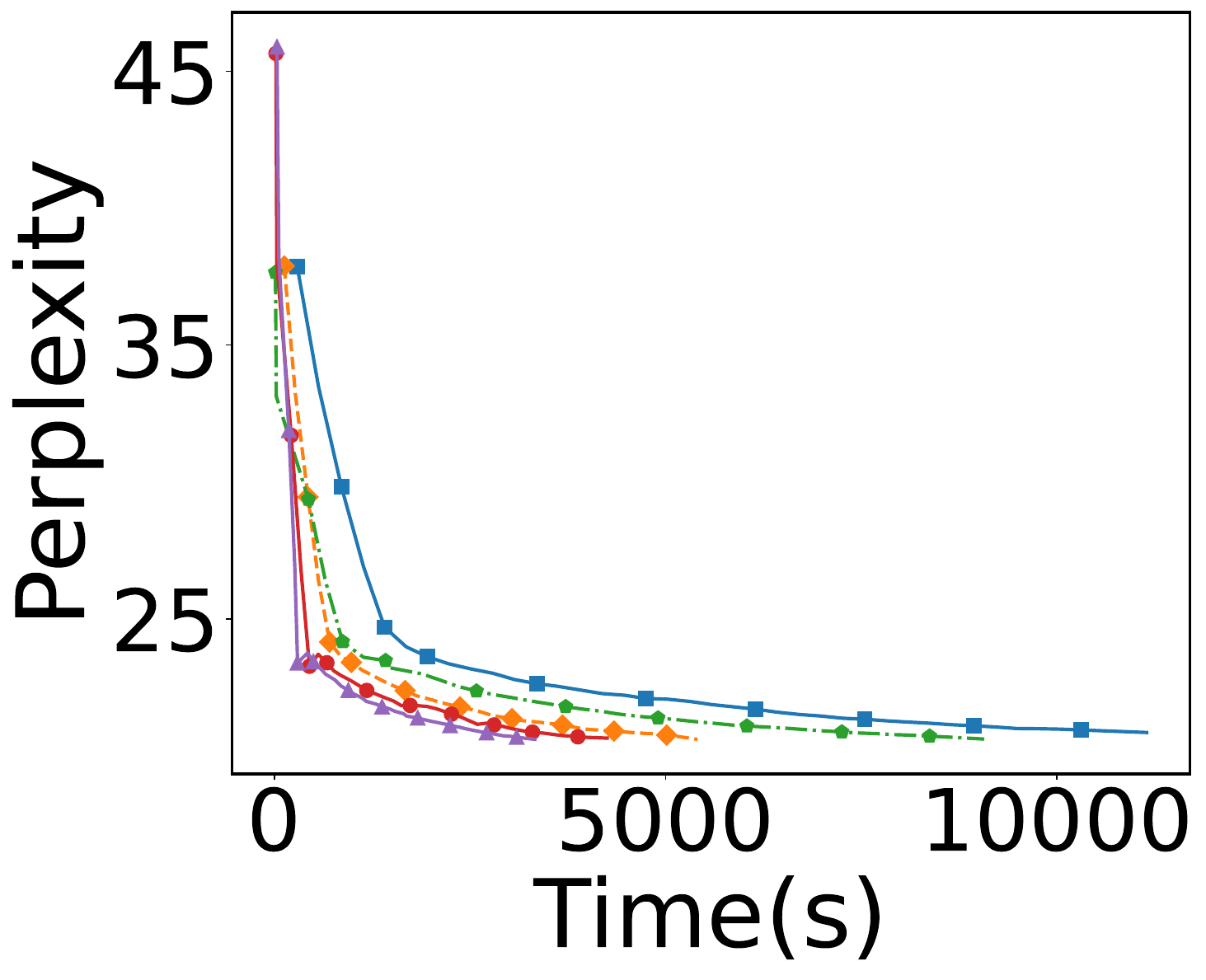}
    \caption*{\shortstack{(g) GPT@Wikitext \\ $n=16$}}
\end{minipage}
\hfill
\begin{minipage}{0.23\linewidth}
    \centering
    \includegraphics[width=\linewidth]{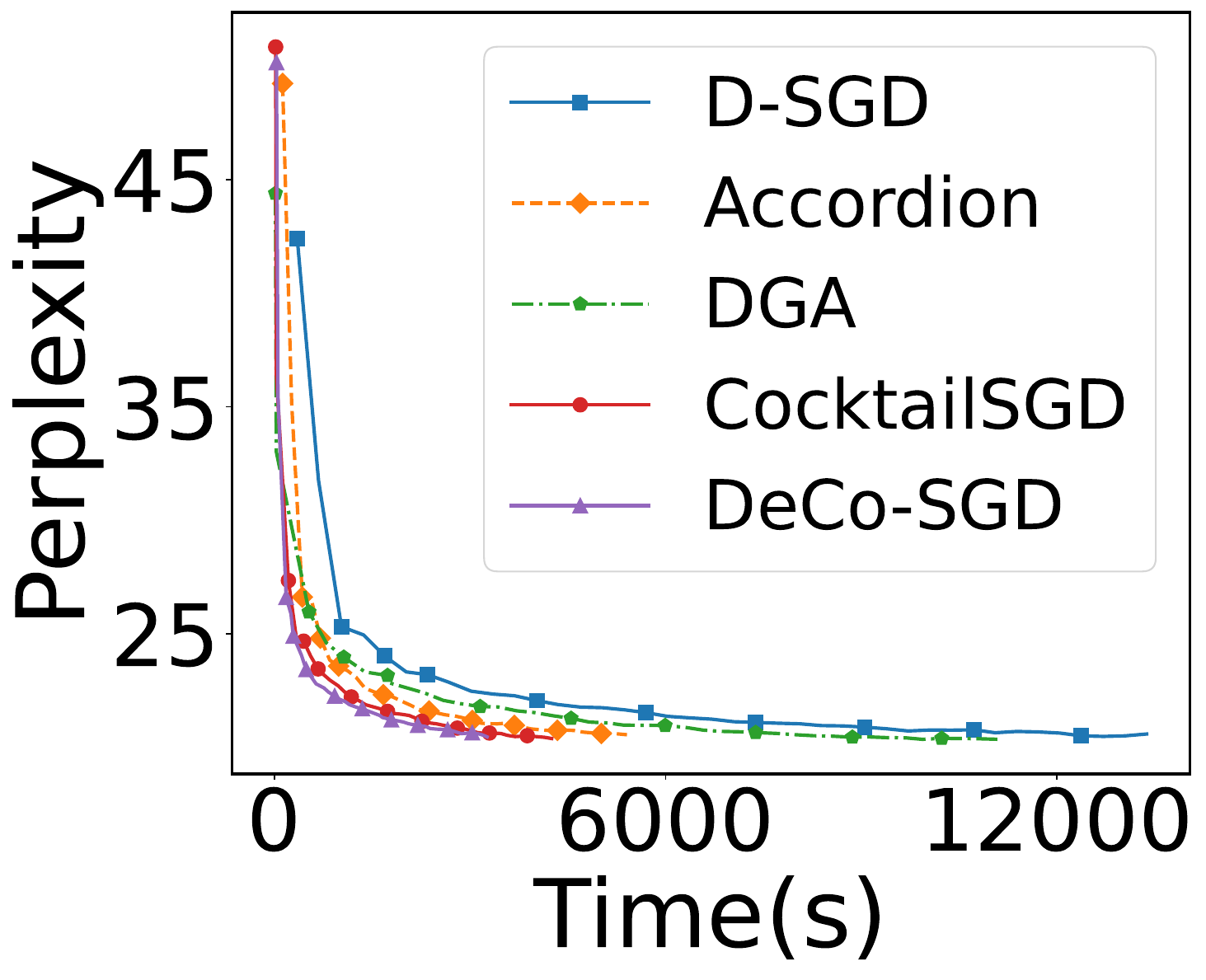}
    \caption*{\shortstack{(h) GPT@Wikitext \\ $n=32$}}
\end{minipage}

\caption{Convergence performance across different numbers of worker nodes ($n=4,8,16,32$) for five distributed optimization methods. \textbf{Top row:} Accuracy over time for ViT@ImageNet. \textbf{Bottom row:} Perplexity over time for GPT@Wikitext.}
\label{fig:combined_figure}
\end{figure}

Fig.~\ref{fig:combined_figure} provides the detailed accuracy/perplexity curves corresponding to Fig.~\ref{fig:bar}.

\subsection{Detailed Results of Section~\ref{exp.4}}

\begin{figure}[h]
\centering
\begin{minipage}{0.23\linewidth}
    \centering
    \includegraphics[width=\linewidth]{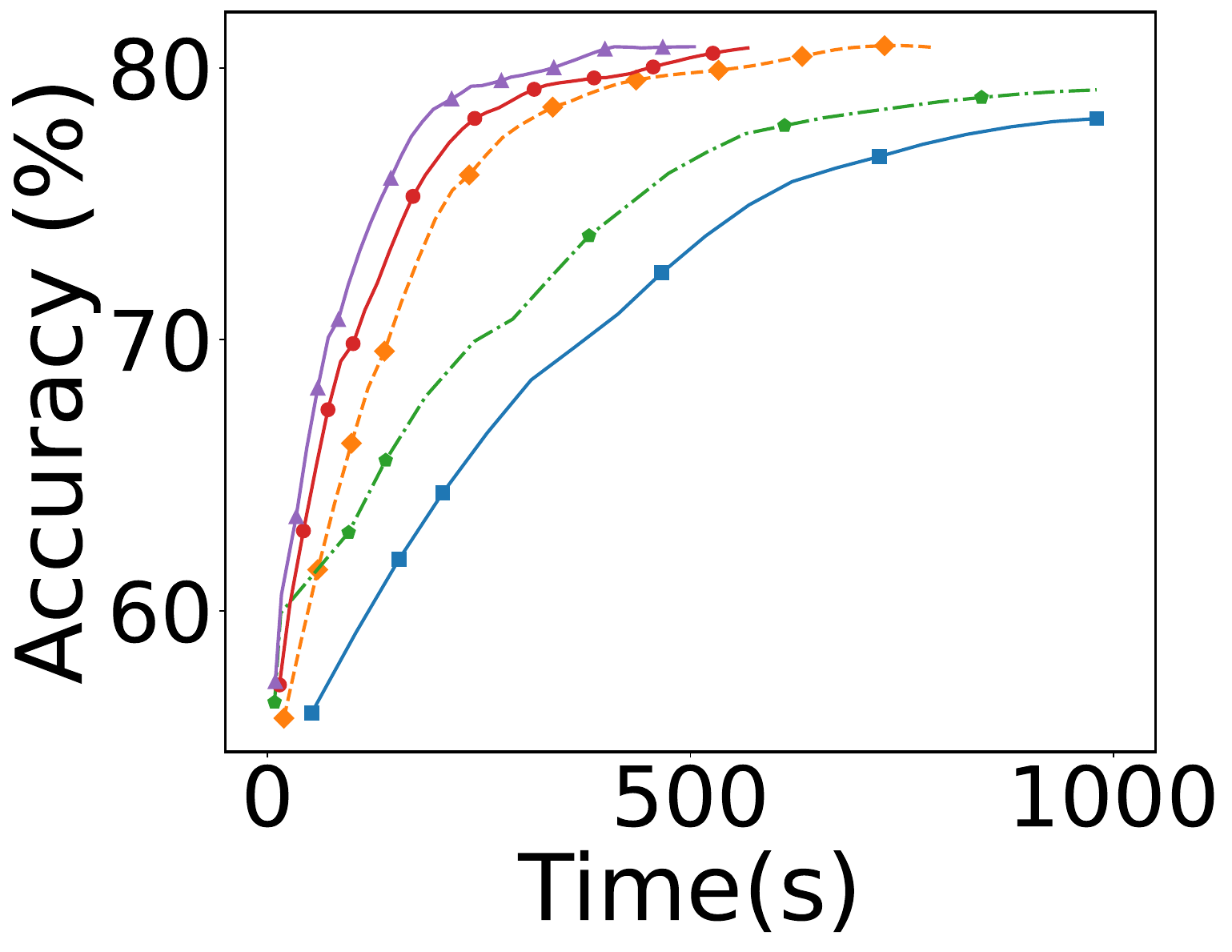}
    \caption*{\shortstack{(a) ViT@ImageNet \\ bandwidth=$100$Mbps \\ latency=$100$ms}}
\end{minipage}
\hfill
\begin{minipage}{0.23\linewidth}
    \centering
    \includegraphics[width=\linewidth]{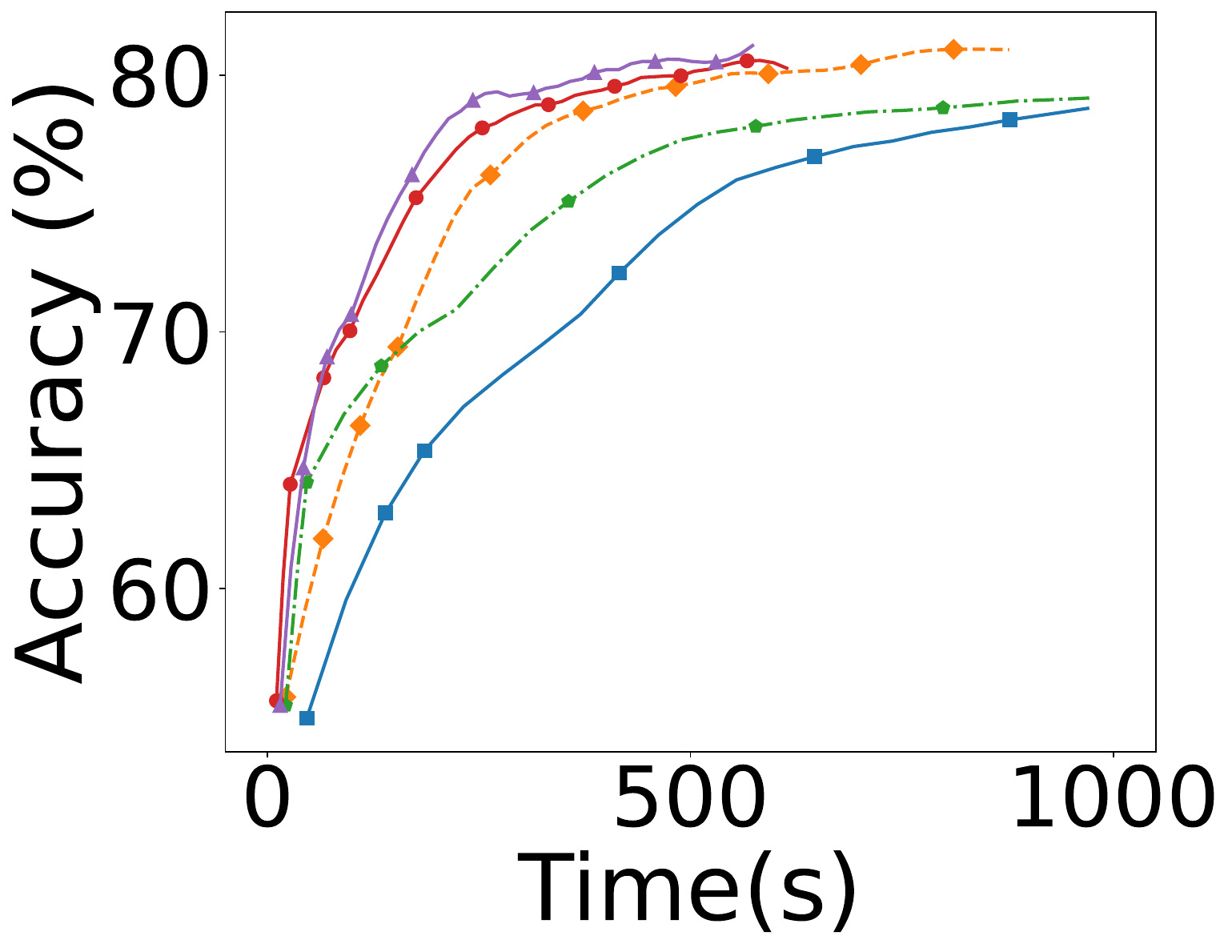}
    \caption*{\shortstack{(b) ViT@ImageNet \\ bandwidth=$100$Mbps \\ latency=$1000$ms}}
\end{minipage}
\hfill
\begin{minipage}{0.23\linewidth}
    \centering
    \includegraphics[width=\linewidth]{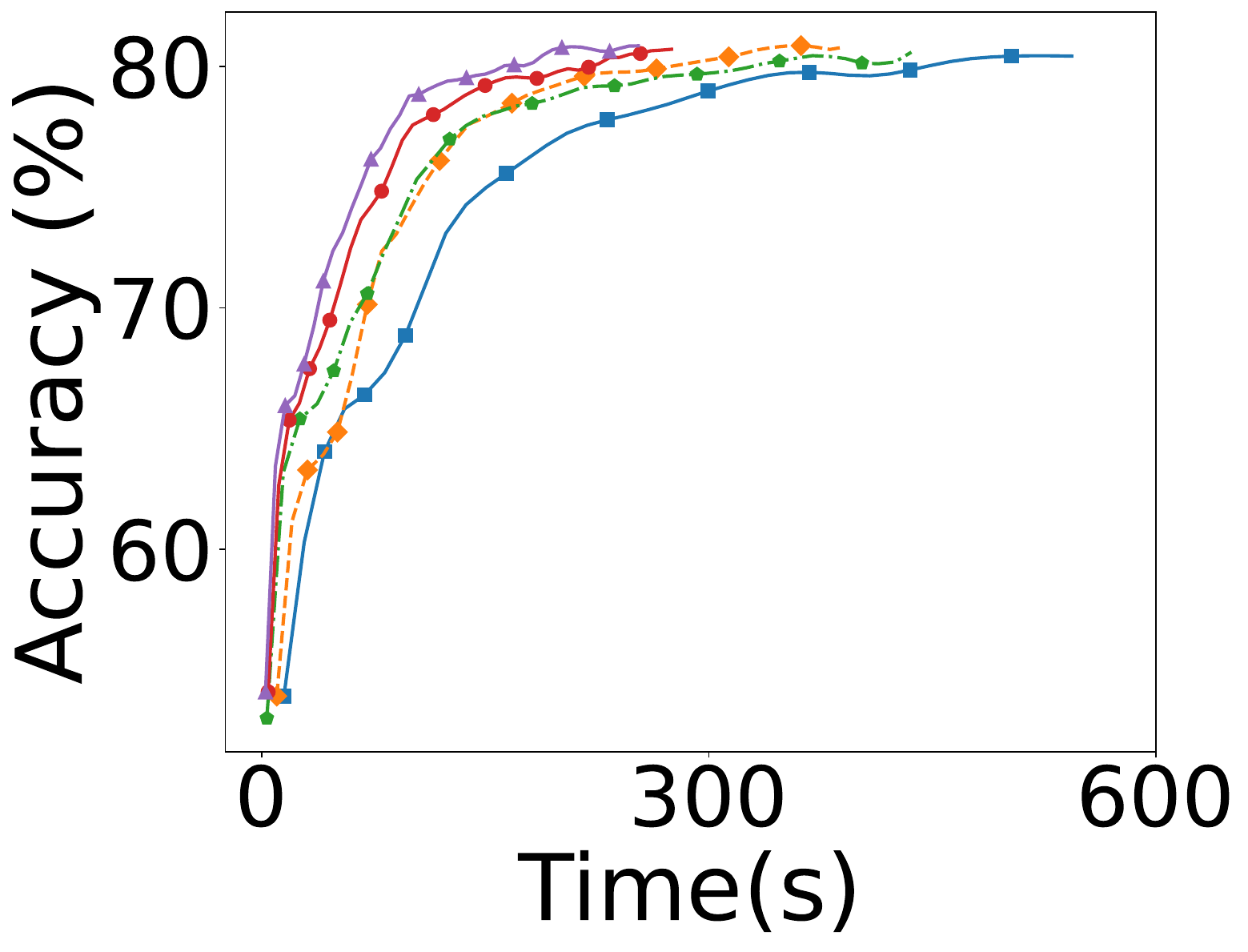}
    \caption*{\shortstack{(c) ViT@ImageNet \\ bandwidth=$500$Mbps \\ latency=$100$ms}}
\end{minipage}
\hfill
\begin{minipage}{0.23\linewidth}
    \centering
    \includegraphics[width=\linewidth]{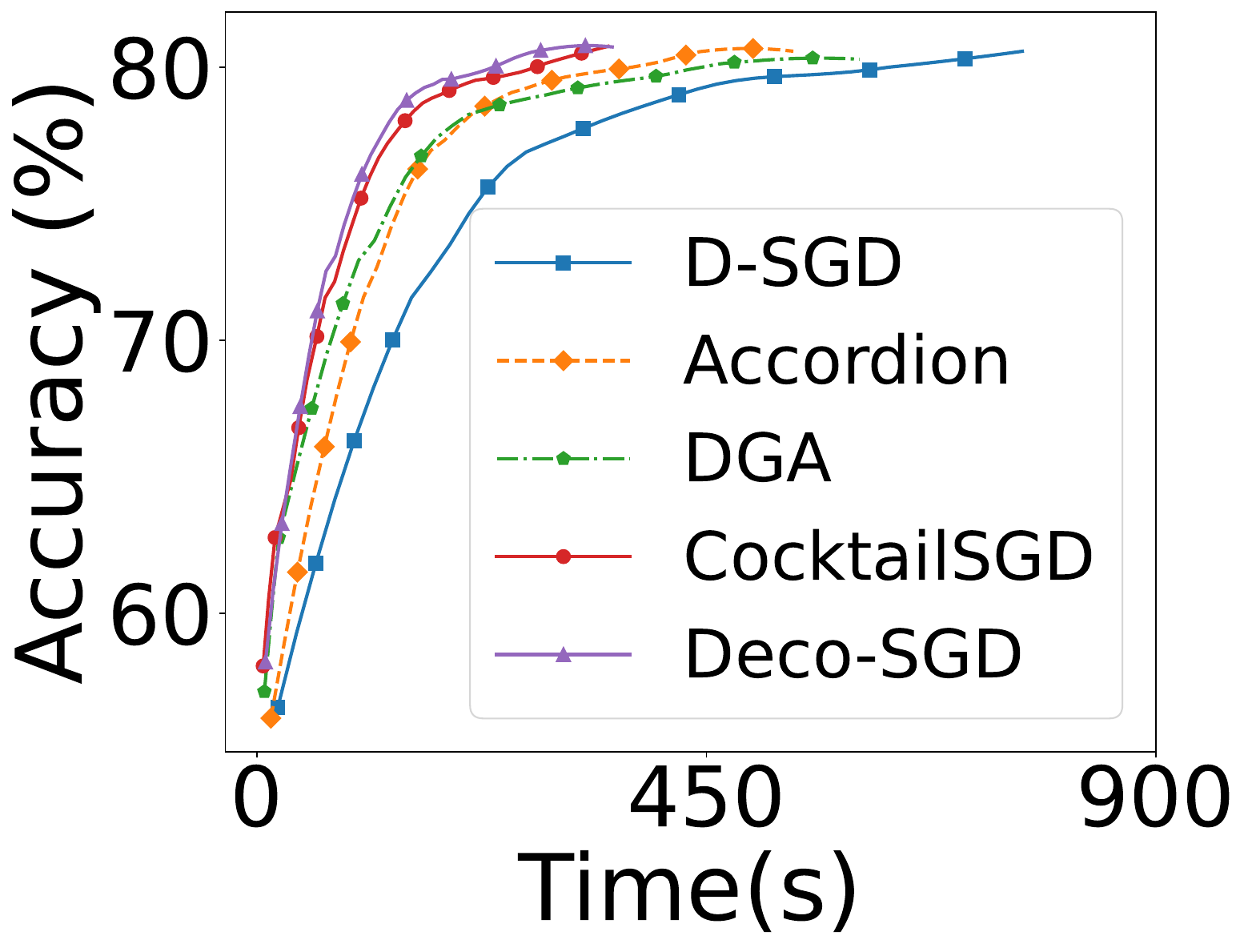}
    \caption*{\shortstack{(d) ViT@ImageNet \\ bandwidth=$500$Mbps \\ latency=$1000$ms}}
\end{minipage}

\vspace{2mm} 

\begin{minipage}{0.23\linewidth}
    \centering
    \includegraphics[width=\linewidth]{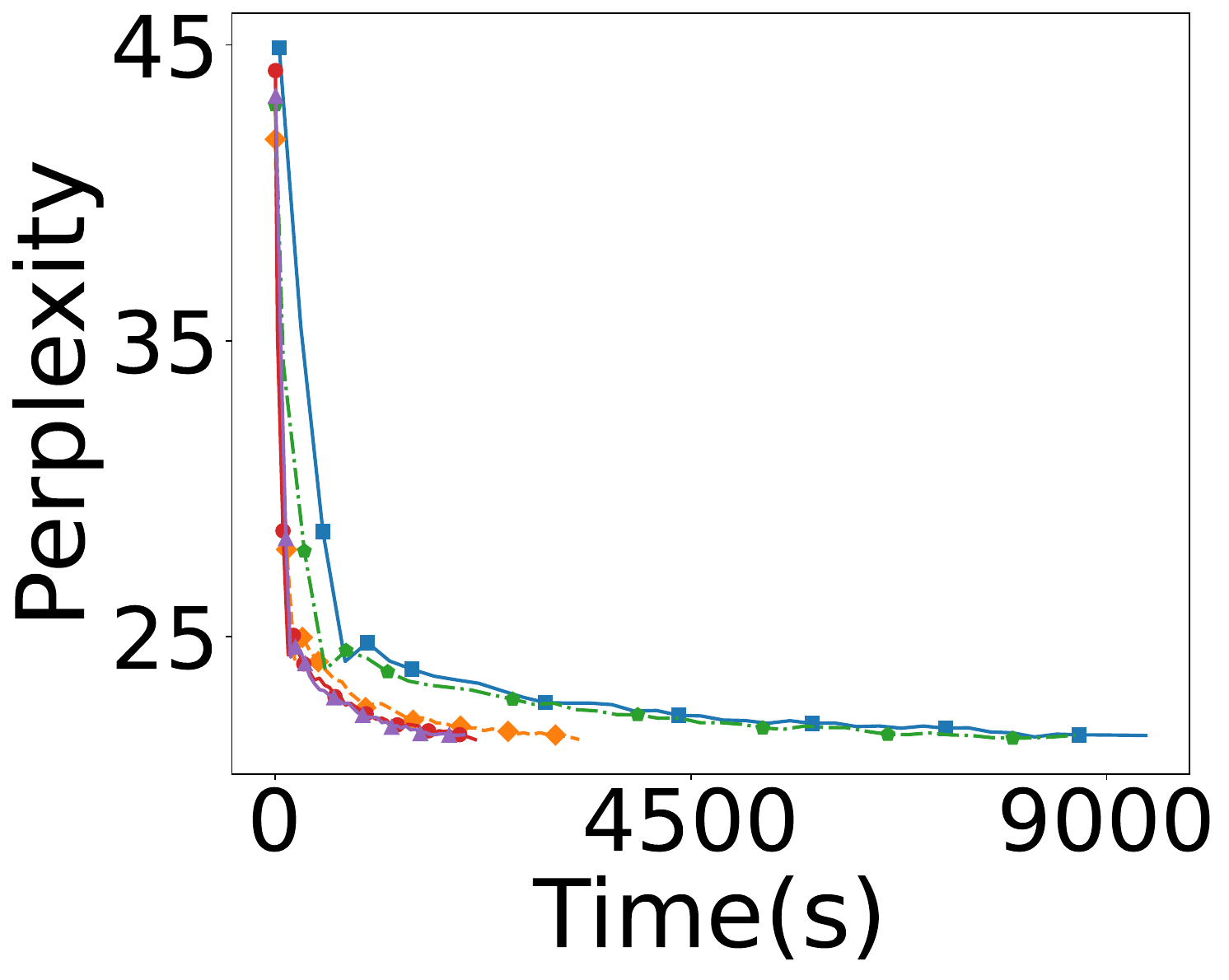}
    \caption*{\shortstack{(e) GPT@Wikitext \\ bandwidth=$100$Mbps \\ latency=$100$ms}}
\end{minipage}
\hfill
\begin{minipage}{0.23\linewidth}
    \centering
    \includegraphics[width=\linewidth]{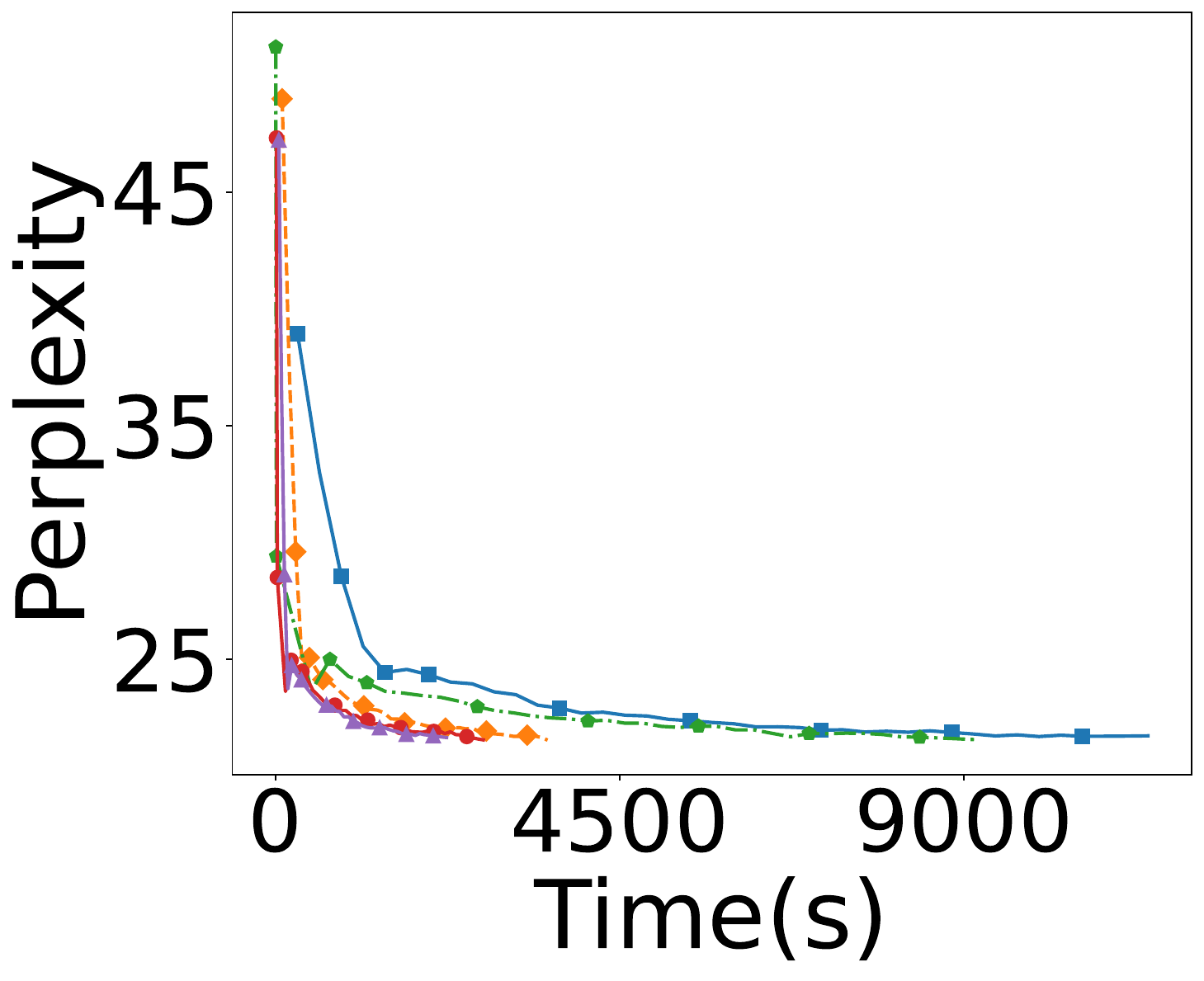}
    \caption*{\shortstack{(f) GPT@Wikitext \\ bandwidth=$100$Mbps \\ latency=$1000$ms}}
\end{minipage}
\hfill
\begin{minipage}{0.23\linewidth}
    \centering
    \includegraphics[width=\linewidth]{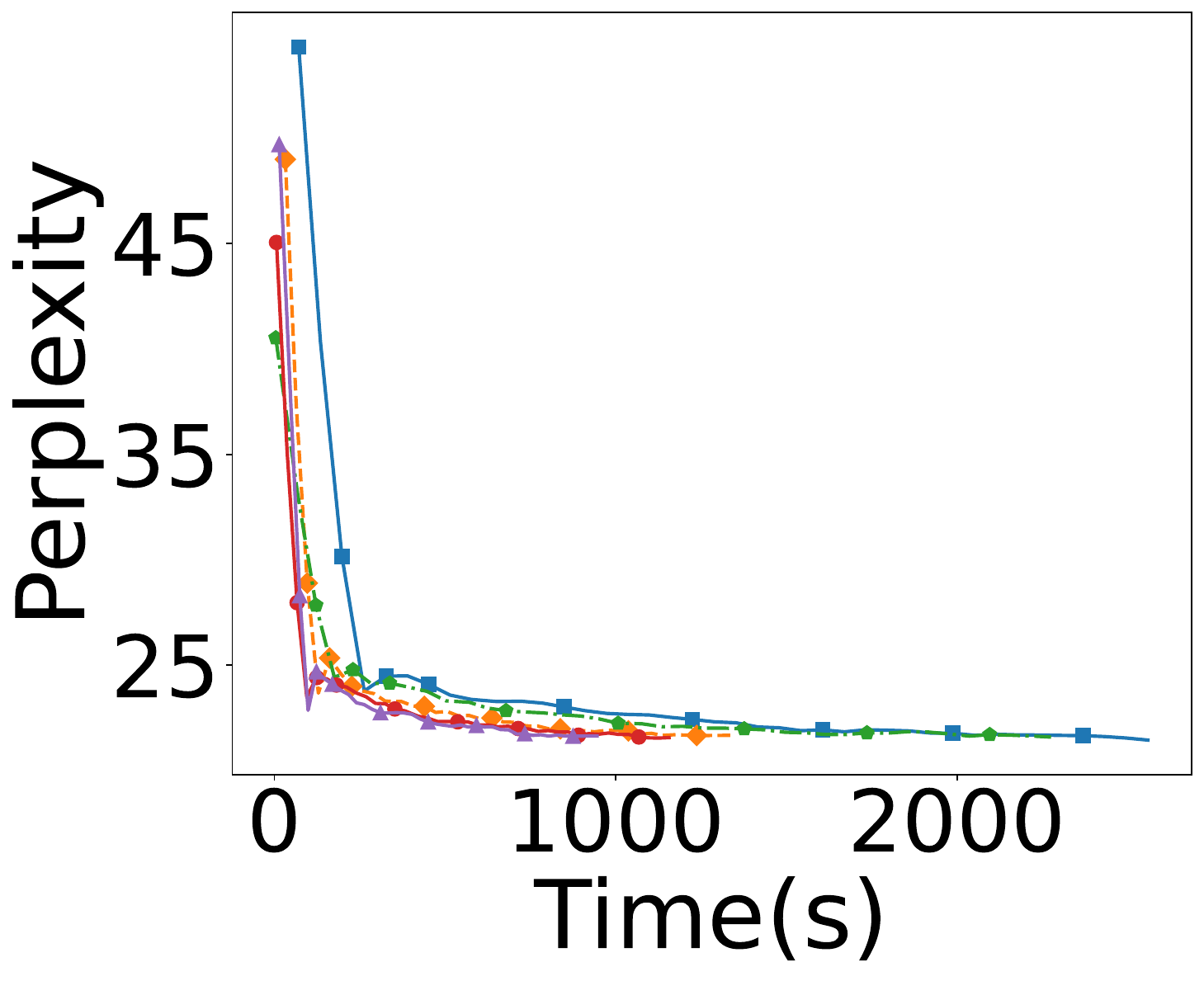}
    \caption*{\shortstack{(g) GPT@Wikitext \\ bandwidth=$500$Mbps \\ latency=$100$ms}}
\end{minipage}
\hfill
\begin{minipage}{0.23\linewidth}
    \centering
    \includegraphics[width=\linewidth]{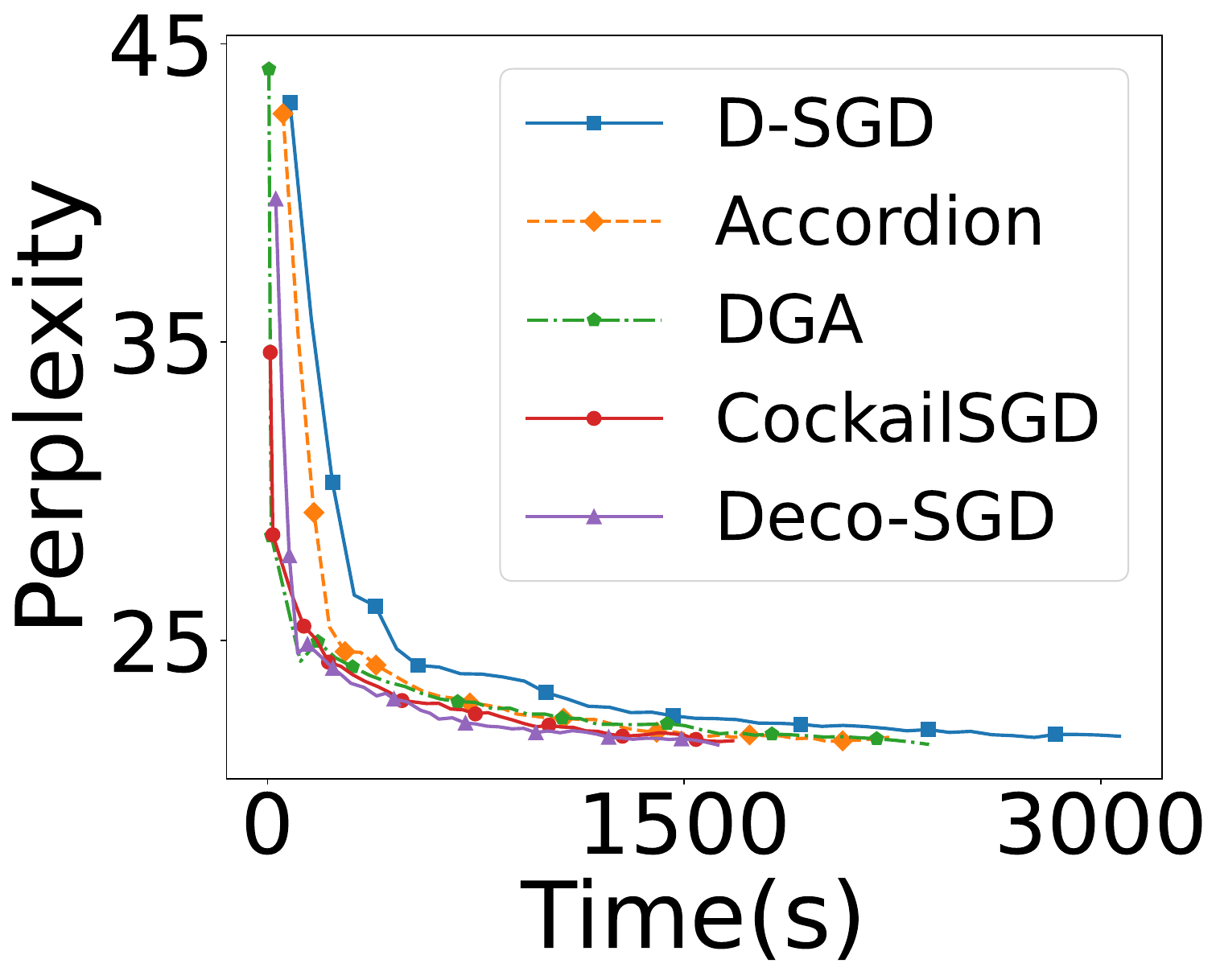}
    \caption*{\shortstack{(h) GPT@Wikitext \\ bandwidth=$500$Mbps \\ latency=$1000$ms}}
\end{minipage}

\caption{Convergence performance under different bandwidth-latency settings for five distributed optimization methods. \textbf{Top row:} Accuracy over time for ViT@ImageNet. \textbf{Bottom row:} Perplexity over time for GPT@Wikitext. Columns correspond to $(100\mathrm{Mbps},100\mathrm{ms})$, $(100\mathrm{Mbps},1000\mathrm{ms})$, $(500\mathrm{Mbps},100\mathrm{ms})$, and $(500\mathrm{Mbps},1000\mathrm{ms})$.}
\label{fig:combined_vary}
\end{figure}

Fig.~\ref{fig:combined_vary} provides a visual counterpart to the data presented in Table~\ref{tab:iid} in the main text. 

\subsection{Experiment Results in non-IID settings}
The non-IID partition strategy is distribution-based label imbalance, which allocates a proportion of samples of each label according to Dirichlet distribution, with the concentration parameter set to 0.5. The results demonstrate that under non-IID settings, \sysname \ consistently outperforms other methods.

\subsection{Experiment Results with other sparsifiers}
We compare \sysname\ with other optimizers under another sparsifier Random-K, which randomly selects k\% gradient elements to transmit. The results below show that \sysname\ consistently outperforms other methods with other sparsifiers.

\begin{table*}[!t]
\centering
\caption{Comparison of training time for \sysname \ and other optimizers under the Random-K sparsifier, which randomly selects k\% gradient elements to transmit.}
\label{tab:randk}
\scalebox{0.45}{
\begin{tabular}{@{}lcccccccc@{}}
\toprule
\makecell{\textbf{Model} \\ \textbf{@DataSet}} & \makecell{\textbf{a(Gbps)}, \\ \textbf{b(s)}}  & \textbf{D-SGD} & \textbf{DAGC} & \textbf{Accordion} & \textbf{DC2} & \textbf{DGA} & \textbf{CocktailSGD} & \textbf{\sysname} \\
\midrule
\multirow{4}{*}{\makecell{GPT\\@Wikitext}}
  & $0.1, 0.1$  & $7276.34 \pm 49.33$ ($5.13\times$) & $ 2283.53 \pm 12.55$ ($1.61\times$)& $ 2246.34 \pm 12.24$ ($1.58\times$)& $ 2244.73 \pm 12.21$ ($1.58\times$)& $ 6108.13 \pm 35.23$ ($4.31\times$)  & $1732.25 \pm 12.58$ ($1.22\times$) & $\mathbf{1417.55 \pm 11.23}$ \\
  & $0.5, 0.1$   & $1785.64 \pm 12.55$ ($2.41\times$) & $969.11 \pm 7.12$ ($1.31\times$)  & $921.27 \pm 7.56$ ($1.24\times$)  &  $917.54 \pm 7.42$ ($1.24\times$)  & $1515.61 \pm 12.48$ ($2.04\times$) & $815.52 \pm 6.45$ ($1.10\times$) & $\mathbf{741.35 \pm 5.73}$ \\
  & $0.1, 1.0$  & $8179.46 \pm 54.32$ ($4.76\times$) & $2841.85 \pm 20.07$ ($1.65\times$) & $2812.31 \pm 20.35$ ($1.64\times$) & $2827.36 \pm 20.91$ ($1.65\times$) & $6517.52 \pm 34.66$ ($3.79\times$) & $ 2018.44 \pm 14.65$ ($1.17\times$) & $\mathbf{1718.41 \pm 13.46}$ \\
  & $0.5, 1.0$  & $2736.48 \pm 19.64$ ($2.31\times$) & $1576.29 \pm 12.98$ ($1.33\times$) & $1566.14 \pm 12.11$ ($1.32\times$) & $1570.48 \pm 12.07$ ($1.32\times$) & $1775.52 \pm 13.29$ ($1.50\times$) & $1459.16 \pm 11.63$ ($1.23\times$) & $\mathbf{1186.33 \pm 9.27}$ \\
\midrule
\multirow{4}{*}{\makecell{ViT\\@ImageNet}} 
  & $0.1, 0.1$   & $1273.93 \pm 9.32$ ($4.64\times$) & $567.73 \pm 3.73$ ($2.07\times$) & $545.18 \pm 3.45$ ($1.98\times$) & $537.85 \pm 3.14$ ($1.96\times$) & $1324.88 \pm 10.26$ ($4.82\times$) & $368.39 \pm 1.82$ ($1.34\times$) & $\mathbf{274.76 \pm 1.45}$\\
  & $0.5, 0.1$   & $434.16 \pm 2.15$ ($2.34\times$) & $364.22 \pm 1.80$ ($1.96\times$) & $347.22 \pm 1.73$ ($1.87\times$) & $353.70 \pm 1.79$ ($1.90\times$) & $467.97 \pm 2.10$ ($2.52\times$) & $224.09 \pm 1.38$ ($1.21\times$) & $\mathbf{185.90 \pm 0.68}$\\
  & $0.1, 1.0$   & $1724.74 \pm 10.47$ ($4.22\times$)  & $672.44 \pm 3.21$ ($1.64\times$) & $654.28 \pm 3.09$ ($1.60\times$) & $644.02 \pm 3.14$ ($1.57\times$) & $1367.39 \pm 8.25$ ($3.34\times$) & $555.26 \pm 3.11$ ($1.36\times$) & $\mathbf{409.18 \pm 2.48}$\\
  & $0.5, 1.0$   & $677.23 \pm 4.21$ ($2.78\times$) & $489.41 \pm 2.21$ ($2.01\times$) & $467.39 \pm 2.07$ ($1.92\times$) & $478.93 \pm 1.85$ ($1.97\times$) & $599.93 \pm 2.83$ ($2.46\times$) & $283.89 \pm 1.62$ ($1.17\times$) & $\mathbf{243.52 \pm 1.02}$\\
\bottomrule
\end{tabular}
}
\end{table*}

\subsection{Visualization of the Coupling Term $\phi$}\label{appendix:phi-vis}
\begin{figure}[h]
    \centering
    \includegraphics[width=0.7\linewidth]{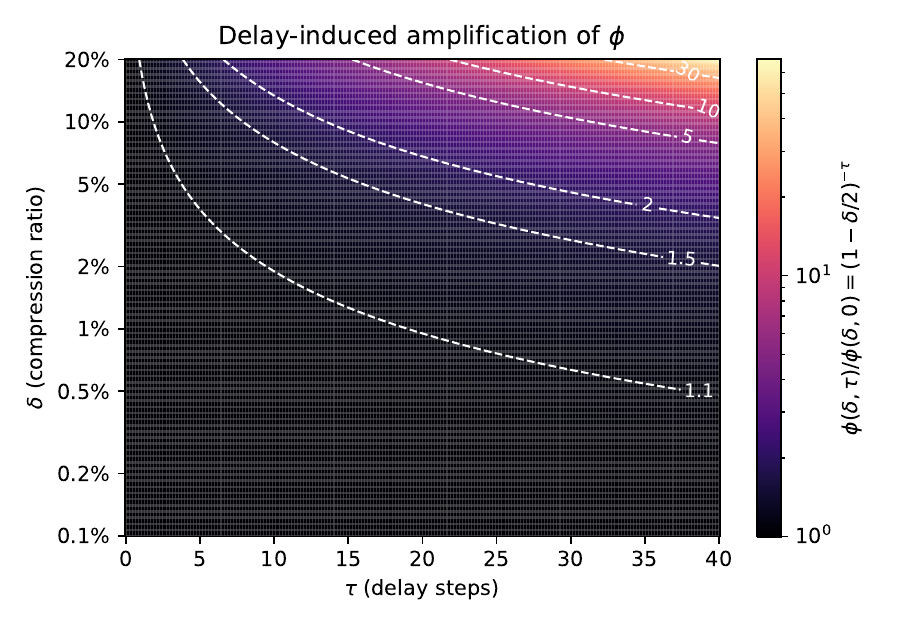}
    \caption{Delay-induced amplification factor $\phi(\delta,\tau)/\phi(\delta,0)=(1-\delta/2)^{-\tau}$ of the coupling term $\phi$.}
    \label{fig:phi-amplification}
\end{figure}

Fig.~\ref{fig:phi-amplification} visualizes the delay-induced amplification factor $(1-\delta/2)^{-\tau}$ of $\phi(\delta,\tau)$ over a wider $(\delta,\tau)$ range, complementing the discussion in Remark~\ref{remark phi}.

\subsection{Profiling of Computation and Communication Overheads}
\label{app:profiling-overhead}
Table~\ref{tab:profiling-overhead} profiles the runtime overhead under the same setting as Sec.~\ref{exp.2}. The compression overhead includes Top-$k$ selection, compression, and decompression. The results show that compression/decompression accounts for less than $1\%$ of $T_{\mathrm{comp}}$, while DeCo decision-making takes less than $10^{-6}$ s.

\begin{table}[!t]
\centering
\caption{Profiling of computation and communication overheads under the same setting as Sec.~\ref{exp.2}.}
\label{tab:profiling-overhead}
\resizebox{\columnwidth}{!}{
\begin{tabular}{@{}lccccc@{}}
\toprule
\makecell{\textbf{Model}\\\textbf{@DataSet}} &
\makecell{\textbf{Comp.}\\\textbf{overhead (s)}} &
\makecell{\textbf{Comp.}/\\\textbf{$T_{\mathrm{comp}}$}} &
\makecell{\textbf{$T_{\mathrm{comp}}$}\\\textbf{(s)}} &
\makecell{\textbf{DeCo}\\\textbf{decision (s)}} &
\makecell{\textbf{Comm. time}\\\textbf{saved (s)}} \\
\midrule
GPT@Wikitext & 0.0236 & 0.71\% & 3.33 & $<10^{-6}$ & 8.96 \\
ViT@ImageNet & 0.0205 & 0.62\% & 3.29 & $<10^{-6}$ & 6.22 \\
\bottomrule
\end{tabular}
}
\end{table}

\subsection{Additional Scalability Results (32-Worker Detailed Numbers)}
\label{app:scalability32}
Table~\ref{tab:scalability32} reports the detailed $32$-worker results under the same bandwidth-latency settings used in Table~\ref{tab:iid}. Across both GPT@Wikitext and ViT@ImageNet, \sysname{} consistently achieves the shortest training time, confirming that its adaptive selection of $\delta$ and $\tau$ remains effective as the number of workers increases.

\begin{table*}[!t]
\centering
\caption{Comparison of training time (s) under different methods and network conditions with $32$ workers. $a$ is the maximum bandwidth and $b$ is the end-to-end latency.}
\label{tab:scalability32}
\scalebox{0.77}{
\begin{tabular}{@{}lcccccccc@{}}
\toprule
\makecell{\textbf{Model} \\ \textbf{@DataSet}} & \makecell{\textbf{a(Gbps)}, \\ \textbf{b(s)}} &  \textbf{D-SGD}  & \textbf{Accordion} & \textbf{DC2} & \textbf{DGA} & \textbf{CocktailSGD} & \textbf{\sysname} \\
\midrule
\multirow{4}{*}{\makecell{GPT\\@Wikitext}}
  & $0.1, 0.1$  & $4124.49 \pm 31.05$ ($3.78\times$) & $1972.88 \pm 10.85$ (\sr$1.81\times$\er) & $1812.74 \pm 14.79$ ($1.66\times$) & $2706.01 \pm 21.69$ ($2.48\times$) & $1326.91 \pm 10.22$ ($1.22\times$) & $\mathbf{1091.51 \pm 8.42}$ \\
  & $0.5, 0.1$  & $1013.15 \pm 8.39$ ($1.89\times$) & $736.52 \pm 4.42$ ($1.37\times$) & $751.48 \pm 4.66$ ($1.40\times$) & $930.76 \pm 5.86$ ($1.73\times$) & $616.20 \pm 4.50$ ($1.15\times$) & $\mathbf{537.10 \pm 4.12}$ \\
  & $0.1, 1.0$  & $4663.46 \pm 34.45$ ($3.87\times$) & $2232.86 \pm 18.73$ ($1.85\times$) & $2291.65 \pm 16.73$ ($1.90\times$) & $3832.70 \pm 23.34$ ($3.18\times$) & $1454.74 \pm 10.04$ ($1.21\times$) & $\mathbf{1203.73 \pm 9.33}$ \\
  & $0.5, 1.0$  & $1524.04 \pm 11.37$ ($1.83\times$) & $1137.05 \pm 7.39$ ($1.36\times$) & $1128.24 \pm 7.22$ ($1.35\times$) & $1430.38 \pm 9.42$ ($1.72\times$) & $978.95 \pm 7.44$ ($1.17\times$) & $\mathbf{833.28 \pm 6.40}$ \\
\midrule
\multirow{4}{*}{\makecell{ViT\\@ImageNet}}
  & $0.1, 0.1$  & $989.45 \pm 7.47$ ($5.08\times$) & $349.86 \pm 1.85$ (\sr$1.80\times$\er) & $330.42 \pm 2.72$ ($1.70\times$) & $718.05 \pm 5.05$ ($3.69\times$) & $242.32 \pm 1.16$ ($1.24\times$) & $\mathbf{194.83 \pm 1.08}$ \\
  & $0.5, 0.1$  & $375.16 \pm 2.12$ ($3.22\times$) & $186.51 \pm 0.93$ ($1.60\times$) & $183.94 \pm 1.67$ ($1.58\times$) & $232.55 \pm 1.28$ ($2.00\times$) & $137.30 \pm 0.62$ ($1.18\times$) & $\mathbf{116.41 \pm 0.93}$ \\
  & $0.1, 1.0$  & $1081.36 \pm 6.61$ ($3.84\times$) & $531.97 \pm 2.77$ ($1.89\times$) & $540.15 \pm 3.86$ ($1.92\times$) & $986.22 \pm 5.72$ ($3.50\times$) & $345.48 \pm 1.73$ ($1.23\times$) & $\mathbf{281.63 \pm 1.36}$ \\
  & $0.5, 1.0$  & $562.76 \pm 3.44$ ($3.36\times$) & $314.26 \pm 1.61$ ($1.87\times$) & $305.24 \pm 1.53$ ($1.82\times$) & $402.48 \pm 2.25$ ($2.40\times$) & $204.92 \pm 0.96$ ($1.22\times$) & $\mathbf{167.63 \pm 0.84}$ \\
\bottomrule
\end{tabular}
}
\end{table*}

\end{document}